\documentclass[default]{sn-jnl}

\usepackage{url}
\usepackage[utf8]{inputenc}
\usepackage{graphicx}
\usepackage{amsmath}
\usepackage{amsthm}
\usepackage[small]{caption}
\usepackage{caption}
\usepackage{subcaption}
\usepackage{latexsym}
\usepackage{comment}
\usepackage{float}
\usepackage{booktabs}

\jyear{2021}%

\theoremstyle{thmstyleone}%
\theoremstyle{thmstyletwo}%

\theoremstyle{thmstylethree}%

\begin{document}

\title[Telling More with Concepts and Relations]{Telling More with Concepts and Relations: Exploring and Evaluating Classifier Decisions with CoReX}


\author*[1]{\fnm{Bettina} \sur{Finzel}}\email{bettina.finzel@uni-bamberg.de}

\author[1]{\fnm{Patrick} \sur{Hilme}}\email{patrick-jeremias.hilme@stud.uni-bamberg.de}

\author[1]{\fnm{Johannes} \sur{Rabold}}

\author[1]{\fnm{Ute} \sur{Schmid}}\email{johannes.rabold,ute.schmid@uni-bamberg.de}

\affil*[1]{\orgdiv{Cognitive Systems}, \orgname{University of Bamberg}, \orgaddress{\street{Weberei 5}, \city{Bamberg}, \postcode{96047}, \country{Germany}}}

\abstract{}
\scriptsize
Explanations for Convolutional Neural Networks (CNNs) based on re\-levance of input pixels might be too unspecific to evaluate which and how input features impact model decisions. Especially in complex real-world domains like biology, the presence of specific concepts and of relations between concepts might be discriminating between classes. Pixel relevance is not expressive enough to convey this type of information. In consequence, model evaluation is limited and relevant aspects present in the data and influencing the model decisions might be overlooked. This work presents a novel method to explain and evaluate CNN models, which uses a concept- and relation-based explainer (CoReX). It explains the predictive behavior of a model on a set of images by masking (ir-)relevant concepts from the decision-making process and by constraining relations in a learned interpretable surrogate model.
We test our approach with several image data sets and CNN architectures. Results show that CoReX explanations are faithful to the CNN model in terms of predictive outcomes. We further demonstrate through a human evaluation that CoReX is a suitable tool for generating combined explanations that help assessing the classification quality of CNNs. We further show that CoReX supports the identification and re-classification of incorrect or ambiguous classifications.


\normalsize



\maketitle

%
%
%
%
%
%

\normalsize
\section{Introduction}

In recent years, research in Explainable Artificial Intelligence (XAI) has produced a vast amount of explanatory methods that make transparent what deep learning models, such as Convolutional Neural Networks (CNNs), have learned \cite{adadi2018survey,arrieta2020xai,schwalbe2023comprehensive}. Most methods developed for CNNs highlight pixels or groups of pixels in images that were relevant to class predictions for individual samples (via heatmaps), e.g., LIME \cite{ribeiro2016lime}, or layer-wise relevance propagation (LRP) \cite{lapuschkin2019lrp}. It remains the task of the human to interpret what features the highlighted image regions contain and to evaluate whether meaningful features were learned for class discrimination. As long as it is a matter of the presence of relevant or irrelevant simple features, such visual explanations are sufficiently expressive to fit the evaluation task. However, as soon as several feature expressions apply simultaneously (colors, textures, shapes, etc.) and also when the spatial constellation of features in an image contribute to the characterization of a class, heatmaps reach their limits in expressiveness. In such cases, explanatory approaches are needed that ascribe meaningful and distinguishable concepts to the learned features and which can take into account relations between these concepts.

\begin{figure}
        \centering
        \includegraphics[width=\columnwidth]{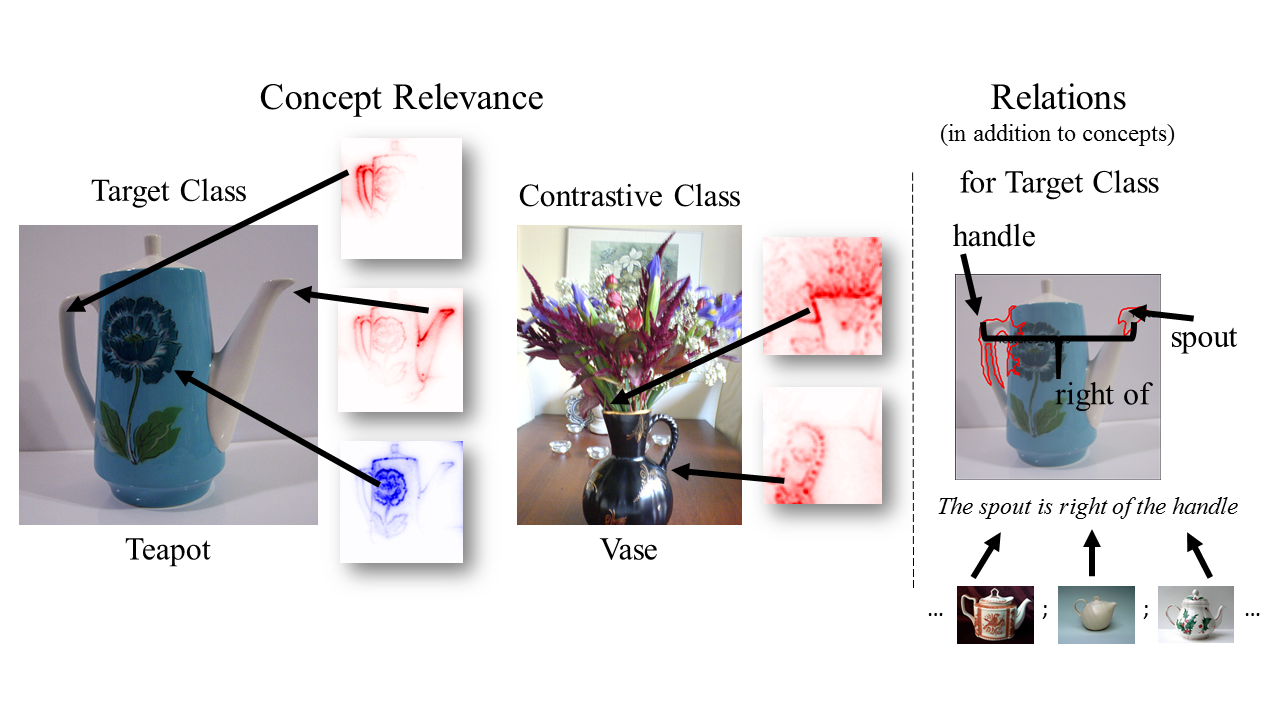} 
    \caption{Explaining why a sample image belongs to the target class ``teapot'' by contrasting it with a sample from the contrastive class ``vase'' based on identified concepts (handle, spout) and their relations (``spout right of handle'')}
    \label{fig:running_example}
\end{figure}

In general, concepts are defined as mental representations for categories of objects \cite{smith2013categories}. Concepts can be referred to with natural language, typically nouns. For instance, the objects in Figure~\ref{fig:running_example} are referred to as ``vase'' and as ``teapot''. Natural categories such as different animals and also many human-made objects often have no crisp decision boundary \cite{rosch1973natural}. Depending on the shape, and the presence of a handle or lid, an object might be identified as a teapot or not. 
Many concepts are members of a super-concept (a teapot belongs to crockery) and also have sub-concepts (e.g., Japanese teapots) \cite{rosch1973natural}. Often, concepts can be characterized by the structure of their parts \cite{simons2000parts}. For instance, a handle, a spout, and a lid are parts of a teapot.

In machine learning, concept learning is a special case of classification learning which approximates a binary target function that decides whether an instance belongs to a target class (concept) or not \cite{mitchell1997machine}.
In knowledge representation, ontologies are used to model domains of discourse by concept hierarchies \cite{simons2000parts}. To derive such ontologies, concept analysis has been introduced \cite{wille2005formal}. In general, the inclusion of domain knowledge can lead to more expressiveness and allows for more comprehensive evaluations of learned models. For example, knowledge about spatial relations can be used to perform reasoning on the structure of found concepts in images \cite{renz2002qualitative}. In the spatial domain, reasoning on position, orientation, and distance are of particular interest (`It is a teapot and not a vase because there is a handle right of a spout.', see image on the right in Figure~\ref{fig:running_example}). Why considering concepts \textit{and} relations at the same time? The separation of classes (here: ``teapot'' or ``vase'') may be based on a common set of present concepts (e.g., handle, spout, flowers) and may therefore only be explainable if relevance of \textit{and} spatial relations between those concepts are taken into account. 
Our approach extracts and localizes concepts for similar classes based on concept relevance representing the contribution of conjunct pixel groups to an image classification. Concepts positively contributing to a classification get positive relevance scores (red in Figure~\ref{fig:running_example}), negatively contributing concepts get negative scores (blue in Figure~\ref{fig:running_example}), whereas irrelevant concepts get a score near or equal zero (white in Figure~\ref{fig:running_example}). Concepts can be labelled according to their appearance in images, where they have the highest absolute relevance score. We can learn from the above ``teapot vs. vase'' example that high relevance scores may correlate with similar concepts across classes (e.g., spout, handle) or that classes may differ by concept relevance (e.g., negative relevance for flowers on teapots). For our selected teapot example, we can explain the classification for the teapot by the relation ``The example shows a teapot and not a vase, because the spout is right of the handle'', which holds, e.g., for all teapots oriented to the right in contrast to vases. Apparently, this finding does not hold for all teapots, but for a sub-group or cluster within the data.

For many real-world domains, objects are more or less typical members of their category \cite{rosch1973natural}. 
Where categories are similar, it is more difficult to correctly classify objects near the category border \cite{mckee2014task}. A vase with a bulbous body might be confused with a teapot; a teapot missing a handle might be confused with a vase. The same problem occurs in critical domains such as medicine. For instance, whether a tissue sample is indicative of a tumor class is harder to decide for borderline cases than for prototypical tissue patterns \cite{bruckert2020companions}.

The application of XAI methods supports humans in comprehending how black box models such as CNNs derive a specific class output for a given input instance. This information can be used to examine the learned model. For instance, \textit{Clever Hans} effects can be identified, that is, that the model output is right, but for the wrong reason, e.g., an image is classified as a horse, but based on copyright text or the background \cite{lapuschkin2019unmasking,stammer2022rrr}. Furthermore, learned models might be prone to make more classification errors near the decision boundaries between classes than near the center of the decision area. Several XAI methods have been proposed to support the exploration of decision boundaries of learned models. First, example-based explanations providing information about prototypes and critics have been suggested \cite{kim2016examples}. Furthermore, contrastive explanations, where the neighborhood of a classified example in the instance space is explored have been shown to be highly efficient \cite{teso2019explanatory,miller2021contrastive,rabold2022nearmisses,stammer2022rrr}. Finally, cluster analysis, where clusters may be formed around very common or unusual or biased samples, has been demonstrated to be helpful \cite{lapuschkin2019unmasking}. In general, contrastive explanations provide information about what has to be given or what has to be absent to make an object an instance of a specific concept. When used to explain a model to a human, contrastive examples uncover classification errors which might be corrected by providing new examples or constraints on the learning and prediction process to shift the decision boundary \cite{teso2019explanatory,dash2022review}.

Attribution and backpropagation methods explain class predictions of a CNN (or other model types) by relating the class output to relevant features in the input. For image data, these features correspond to pixels or groups of pixels. Prediction boundaries are observed then on the prediction level, that is, evaluating whether a (contrastive) example is classified correctly or not. More recently, different approaches have been proposed to take into account information represented in intermediate convolutional layers of CNNs which will be discussed below \cite{bau2017network,fong2018net2vec,rabold2019relational,achtibat2022crp,achtibat2023from}. Given information about concepts represented in intermediate layers of a CNN together with relations between such concepts allows us to provide more specific explanations, namely whether the prediction is based on the right concepts and relations or not. Representing both, concepts and relations in a human-understandable way further supports meaningful interaction with data, explanations, and the model itself. 

Given that heatmaps cannot adequately explain the classification of a CNN model on complex, spatially-descriptive image data and that human-understandable explanations are needed that allow the decision boundary to be explored, we present a novel approach: the Concept- and relation-based explainer (CoReX). Our approach learns relations on top of concepts with the help of Inductive Logic Programming (ILP), which is inherently interpretable \cite{muggleton2018ultra}. CoReX makes the following contribution to the state of the art:

\begin{itemize}
    \item It combines relevance-based concept extraction with interpretable relational learning to validate the predictive performance of a CNN w.r.t. contrastive classes that cannot be distinguished easily solely based on concepts, but rather general, domain-relevant spatial relations. Combining relevance information with ILP has already been researched \cite{srinivasan2019logical}, but not for extracted concepts and not for intermediate layers of CNNs.
    \item For a collection of data sets (abstract, in-the-wild, scientific), we show quantitatively that our novel, combined explainer is faithful to a CNN model's predictive outcomes. We further extend the quantitative evaluation of fidelity by a concept-based ablation study, that examines the performance of a CNN, when the CNN is permitted to use relevant and / or irrelevant concepts. We provide a complete code basis to perform experiments presented here.
    \item A human evaluation of concept-based and relational explanations involving experts demonstrates the usefulness of combined explanations in comparison to only visual or verbal explanations for assessing a models classification performance.
    \item In order to facilitate the exploration and adaptation of a CNN model's decision boundary, we include contrastive explanations, rule-based cluster analysis and user-defined constraints into our approach.
\end{itemize}

Consequently, our main research question is, whether CoReX can identify concepts and relations from features learned by a CNN that are representative of contrastive classes and that are the main contributors to a model's predictive performance. We further examine the explanatory capabilities of CoReX toward rectifiable models, in particular, its suitability for interactive learning by constraints with a human-in-the-loop.

The paper is organized as follows: In Section \ref{sec:related} we present works that are related to our approach. Section \ref{sec:background} introduces the technical background as a basis to the implementation of CoReX and defines important terminology. CoReX itself is described in Section \ref{sec:method}. In Section \ref{sec:experiments} we introduce used materials and experimental settings in preparation of our evaluation, which is presented in Section \ref{sec:evaluation}, including quantitative and qualitative results. We conclude this article with future prospects in Section \ref{sec:conclusion}.

\section{Related Work}
\label{sec:related}

Our approach to concept- and relation-based explanations relates to methods for disentangling representations in intermediate layers of CNNs. Furthermore, 
we focus on, rule-based, contrastive explanations which support identifying classification errors near decision boundaries. This type of explanation is especially relevant for explainable interactive approaches of machine learning (XIML). Our approach targets a wide variety of users of different expertise, since the generated symbolic rules can be transformed into comprehensible explanations for experts and developers as well as novices \cite{finzel2024human}. See~\cite{cambria2023survey} and~\cite{gerlings2022explainable} for an extended discussion of fit between explanations and their audience.

\subsection{Concepts, Relations, and Disentangled Representations}

Research in human perception \cite{miller2013language} shows that the classification of an image depends not only on information about shape and texture but also on the constituents it is composed of and their relations. These constituents typically are concepts that can be named. Global explanations of categories typically rely on verbal or symbolic descriptions of their defining concepts and relations \cite{du2019techniques}. For instance, the concept of a grandparent can be explained as a person who is the parent of a parent \cite{muggleton2018ultra,rabold2022nearmisses} or the concept of a teapot could be explained as an item having a spout, a handle, and a lid where the lid is on top and the spout is on the side (see Figure~~\ref{fig:running_example}).
This observation that perception considers composition and relations has inspired approaches to 
disentangling representations.

One of the first approaches, NetDissect \cite{bau2017network}, considers a predefined pool of concepts given as objects, textures, and colors and finds units in convolution layers in a CNN whose activation maps are highly correlated with ground-truth masks of concepts in an image data set. The authors could confirm that representations at different layers disentangle different categories of meaning and that such disentanglements support interpretability of the representation learned by hidden units.
Net2Vec \cite{fong2018net2vec} is a modification of NetDissect, which learns a concept segmentation layer for locating concepts on images. The weight vector of this layer can then be understood as an embedding for the concept. Both methods rely on a predefined set of concepts and predefined masks which restricts explainability to the preconceptions of the model designers. 

A recent method that builds upon the relevance-based approach LRP \cite{lapuschkin2019lrp} is Concept Relevance Propagation (CRP) \cite{achtibat2022crp,achtibat2023from}. CRP provides interpretable and class-specific features as concepts that are identified through class-conditioned decomposition of relevance maps instead of activation. In contrast to NetDissect and Net2Vec, CRP is able to automatically find sets of pixels that represent interpretable features in terms of concepts. Instead of compa\-ring upscaled activation maps with ground-truth masks as in the previous approaches, CRP propagates class-specific relevance all the way to the input layer. We introduce the method in more detail in Section \ref{sec:method}. A survey on other related concept-based explanation methods can be found in a recent survey by Poeta et al. \cite{poeta2023concept}.

Approaches to relational symbolic learning, such as Inductive Logic Programming (ILP, \cite{muggleton1991ilp}) explicitly learn conceptual structures. ILP-learned models are sets of rules for a target concept defined over sub-concepts and their relations. The set of rules constitutes a global explanation of a concept. Local explanations are generated when the model is evaluated for a specific instance \cite{finzel2024human}. The reasoning trace generated for the instantiated rule constitutes the explanation \cite{finzel2024explaining}. ILP has been applied as an explanatory surrogate model for image classification with CNNs \cite{rabold2020expressive}. Furthermore, approaches to generating rule-based or contrastive explanations based on ILP models have been proposed \cite{rabold2022nearmisses,finzel2024explaining}. An extensive survey on these and related methods can be found in~\cite{schwalbe2023comprehensive}.

Some related post-hoc rule-generating explanation methods are introduced in the following paragraphs. Anchors~\cite{ribeiro2018anchors} are minimal sets of features, that, when present in a to be explained instance, altering other features does not change the classification of it by the decision system. LORE~\cite{guidotti2018local} and EXPLAN~\cite{rasouliY2020explan} generate decision trees of attribute checks that locally explain an instance of a tabular dataset. LORE additionally provides contrastive rules to help users identify, which attributes would need to be changed in order to receive a different classification. These described explanation methods however are not expressive enough for many problems, since their generated symbolic surrogates are purely propositional, whereas we want to generate explanations for relational domains.

\subsection{Explanations for Interactive Learning}

While measures of predictive accuracy give an overall indication of the quality of a learned model w.r.t. its performance on new inputs, explanations can provide a more specific understanding of the inner workings of a model. Global explanations reflect the general structure of a model. For relational data, ILP models constitute global explanations. ILP models can be learned stand-alone or in neuro-symbolic settings \cite{manhaeve2021neural}. For relevance maps generated with LRP, Spectral Relevance Analysis (SpRAy) has been proposed \cite{lapuschkin2019unmasking} as an approach for global explanation generation. SpRAy applies spectral clustering on LRP explanations and thereby identifies different decision behaviors of the learned model. Prototypes are another form of global explanations which highlight the typical pattern of examples which are classified as belonging to a specific class \cite{kim2016examples}. 

While global explanations help to understand the general structure of the model, local explanations allow for understanding the class decision of a model for a specific instance. Heatmaps are an instance of local explanations in the form of visualizations on input images. Contrastive explanations can be given globally -- by aligning two prototypes \cite{artelt2021efficient,miller2021contrastive} -- or locally, by contrasting the current instance with one which is similar but belonging to a different class \cite{herchenbach2022explaining,rabold2022nearmisses,finzel2024explaining}. Contrastive explanations have been identified as especially helpful to understand the underlying reasons of classification errors. Therefore, they are helpful as guidance for interactive approaches to machine learning. Explanatory Interactive Machine Learning (XIML) has been introduced as a term for approaches that combine explanation generation and interactive model correction based on such explanations \cite{teso2019explanatory}. Corrections are given by revising not only erroneous class decisions (i.e., labels) but also the explanations, thereby constraining model adaption \cite{teso2022overview,schramowski2020making}.

\section{Background and Terminology}\label{sec:background}

This section introduces the theoretical background of methods integral to our approach as well as the basic terminology relevant to this work.

\subsection{Extracting Visual Features with CRP}

The basis to CRP \cite{achtibat2022crp,achtibat2023from} is LRP \cite{lapuschkin2019lrp,montavon2019lrp}. LRP can be used to compute the contribution of individual pixels to a predictive outcome (relevance maps visualized as heatmaps). This is achieved by decomposing the relevance that is aggregated in an output neuron through a backward pass. We shortly introduce its basic computational rule:

\begin{equation}
\label{eqn:basicLrp}
  R_i^{(l)} = \sum_j \frac{a_i w_{ij}}{\Sigma_{i} a_i  w_{ij}} R_j^{(l+1)}
\end{equation}

\noindent
The relevance of a neuron $i$ in a layer $l$ can be determined as follows. Given the relevance $R_j$ of a neuron $j$ in a higher layer $(l+1)$ (e.g., the output neuron), the backward relevance flow from $j$ to $i$ is derived from dividing the product of the activation $a_i$ and the weight $w_{ij}$ by the sum of all activation weight products. If there is more than one neuron in $l+1$, the ratios get summed up by the outer sum $\sum_j$. Applying this rule in a complete backward pass leads to assigning relevance to every pixel $p$ of the input image. This score is computed in a conservative manner, meaning that $\sum_p R_p = f(x)$ holds for a classifier $f(x)$ on input $x$. The rule presented in Equation \ref{eqn:basicLrp} has variants for different layers \cite{montavon2019lrp}. Selecting the best variant for the given CNN architecture is crucial to the quality of relevance computation. Here, we apply the $\epsilon$-variant, which was developed for intermediate layers in CNNs. These layers are usually convolutional layers. Thus, the rule works well for our approach as we only consider relevance in the last convolutional layer. The benefit of the $\epsilon$ rule is, that it preserves the most salient relevance and reduces noise resulting from weight sharing in convolutions \cite{montavon2019lrp}. 

The relevance values of all pixels of an input constitute a relevance map. The CRP method we use to extract concepts is based on the assumption that a relevance map does not describe just one individual concept. Various filters contribute to the relevance that is distributed across pixels in an image \cite{achtibat2023from}. The advantage of CRP over LRP is that it can compute a class-conditional relevance map $R(x \mid y)$ for input $x$ w.r.t. a class $y$ and concept-conditional relevance maps by masking (intermediate) network outputs before applying LRP. It extends LRP as noted in Equation \ref{eqn:basicLrp}:

\begin{equation}
\label{eqn:basicCrp}
R_{i}^{(l)}(x\mid \theta \cup \theta_l) = \frac{a_iw_{ij}}{\sum_i{a_i w_{ij}}} * \delta_{jc_l} * R_j^{l+1}(x\mid \theta)
\end{equation}

Conditioning the relevance is achieved by setting all but the desired layer outputs to zero by multiplying a model output $f_j(x)$ with a Kronecker-Delta $\delta_{jy}$, such that $R_j^L(x \mid y) = \delta_{jy}f_j^L(x)$, where $L$ denotes an output layer. Here, $j$ is a convolutional layer represented by a tensor $(p,q,j)$, where $p,q$ denote the first and the second spatial axis of the tensor and $j$ the concept-axis, i.e., the filter. Conditioning the relevance by $\theta$ is possible for an individual network output for one or more filters in a convolutional layer. Achtibat et al. further assume that each filter encodes one particular concept throughout all model applications since their weights stay the same \cite{achtibat2022crp,achtibat2023from}. This is the basis for "filtering" specific concepts that were learned by the model.

The impact of a single concept $c$ on the prediction outcome of $x$ is represented by the sum of relevance in the respective layer (see Equation \ref{eqn:RelCrp}).
It helps to select the concepts contributing most to a prediction. Achtibat et al. use Relevance Maximization (RelMAx) to find for every concept a set of input samples (reference samples), where the concept contributed most to the prediction (see Equation \ref{eqn:relmax}).
    
\begin{equation}
\label{eqn:RelCrp}
  R^l(x \mid \theta_l) = \sum_{i} R_i^l(x \mid \theta_c)
\end{equation}

\begin{equation}
\label{eqn:relmax}
\tau_{max}^{rel}(x) = \max_iR_i(x \mid \theta).
\end{equation}

By using a concept-specific constraint $\theta$ on RelMax, a set of reference samples from the target class is selected. For this set, a human expert can decide upon the overall concept label for the filtered pixel regions in reference samples. We integrated this component into our implementation in order to optionally support labeling of concepts by human users (e.g., a teapot handle). Localization of a concept within a reference image is done by extracting the receptive field information \cite{achtibat2022crp,achtibat2023from}. The rest of the image is then masked out to emphasize on the region of interest, where the concept displays. As this work focuses on the technical evaluation of our proposed approach, we did not integrate the reference sampling into our experiments, however, we used it for validity checks on identified concepts.

Furthermore, reference sampling helps to label concepts more efficiently. Presenting reference samples to a (human) labeler according to the relevance rank of a concept helps to prioritize which concepts should be labeled first. In Section \ref{sec:method}, we will explain how we take advantage of concept ranking to learn an interpretable relational surrogate model on fewer data, yet faithful to a CNN model's predictions.

In advance to describing how we learn a surrogate model on top of extracted concepts and their relations, it is worth mentioning that CRP assigns concepts with positive or negative relevance, depending on whether they contribute to a decision in favor of the target class or not (likewise to LRP \cite{lapuschkin2019lrp}). A relevance value equal or near zero defines a low or missing contribution in any direction. To be able to interpret the explanations produced by our approach, we define the meaning of relevance such that if the relevance of some concept $c$ is positive, this concept is located in the image region for desirable reasons. If the relevance of some concept (not necessarily the same $c$) is negative, the concept should not be there, whichever concept is located at the relevant pixel area instead of the desired one.
We want to point out that images cannot contain nothing. If a concept of interest is not displayed, there always will be another concept, pixel group or pixel value with possibly no defined meaning instead.

\subsection{Learning Symbolic Hypotheses with ILP}

ILP is a machine learning approach that allows for induction and deduction. It has been introduced back in the year 1991 by Stephen Muggleton \cite{muggleton1991ilp}. Learned models are logic programs, which are induced from examples. These logic programs can then be executed to derive explanations for classifications \cite{schwalbe2023comprehensive}. The overall goal of ILP is to derive a hypothesis $H$, also called theory or model, from a set of examples $E^+$ belonging to the target class and a set of examples $E^-$ not belonging to the target class and background knowledge $B$. The background knowledge $B$ holds features of examples and optionally reasoning rules (called domain theory) to derive further properties, e.g., transitive relations. $H$ is induced such that $\forall e \in E^+ : B \cup H \models e $ and $\forall e \in E^- : B \cup H \not\models e $. The first condition demands $H$ to be complete, i.e., covering all positive examples $E^+$. The second condition demands it to be consistent, i.e., not covering negative examples $E^-$.

\begin{algorithm}[ht!]
    \caption{Basic ILP algorithm}
    \label{fig:algo1}
    \textbf{Input:} $(B,E^+,E^-)$: an ILP problem\\
    \textbf{Output:} $M$: a classification model
    \begin{algorithmic}[1] 
        \State $M \gets \emptyset$ \Comment{initialise the model}
        \State $C \gets \emptyset$ \Comment{temporary clause} 
        \State $Pos \gets E^{+}$ \Comment{the set of positive training examples}
        \While{$Pos \neq \emptyset$}
            \State $C \gets \text{GenerateNewClause}(Pos, E^{-}, B)$
            \textbf{such that} $\exists p \in E^+\!:\ B \cup C \models p$ \State \textbf{and} $\forall n \in E^-\!:\ B \cup C \not\models n$ \textbf{and} $C$ is optimal w.r.t. quality criterion
            \State $A(C)$
	\State $Pos \gets Pos \setminus \{p\}$
	\State $M \leftarrow M \cup \{C\}$
        \EndWhile
        \State \textbf{Return} $M$
    \end{algorithmic}
\end{algorithm}

\noindent
In particular, it can be seen from Algorithm~\ref{fig:algo1} that ILP learns in a top-down approach a set of clauses $C$ which covers the positive examples while not covering the negative ones.
In the subsequent paragraphs we will refer to the learned hypothesis as the model $M$, which is induced by iteratively generating new clauses $C$. In Figure \ref{fig:algo1} we summarize this step using the function
$GenerateNewClause$, according to~\cite{gromowski2020process}. The ILP problem introduced above is then defined by learning from examples $E^+$ and $E{^-}$, given the background knowledge $B$,
such that for each $C$ there exists a positive
example $p$ that is entailed by $C \cup B$, i.e., $C \cup B \models p$,
while no negative example $n$ is entailed by $C \cup B$.
Furthermore, it is of interest to find clauses such that each $C$ is optimal with respect to some \textit{quality criterion} $A(C)$, such as the total number of entailed positive examples \cite{finzel2021explanation}.
Such \emph{best} clauses are added to model $M$.

The formal properties of consistency and completeness make ILP susceptible to noise in the underlying data. This apparent downside comes in favor for generating explanations for flawed models since ILP will not be robust to examples incorrectly labeled by some base model, the explanandum. When instances contain noisy concepts or relations which the model deems to be important, hypotheses generated by ILP can make the noise explicit to the user which in turn gives them the power to evaluate the model as well as the data.

For theory induction, this work mainly uses the Prolog-based ILP framework Aleph~\cite{srinivasan2007aleph}. The basic procedure that is performed by Aleph follows four steps. First, for as long as there is an example in $E^+$, a candidate example $e \in E^+$ gets selected. In the second step, a most specific clause (so-called bottom clause) is constructed, which entails the selected example and adheres to $B$. Then, Aleph searches for a rule that is more general than the bottom clause, i.e., a subset of all literals in the current clause. Aleph tests which examples from $E^+$ or $E^-$ are covered by the rule and finally removes all (now redundant) examples from the list of candidate examples, that are already covered. This procedure is repeated until all examples are covered, either by a rule or added to the theory as instances without generalization \cite{srinivasan2007aleph}.

An alternative, recently developed ILP approach is Hierarchical Probabilistic Logic Programming (HPLP) that allows for learning both the parameters (probabilities) and the structure (logic rules) from data. HPLP uses an algorithm for parameter learning that initially converts the HPLP input into a collection of arithmetic circuits that share parameters. Subsequently, it employs gradient descent or expectation maximization on these arithmetic circuits \cite{fadja2021learning}.

Another ILP system, Popper~\cite{cropper2021learning}, generates a possible clause $h \in H$, where $H$ is the problem space within given language constraints. In a testing stage, the number of positive and negative examples $E^+$ and $E^-$ covered by $h$ is evaluated. If a clause $h$ covers all of $E^+$ and none of $E^-$, $h$ is returned as a solution of the given problem. If there are generated clauses where there is at least one positive covered example and no negative one, a SAT solver searches for combinations of these clauses, where, in the best case, all of $E^+$ are covered. On the other hand, if there are clauses, where at least one positive example is covered and also some negative examples are covered, the SAT solver tries to find clauses that combine the body literals, such that by these more specific clauses, no $e \in E^-$ is covered. These clauses can then be used for the combine stage as well. Whenever Popper did not yet find the optimal solution, a constraint stage adds more constraints to the search space, such as preventing specifications of clauses, that do not cover any positive example. Popper is also capable of handling noisy data by weakening the optimality constraint and using the minimum description length as a metric for evaluating candidate clauses, which penalizes long clauses that would render many false positives and false negatives.

\section{Method}
\label{sec:method}

\begin{figure}
	\centering
	\includegraphics[width=\textwidth]{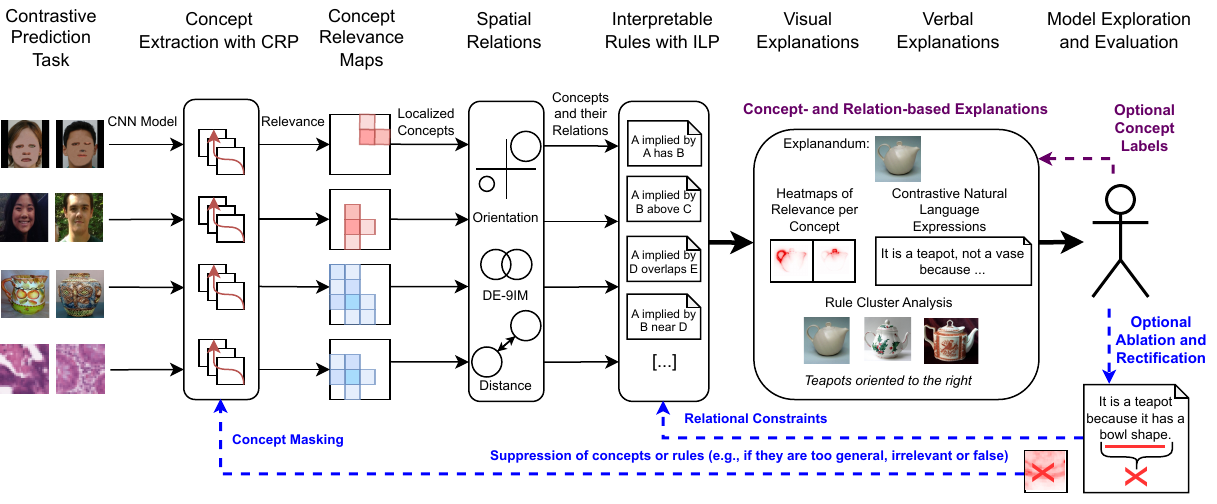}
	\caption{Overview of our CoReX approach for explaining and evaluating CNN image classifications with concept- and relation-based explanations and constraints (concept masking and relational constraints).}
	\label{fig:overview}
\end{figure}

Our approach is illustrated in Figure~\ref{fig:overview}. It combines (1) extracting concepts from visual features learned by CNNs with CRP, and (2) learning symbolic hypotheses by means of relations with ILP, where we integrate spatial relational knowledge for a more expressive evaluation of the CNN output. The visual and verbal explanations resulting from (1) and (2) are the basis for providing (3) contrastive explanations, and (4) rule-based cluster analysis for a given explanandum. To evaluate the importance of the concepts and relations that are learned by the surrogate ILP model, we then suppress concepts in learned CNN models by (5) masking the respective concepts before application on the same data, and we suppress relations by (6) constraining relations in the ILP model to identify and re-classify samples based on human domain-knowledge.

\subsection{Building Background Knowledge from Concepts}

In Section \ref{sec:background} we introduced how concepts are extracted from the relevance of filters in convolutional layers of CNN models. In principle, we only analyze the last convolutional layer of our utilized models for concept extraction. More details on the architectures of our models and their parameters are given in Section \ref{sec:experiments}. For the preparation of learning a surrogate model with ILP, we store concepts in the background knowledge $B$ as introduced earlier. That means, we write a first-order-logic 2-ary predicate to the Prolog background knowledge for each example image if a concept is present (see first line of part D in Figure \ref{fig:ilp_process}).

Depending on the architecture of the respective CNN model, the amount of extracted concepts can be large ($>$ 500) as noted in Section \ref{sec:number} of the appendix. One observation we make is that the majority of these concepts often has low relevance values. This does not necessarily imply that these concepts are useless, however, it is probable that many of them can be removed without seriously harming the performance of a CNN model. For improved computational performance, we therefore decided to write only the concepts to the background knowledge that have a relevance, which exceeds a predefined relevance threshold. Since every model may produce a different distribution of relevance dependent on the data set under consideration, we set the threshold according to a quantile of the relevance distribution. This can result in varying amounts of concepts in the background knowledge accross data sets, however, this strategy is more flexible and less biased compared to a fixed amount of concepts or a fixed value for a threshold.

In our experimental results in Section \ref{sec:evaluation}, we speak of three different kinds of concepts. Based on the thresholding, we distinguish between relevant concepts (in the background knowledge) and irrelevant concepts (not in the background knowledge). Within the concepts that are in the background knowledge, we further introduce another category of concepts: the concepts from the background knowledge that are included in the rules of the interpretable surrogate model after learning a theory with ILP.

Besides storing predicates for concepts that exceed the relevance threshold, we also compute spatial relations between these concepts.

\subsection{Integration of Spatial Relational Knowledge}

The Prolog background knowledge that basically consists of concepts with positive or negative relevance, can be additionally enriched by knowledge that is generally applicable to the image data domain, such as spatial relations. The relations are labeled according to existing spatial frameworks. In particular, we integrate a symbolic representation of relations from the DE-9 Intersection Model \cite{clementini1996DE9im}, orientation in a 8-cell grid and categorized distance. For each framework, we first localize the pre-computed concepts with the help of concept relevance, we then transform the conjunct pixel regions into polygons and relate the polygons in accordance to the respective spatial framework. We then store the labeled relations in Prolog syntax as 3-ary predicates with 2-ary predicates as arguments for denoting the relevance sign of a concept (see bottom lines of part D in Figure \ref{fig:ilp_process}). An overview of all possible predicates is presented in Section \ref{sec:spatial} of the appendix.

\subsection{Model Truth and Explainer Truth}\label{sec:mtxt}

A useful interpretable surrogate model should produce outcomes similar to the machine learning model in question as a precondition to providing explanations faithfully. In particular, observing the output of the base model and the output of the interpretable explainer gives insight in how closely an explainer resembles the model's predictive behavior. To compare the two, we use the terms \emph{model truth} $\hat{y}_d$ and \emph{explainer truth} $\overset{e}{y}_d$ for a given sample $d$. These terms manifest what the model or the explainer deem to be true for $d$. Both of these give either 1 or 0 depending on whether $d$ was being rendered as belonging to a target class or not. We use this notation and terminology for evaluating the fidelity of an explainer (see Subsection~\ref{sec:eval}).

Having introduced concept extraction, rule induction on concepts, spatial relations between them and evaluation terminology, we put all components together that form our approach and introduce its algorithmic foundation in the next section.

\subsection{Algorithm}

We summarize the main procedure of CoReX as presented in Algorithm~\ref{alg:mainAlg}.
For all positive and negative model truth examples, background knowledge $B$ for ILP is generated. For one example $e$, concept relevance maps $C_l$ are found by CRP on all selected layers $l \in L$. The most important concepts exceeding the relevance threshold $\tau$ are used to localize polygons $P_e$ fitted on the relevance maps. Utilizing spatial calculi, relations between the polygons are then found and stored in symbolic form. This information is then used to induce a logic theory $H$ in form of rules by Aleph. Optionally, it is possible to provide the CRP procedure with a parameter $\varphi$ for masking concepts. Furthermore, in ILP it is possible to add a set of constraints $\phi$ on concepts and relations that must not appear in learned theories.

\begin{algorithm}[tb]
    \caption{CoReX Algorithm}
    \label{alg:mainAlg}
    \textbf{Input}: Original model $f$, positive/negative model truth samples $E^+$/$E^-$, set of probed layers $L$, optional masking $\varphi$, relevance threshold $\tau$, set of possible relations $S$, optional ILP constraints $\phi$
    \begin{algorithmic}[1] 
        \State $B \gets \text{domain theory}$
        \ForAll{$\odot \in \{+, -\}$}
            \ForAll{$e \in E^\odot$}
                \State $P_e \gets \emptyset$ 
                \ForAll{$l \in L$}
                    \State $C_l \gets \text{crp}(e, f, l, \varphi)$
                    \State $C^*_l \gets \text{filter\_concepts}(e, l, q)$
                    \ForAll{$c \in C^*_l$}
                        \State $P_e \gets P_e \cup \text{localize}(c, \tau)$
                    \EndFor
                \EndFor
                \State $R_e \gets \text{find\_relations}(P_e, S)$
                \State $B \gets B \cup R_e$
            \EndFor
        \EndFor
        \State $H \gets \text{induce}(E^+, E^-, B, \phi)$
        \State \textbf{return} H
    \end{algorithmic}
\end{algorithm}

An exemplary rule for our running example of separating teapots from vases is presented below. It expresses that some sample A belongs to the target class (here: teapot) \textit{if} there are two concepts in A (concept ``spout'' and concept ``handle''), both concepts contributing with positive relevance (``pos''), and it holds that the ``spout is located right of the handle'' (as presented in Figure~\ref{fig:running_example}).\\

\begin{centering}
    \begin{verbatim}
    is_class(A) :-
        right_of(A, pos(A, spout), pos(A, handle)).
    \end{verbatim}
\end{centering}

\begin{figure}[bt!]
    \centering
	\includegraphics[width=\textwidth]{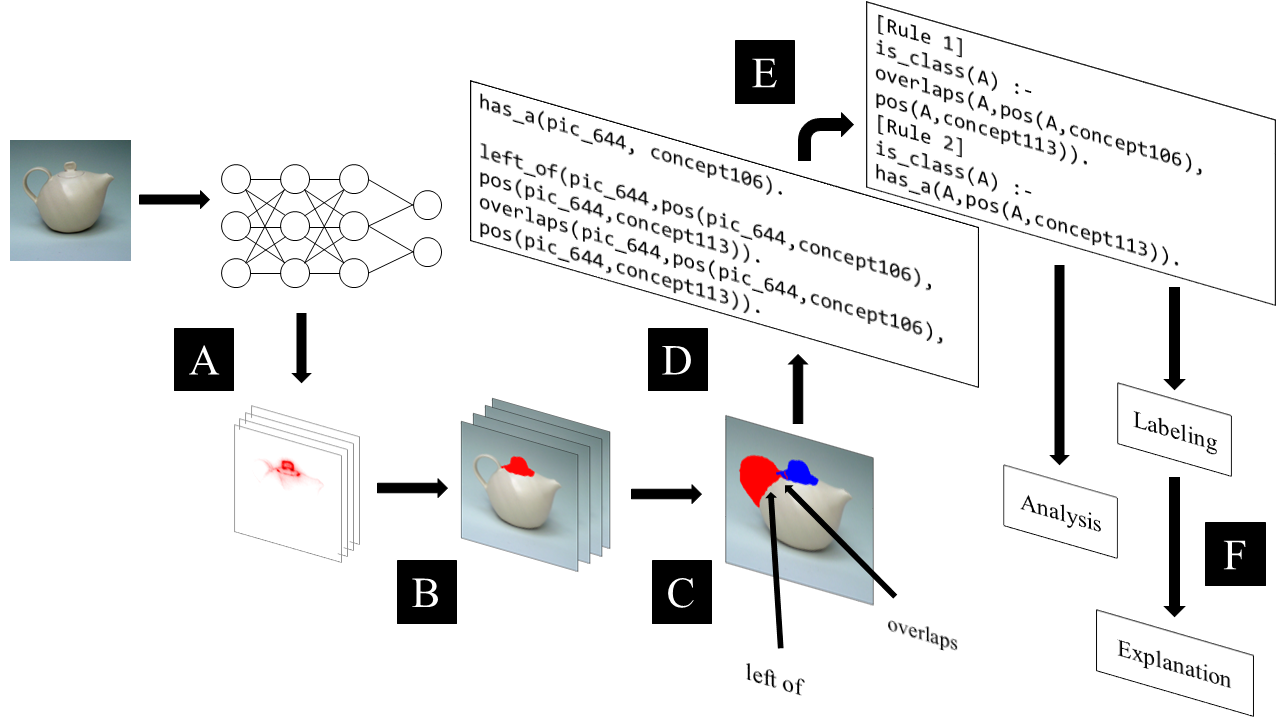}
	\caption{Overview of the process of generating explanations with ILP from extracted concepts and learned relations}
    \label{fig:ilp_process}
\end{figure}

The rule results from the process presented in Figure~\ref{fig:ilp_process}. For each $e \in E^+$ and $e \in E^-$ we get the classification of $e$ and the CRP-based concepts, conditioned by the target (see step A in Figure~\ref{fig:ilp_process}). We then localize and polygonize the extracted concepts (step B). Then, spatial relations between polygons are computed (step C) and the resulting relations are transferred into Prolog syntax as input to ILP (step D). A theory is induced (step E) and finally, evaluation analysis or explanations take place (steps in F). For explanations, the Prolog syntax is translated into natural language based on a fixed scheme.

\section{Experiments}\label{sec:experiments}

This section presents the materials, metrics and the experimental setting of our evaluation.

\subsection{Models}\label{sec:models}

For our experiments, we use two CNN architectures well established, especially for LRP-based explanation methods, namely VGG16 \cite{SimonyanZ14a} and ResNet50 \cite{he2016deep}. For both architectures exist optimized variants of the $\epsilon$ rule for the computation of concept relevances \cite{achtibat2022crp,achtibat2023from}, which is another favorable property of the chosen architectures.

For the classification of most of our selected data sets, we use VGG16 models pre-trained on the ImageNet database \cite{deng2009imagenet} and fine-tune the fully connected layers on our selected data sets. Table~\ref{tab:models} states the performance metrics and hyperparameters we use for the fine-tuning on our selected data sets. This way, we preserve the features learned in the network, while updating the weights. The rationale behind not fine-tuning features for our specific classification tasks is that we assume that the pre-trained features are already general purpose for a variety of tasks. By fine-tuning the fully connected layers, the relevance values for CRP, which is calculated down-stream dependent on weights, become task-specific and thus indicate relevance of features for a specific classification problem. One particular data set (see descriptions below) is classified based on a pre-trained model with a ResNet50 architecture. For this network, we fine-tuned the last layers, in particular the average pooling layer and the linear output layer. This was necessary, since we had to make adaptations to the composition of classes. More details are given in the paragraphs on the chosen data sets.

\begin{table*}[bt!]
\centering
\caption{The characteristics of the non-contrastive pre-trained models used for contrastive classification after fine-tuning (CE is cross entropy; BCE is its binary equivalent)}
\label{tab:models}
\resizebox{\textwidth}{!}{%
\begin{tabular}{lllllllllll}
\toprule
 & \#Train & \#Test & Train F1 & Test F1 & \#Class & Batch Size & Max. Epochs & Optimizer & Loss Function & Learning Rate \\
 \midrule
Picasso (VGG16) & 18,002 & 1998 & 0.9933 & 0.9924 & 1 & 32 & 20 & Adam & BCE & 0.0001  \\
Adience (VGG16)    & 9942 & 2252 & 0.9913 & 0.8702 & 2 & 32 & 20 & Adam & CE & 0.0001  \\
Teapot and Vase (VGG16)   & 231 & 100 & 1.0000 & 0.9200 & 2 & 32 & 20 & Adam & CE & 0.0001  \\
PathMNIST (ResNet50) & 25,765 & 2462 & 0.9974 & 0.9709 & 2 & 128 & 10 & Adam & CE & 0.001  \\
\bottomrule
\end{tabular}%
}
\end{table*}

For all our models, we only examine features in the very last convolution layer of the architecture, although the CRP method would allow for inspection of preceding layers as well. This is of particular interest, when concepts may be organized in a hierarchy. This would, however, require the consideration of the relevance flow between different layers, for which there exists no thoroughly evaluated method for CNNs so far.

\subsection{Data Sets}\label{sec:datasets}

\begin{figure}[tb]
    \centering
    \captionsetup{justification=centering}
    
    \begin{subfigure}[b]{0.4\columnwidth}
        \centering
        \includegraphics[width=\columnwidth]{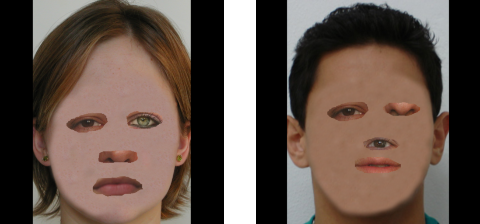}
        \caption{Picasso: correct (left) vs incorrect (right) faces}
    \end{subfigure}
    \hfill
    \begin{subfigure}[b]{0.45\columnwidth}
        \centering
        \includegraphics[width=\columnwidth]{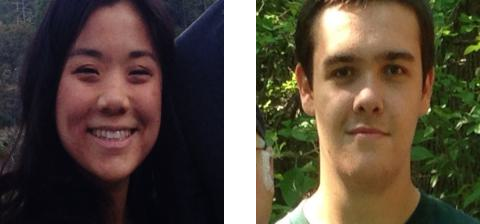}
        \caption{Adience: female (left) vs male (right)}
    \end{subfigure}

    \begin{subfigure}[b]{0.45\columnwidth}
        \centering
        \includegraphics[width=\columnwidth]{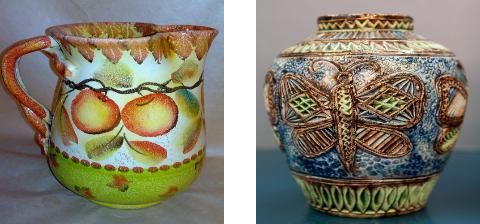}
        \caption{ImageNet: teapots (left) vs vases (right)}
    \end{subfigure}
    \hfill
    \begin{subfigure}[b]{0.4\columnwidth}
        \centering
        \includegraphics[width=\columnwidth]{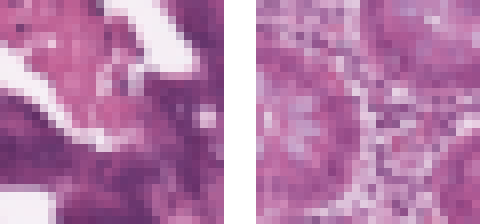}
        \caption{PathMNIST: cancerous (left) vs healthy (right) tissue}
    \end{subfigure}
    
    \caption{Examples of contrastive classes}
    \label{fig:example_images}
\end{figure}

The data sets we have chosen cover a range of characteristics, which makes them suitable for a comprehensive evaluation of our approach. We use artificially generated image data, benchmark image data, showing either static and aligned objects or in-the-wild recorded scenes, and a benchmark that contains scientifically and application relevant samples, e.g., for medical diagnosis. All data sets are thoroughly chosen w.r.t. the criterion that they contain classes, which are very likely to share concepts, while not sharing the same spatial relationships between those concepts. Figure~\ref{fig:example_images} shows example images from the data sets.

\paragraph{Picasso: Correct Faces versus Incorrect Faces.}
The Picasso data set~\cite{rabold2020expressive} is a constructed data set derived from the FASSEG set~\cite{khan2015multi} for segmentation of facial features in frontal faces. Picasso contains 224x224 images where the positive class holds faces with three features, the eyes, nose and mouth, in natural position, whereas in the negative class, the features are swapped randomly. For ILP, we use 250 each from train and test data. The reason, why we evaluate our approach on training data and test data is that explanations may differ depending on the kind of data. Explanations on training data evaluate which representations the model has learned, while explanations on test data evaluate how well these explanations generalize.

\paragraph{Adience: Female versus Male Persons.}
The Adience data set~\cite{eidinger2014age} features 224x224 cropped face images labeled by gender. The images were taken in the wild and contain a variety of backgrounds and illumination conditions. We sampled 200 male and 200 female labeled images. We chose the data set as we expected different spatial relations for (probably biased) properties of persons, like the localization of hair due to its length.

\paragraph{ImageNet: Teapots versus Vases.}
We took 161 images from teapots and 170 images from vases (224x224) from the ImageNet data set~\cite{deng2009imagenet}. Because labels were missing, we let the original VGG16 classify these images. The same images were taken as input to generate our explanations. Due to the small size of the data set, we could only use training data as well as test data for the VGG16 evaluation. For our ILP model, we only considered training data for evaluation.

\paragraph{PathMNIST: Healthy versus Tumorous Tissue.}
The PathMNIST data set \cite{kather2019predicting}, subset of MedMNIST \cite{yang2021medmnist}, consists of 28x28 pathological histology slides taken from healthy as well as cancerous colon tissue. PathMNIST originally is a 9 class classification task. For our experiments, we re-sampled the images to form a 2-class task by keeping all images in the cancer-associated class as positive images (12,885) and then performed a stratified sampling over the other 8 healthy classes to receive approximately as many images (12,880) in the contrastive class. We sampled 250 positive and 246 negative images for ILP. 

\subsection{Evaluation Metrics}\label{sec:eval}

When evaluating the generalization power of our models, we use the standard F1 metric (harmonic mean of precision and recall). The respective scores for the ground truth labels of the train and test sets can be found in Table~\ref{tab:models}.

For the evaluation of fidelity of an explanation model w.r.t. an original black box model, we compare the model truth (output of the original model) with the explainer truth (output of the explanation model) which were introduced in Subsection~\ref{sec:mtxt}.
Adopting the method from~\cite{rabold2020expressive}, in accordance with the notion of fidelity introduced in~\cite{guidotti2018survey}~and~\cite{carvalho2019machine}, we calculate the accuracy of the explainer truth $\overset{e}{y}_d$ w.r.t. the model truth $\hat{y}_d$ when feeding instances $d$ from a data set $D$, where $\delta(a,b)$ is 1, if a and b are equal and 0 otherwise (Equation~\ref{eq:acc}). For better readability, the index of the samples is omitted.

\begin{equation}
    \frac{\sum_{d \in D} \delta(\hat{y}_d, \overset{e}{y}_d)}{\vert D\vert}
    \label{eq:acc}
\end{equation}

\subsection{Ablation Study}
We combine model explanations with ablation studies to evaluate the used CNN models (see section \ref{sec:results} for more details). In machine learning, ablation studies refer to the removal of architectural components or model features. The goal is to evaluate a model w.r.t. its predictive performance and the contribution of architectural components or features to the predictive outcome. Approaches for CNN ablation are, e.g., pruning, drop out or suppression of activation and weight flows between neurons \cite{choi2022neurons}.

To perform an ablation study based on our concept- and relation-based explainer CoReX, we apply masking on the filters of the CNN to suppress concepts. This technique is realized in a first step by setting all filters in the CNN to zero that do not match with concepts appearing in the background knowledge of ILP (masking irrelevant concepts). We measure the predictive performance of the adapted CNN. We then mask all irrelevant concepts as well as the concepts that appear in a theory learned by ILP (masking rule concepts). We hypothesize that, if the performance drop is larger for the second case compared to the first one, the concepts learned by the CNN and provided to ILP have an decreasing effect on the confidence of the CNN in accordance to their relevance. For our experiments we applied Aleph with default settings (see \cite{srinivasan2007aleph}). 

\section{Evaluation}\label{sec:evaluation}

This section presents the main results of our experiments. We include an analysis of the fidelity of our interpretable, relational surrogate model as it serves as a basis for generating explanations. We evaluate the fidelity in terms of a match in predictive outcomes (same classification) as well as in terms of predictive performance of the CNN when masking concepts that occur in the interpretable model generated by the surrogate model. An additional qualitative evaluation of explanations illustrated based on example explanations and demonstrated by a systematic human evaluation conducted with experts, complements the findings. We further examine the explanatory applicability of CoReX, in particular for contrastive explanations and cluster analysis. We further discuss the role of constraints in our approach, which extends beyond the masking and constraining in our ablation study.

\subsection{Fidelity and Predictive Performance}
\label{sec:results}

Table~\ref{tab:model_truth} presents the correct and incorrect predictions for all CNN models on training and test data, which form the input for the ILP-based explainer. The table denotes the amount of positive examples ($E^+$) and negative examples ($E^-$), accordingly. The column ``\# Rule Concepts'' documents the amount of concepts that were included by ILP into rules.

Our results indicate that our explanations (generated for the samples stated in Section~\ref{sec:datasets}) have a high fidelity w.r.t. the original model when measured by the comparison of the model truth and the explainer truth (as calculated by the fidelity Equation~\ref{eq:acc}). Table~\ref{tab:masking} then states the fidelity of the ILP explanations for the training and test samples.

We additionally measured the F1 scores for models that were altered by masking during the performed ablation study. We differentiate two cases: (1) masking concepts that occurred in the learned ILP theory as well as irrelevant concepts versus (2) masking only concepts deemed irrelevant because they are not occurring in the ILP background knowledge. We expected to see a drop in performance when masking concepts occurring in rules. Table~\ref{tab:masking} shows that in almost all scenarios, the performance is reduced (in bold: F1 score of ``Rule + Non-BK-Masking'' is lower than the F1 score of ``Non-BK-Masking''). This further may be an indicator that the learned rules constitute explanations of the original model's prediction in terms of used concepts. The magnitude of the performance drop appears to be rather low, however, related to the amount of masked concepts and the amount of concepts in total (see Section \ref{sec:number}) this drop is not insignificant. Thus, we can observe that the concepts from the background knowledge, which have a high relevance due to the relevance threshold, but are not included in ILP rules cannot fully compensate for the impact of the concepts in learned rules. This may not hold for the Teapot and Vase experiments on the training set as well as the PathMNIST test set, since the performance is equal in both of the masking settings. Thus, the results do not perfectly support our hypothesis of dropping performance upon masking concepts from learned rules, however, they highly support it. Moreover, there was not a single case in which the performance of the network improved when concepts were masked, which is a desirable outcome for relevant concepts. We further examined at which rank in the relevance ranking the concepts from rules with the largest coverage appeared across all samples. The frequency of occurrence per rank is indicated in Section \ref{sec:rankings}. We can see that the top 10 most frequent ranks per concept are usually the first ranks or at least upper 20 \% top ranks of possible ranks. In few cases, higher rank numbers occur in the 10 most frequent ranks. This can be explained by \emph{auxiliary} concepts that occur as part of a relation together with more important concepts, indicating that not only single concept occurrences are important but also the inter-relations between them. In summary, the findings support the claim that not only the calculated relevance can be transferred to the importance of concepts, but also the spatial constellation of concepts as learned by the ILP model contributes to the performance of an image classifier.

\begin{table*}[bt!]
\centering
\caption{Model truth for the train/test data (seen/not yet seen by the model) and the number of concepts occurring in the learned rules}
\label{tab:model_truth}
\resizebox{\textwidth}{!}{
\begin{tabular}{llllllll}
\toprule
\textbf{Experiment} & True Positives & True Negatives & False Positives & False Negatives & \#Rule Concepts & ILP Input ($E^+$) & ILP Input ($E^-$) \\
\midrule
Picasso-Train (VGG16) & 250 & 243 & 7 & 0 & 44 & 257 & 243 \\
Adience-Train-FM (VGG16) & 197 & 195 & 5 & 3 & 10 & 202 & 198 \\
Adience-Train-MF (VGG16) & 195 & 197 & 3 & 5 & 9 & 198 & 202\\
Teapot-Vase-Train (VGG16) & 157 & 166 & 4 & 4 & 10 & 161 & 170 \\
Vase-Teapot-Train (VGG16) & 166 & 157 & 4 & 4 & 13 & 170 & 161 \\
PathMNIST-Train (ResNet50) & 250 & 246 & 0 & 0 & 6 & 250 & 246 \\

Picasso-Test (VGG16) & 250 & 250 & 0 & 0 & 31 & 250 & 250 \\
Adience-Test-FM (VGG16) & 179 & 164 & 36 & 21 & 10 & 215 & 185 \\
Adience-Test-MF (VGG16) & 164 & 179 & 21 & 36 & 6 & 185 & 215 \\
PathMNIST-Test (ResNet50) & 238 & 239 & 7 & 12 & 7 & 245 & 251 \\
\bottomrule
\end{tabular}%
}
\end{table*}

\begin{table*}[bt!]
\centering
\caption{Explainer fidelity and train/test data metrics for the networks after masking of rule + non-background-knowledge (BK) concepts compared to only masking non-background-knowledge concepts. Bold values indicate an expected drop in performance}
\label{tab:masking}
\resizebox{\textwidth}{!}{
\begin{tabular}{llllll}
\toprule
& & \multicolumn{2}{l}{\textbf{Rule + Non-BK-Masking}} & \multicolumn{2}{l}{\textbf{Non-BK-Masking}} \\
\midrule
\textbf{Experiment} & Explainer Fidelity & Amount of Masked Concepts & F1 score & Amount of Masked Concepts & F1 score \\
\midrule
Picasso-Train (VGG16) & 0.9860 & 208 & \textbf{0.2985} & 164 & 0.9921 \\
Adience-Train-FM (VGG16) & 1.0000 & 18 & \textbf{0.9437} & 8 & 0.9913 \\
Adience-Train-MF (VGG16) & 1.0000 & 17 & \textbf{0.9896} & 8 & 0.9913 \\
Teapot-Vase-Train (VGG16) & 0.9970 & 24 & 1.0000 & 14 & 1.0000 \\
Vase-Teapot-Train (VGG16) & 0.9970 & 27 & 1.0000 & 14 & 1.0000 \\
PathMNIST-Train (ResNet50) & 1.0000 & 1384 & \textbf{0.9872} & 1378 & 0.9873 \\

Picasso-Test (VGG16) & 0.9980 & 193 & \textbf{0.6689} & 162 & 0.9924 \\
Adience-Test-FM (VGG16) & 0.9975 & 15 & \textbf{0.8079} & 5 & 0.8708 \\
Adience-Test-MF (VGG16) & 0.9975 & 11 & \textbf{0.8673} & 5 & 0.8708 \\
PathMNIST-Test (ResNet50) & 0.9980 & 1374 & 0.9367 & 1367 & 0.9367 \\
\bottomrule
\end{tabular}%
}
\end{table*}

Complementary results obtained from experiments with Popper in comparison to Aleph can be found in Appendix \ref{tab:fidelity_popper}.

\subsection{Evaluation of Explanations}

The following subsections illustrate the usefulness of explanations generated by CoReX.

\subsubsection{Example Explanations}

\begin{figure}[bt!]
    \centering
        \begin{subfigure}[b]{0.45\textwidth}
        \centering
        \footnotesize
        Aleph:\\
        \emph{Teapot, if positive concept c30 (spout) is right of positive concept c9 (handle).}\\
        \vspace{.6cm}
        Popper:\\
        \emph{Teapot, if positive concept c30 (spout) is contained or positive concept c398 (lid) is contained.}
        \vspace{.6cm}
    \end{subfigure}
    \begin{subfigure}[b]{0.45\textwidth}
    \begin{subfigure}[b]{\textwidth}
        \frame{\includegraphics[width=.32\columnwidth]{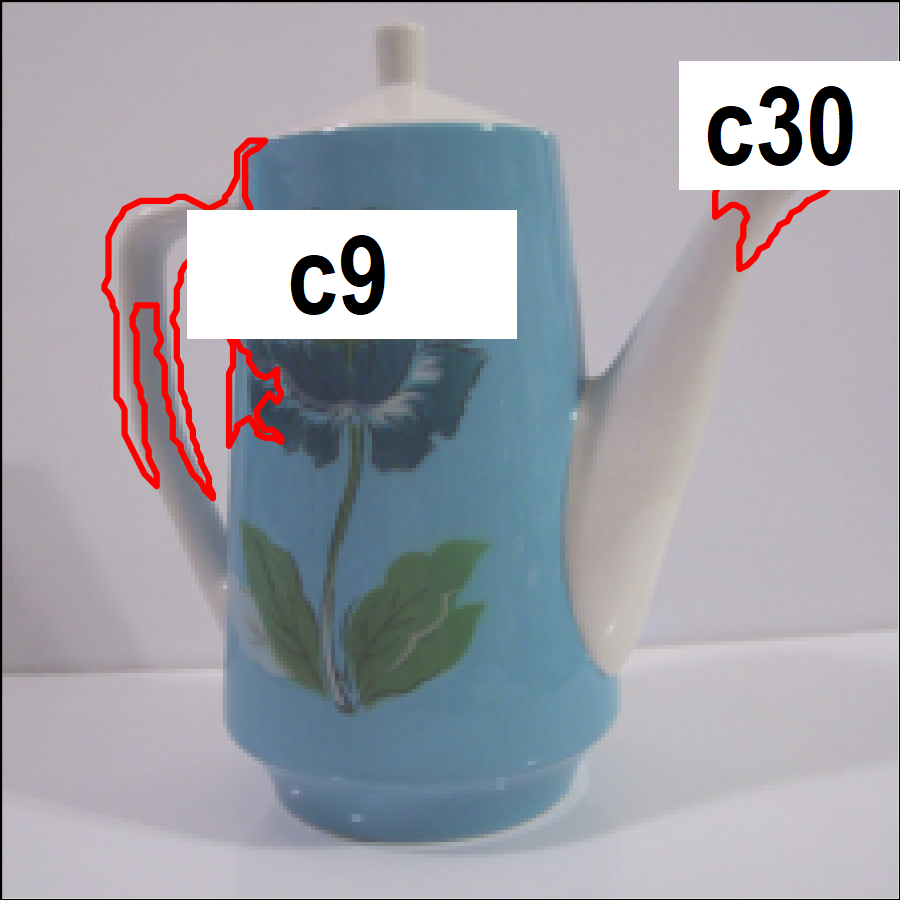}}
        \frame{\includegraphics[width=.32\columnwidth]{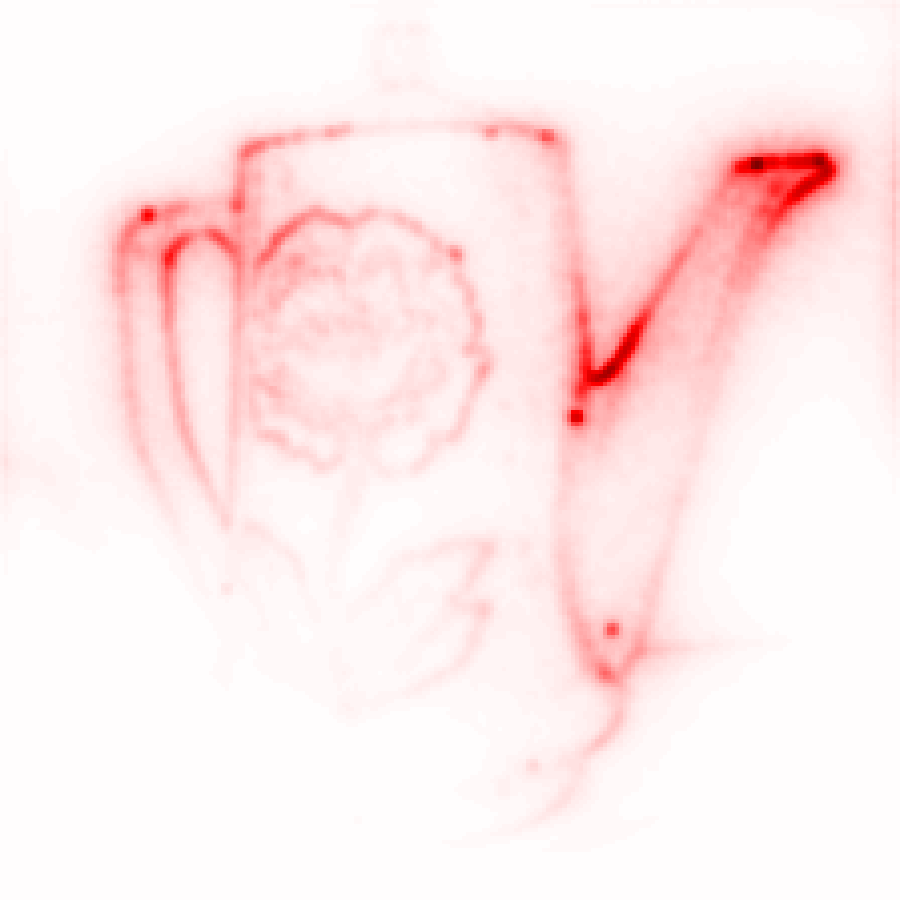}}
        \frame{\includegraphics[width=.32\columnwidth]{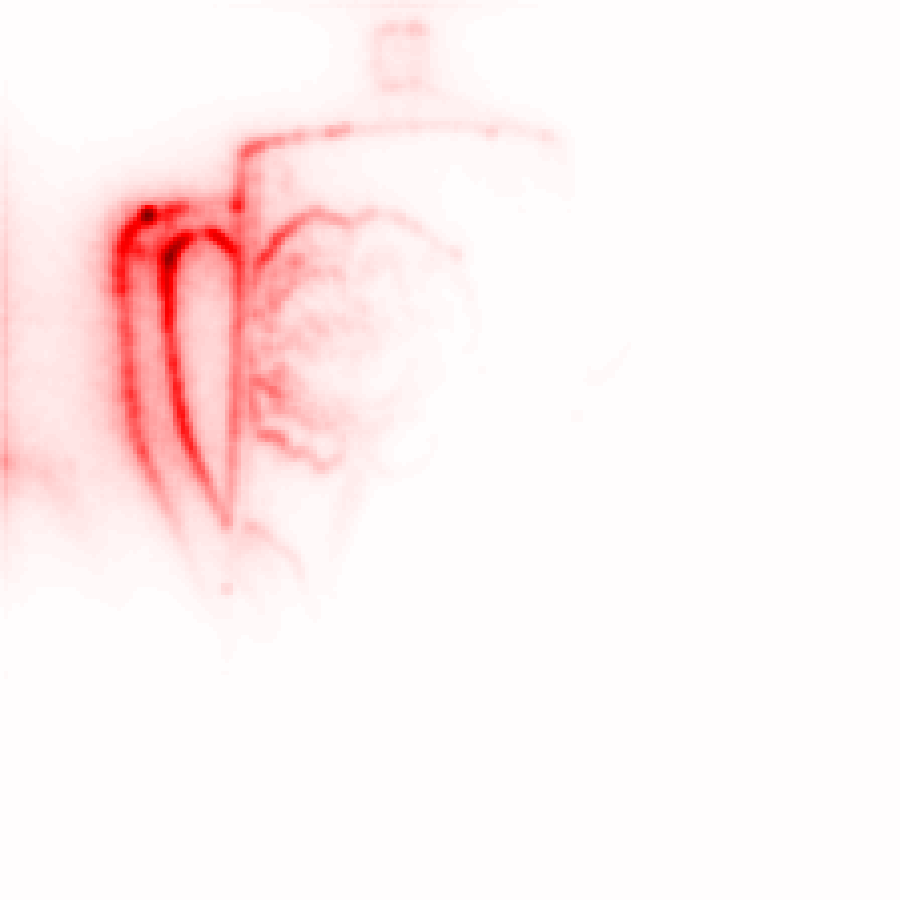}}
    \end{subfigure}
    \begin{subfigure}[b]{\textwidth}
        \frame{\includegraphics[width=.32\columnwidth]{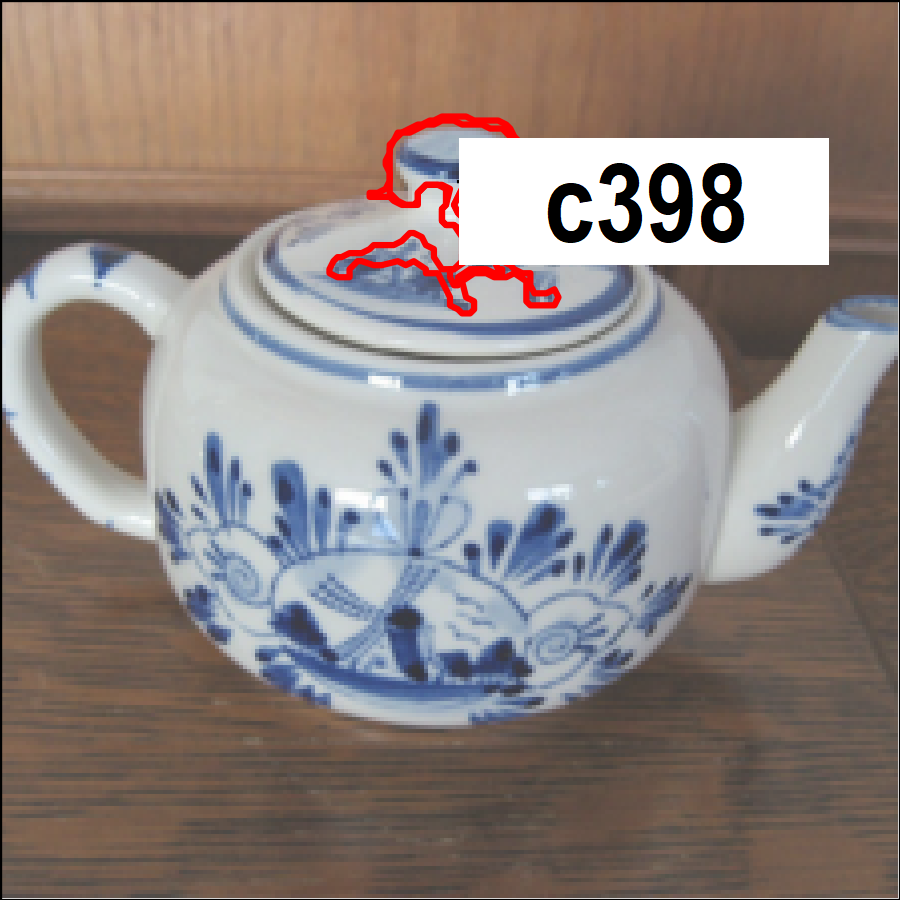}}
        \frame{\includegraphics[width=.32\columnwidth]{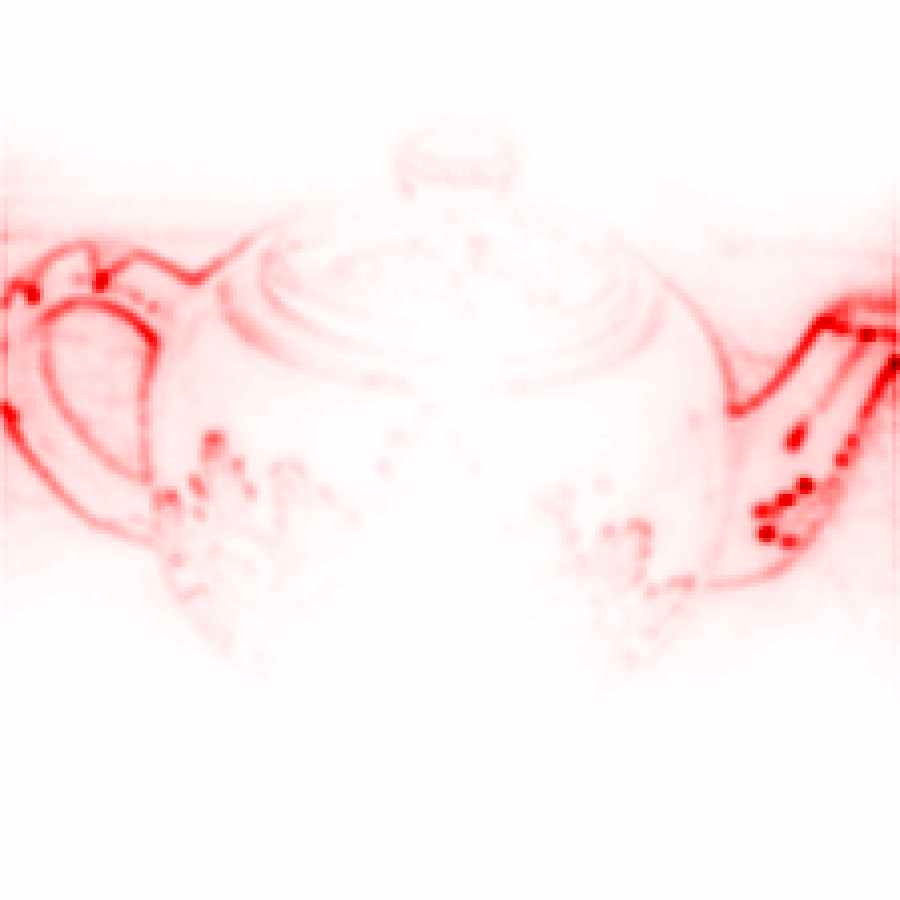}}
        \frame{\includegraphics[width=.32\columnwidth]{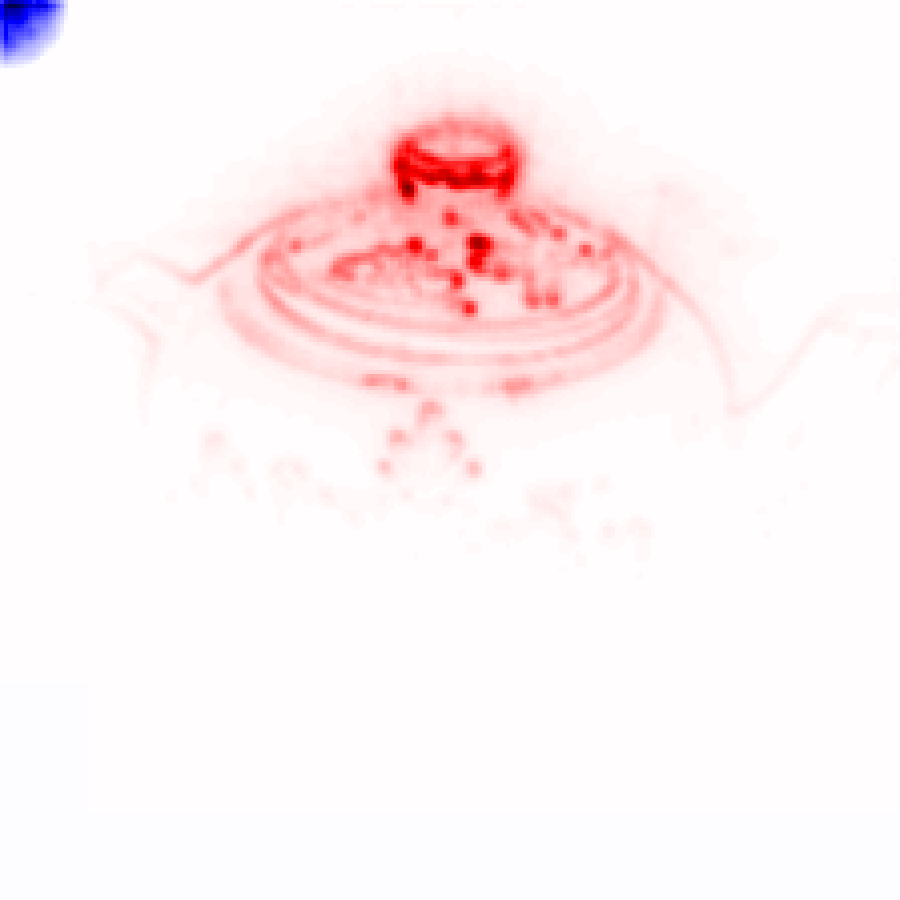}}
        
    \end{subfigure}
    \end{subfigure}
    \caption{The generated CoReX explanations for Aleph and Popper, containing the concepts ``handle", ``spout'' and ``lid". Additionally, samples that are covered by the explanations are shown, with the concept polygon borders and relevance values highlighted in heatmaps.}
    \label{fig:teapot_vase_explanations}
\end{figure}

\begin{figure}[bt!]
    \centering
        \begin{subfigure}[b]{0.45\textwidth}
        \centering
        \footnotesize
        Aleph:\\
        \emph{Female, if positive concept c157 (eyes) is above of positive concept c439 (smiling mouth) or if positive concept c157 (eyes) is left of positive concept c259 (hair)}.\\
        \vspace{.15cm}
        Popper:\\
        \emph{Female, if positive concept c15 (hair concept) is contained or positive concept c259 (hair concept) is contained.}
        \vspace{.15cm}
    \end{subfigure}
    \begin{subfigure}[b]{0.45\textwidth}
        \frame{\includegraphics[width=.49\columnwidth]{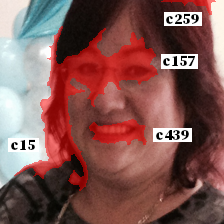}} 
        \frame{\includegraphics[width=.49\columnwidth]{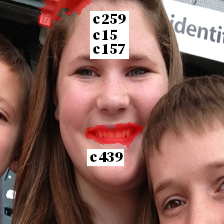}}
    \end{subfigure}
    \caption{The generated CoReX explanations for Aleph and Popper, containing the concepts ``eyes'', ``smiling mouth'' and ``hair". Additionally, samples that are covered by the explanations are shown, with the concept polygons highlighted as overlays on original images.}
    \label{fig:adience_explanations}
\end{figure}

First, some examples of explanations that CoReX generates are illustrated. For the Teapot-Vase experiment, common concepts occurring in the rules of Aleph and Popper were that of the handle, the spout and the lid of the teapot. An indicator of female persons in the Adience-FM data were concepts indicating hair, eyes and a smiling mouth. Figure~\ref{fig:teapot_vase_explanations} shows the fitting generated explanations as well as the highlighted ``handle", ``spout'' and ``lid" concepts. Figure~\ref{fig:adience_explanations} shows explanations and the highlighted ``hair", ``eyes" and ``mouth" concepts. For both Aleph and Popper, similar concepts are present in their respective generated explanations, hinting on robustness of our approach across different ILP methods. Note, that the abstract concepts need to be labeled by examining samples of multiple images and the location of their concepts, e.g., via reference sampling, introduced earlier.

\subsubsection{Human Study}

To evaluate the usefulness of concept-based and relational explanations a study involving human experts in ornithology and birding was conducted. In comparison to the domains of the other data sets (e.g., medicine for PathMNIST), it was considered easier to find experts for the evaluation of explanations generated for a CNN model for birds classification. Usefulness was measured by asking, whether attributes relevant to the classification of given bird examples were emphasized by presented explanations and whether they were helpful to understand the classification. The experts were given the task to rank explanations according to their usefulness in distinguishing two different bird species from given images.

\paragraph{Preparation:} The material of the study has been developed in cooperation with two experts from the Bavarian Association for Bird and Nature Conservation and one gradu\-ate biologist, trained in species identification (birds) by the German nature conservation association BUND (\textit{Friends of the Earth}). The study preparation consisted of interviews with the experts about selected data samples and explanations as well as pre-studies. All experts have been active in birding or related activities since more than 20 years. It should be noted that the experts' native language was German. The explanations together with the other study material were therefore translated from English to German and feedback regarding appropriateness was collected during material discussions with the experts.

\paragraph{Participants:}
The conduct of the study involved three experts different from the ones in the preparation phase, either trained in bird species identification or birding and bird counting as well as one professionalized bird photographer having significant international experience. One of the four experts was member of the Bavarian Association for Bird and Nature Conservation. Another one recently underwent intensive training in species identification and birding. At the time of conduct, all experts were still active in birding or related activities.

\paragraph{Material:} The images of bird species were selected from the CUB\_200\_2011 data set. Two particular bird families were chosen, Passerellidae (sparrows) and Icteridae (blackbirds), due to their interesting contrastive properties (high similarity and/or relational properties in bird appearance). For each bird family (sparrows and blackbirds), four classes were selected, which were then combined to two classes within each bird family. To select and group these birds, similarity graphs were used and complemented by automated image-based clustering. In this manner, the pairing brewer’s and vesper sparrows versus fox and song sparrows and the pairing brewer’s and rusty blackbirds versus red-winged and tricolored blackbirds were formed. These pairings were approved by the involved experts.

For the study, the images served as input to an image classifier that was trained and tested to create concept-based (visual) and relational (verbal) explanations. Table \ref{tab:cub_instances} denotes the instances per class and species from the selection drawn from the CUB\_200\_2011 data set for training and testing the image classifier. The pairing brewer’s and vesper sparrow contained 559 training samples (B: 280, V: 279) and the fox and song sparrows pairing contained 556 samples (S: 284, F: 282). The pairing brewer’s and rusty blackbirds consisted of 559 samples for training (B: 274, Ru: 285) and red-winged and tricolored blackbirds of 524 samples (Re: 285, Tri: 239). Likewise, the sizes of the testing data sets were 71, 70, 72 and 73. For improved reproducibility of our study Table \ref{tab:cub_model} shows the parameters of the model trained and tested on the selected CUB\_200\_2011 data set. For training the images were resized to 224x224 and augmented by rotating, zooming, changing the brightness, flipping them horizontally, and applying a shear. The model was trained and tested for the different pairings, such that two species formed the target class and the other two species the contrastive class. Table  \ref{tab:cub_model} lists the different settings and their F1 scores respectively: For the sparrows B. and V. as target or S. and F. as target the F1 score was above 0.9. For the blackbirds B. and Ru. as target or Re. and Tri. as target the scores reached even higher numbers. This was considered an important precondition to generate explanations that represent relevant class-discriminating features.

\begin{table}[]
\centering
\caption{Number of instances per class and species for training and testing an image classifier.}
\label{tab:cub_instances}
\begin{tabular}{@{}lll|lll@{}}
\toprule
Sparrows    & Training    & Testing    & Blackbirds      & Training   & Testing    \\ \midrule
Brewer's     & 280    & 36    & Brewer's      & 274    & 36    \\
Vesper       & 279    & 35    & Rusty         & 285    & 36    \\
Song         & 284    & 35    & Red-winged    & 285    & 36    \\
Fox          & 282    & 35    & Tricolored    & 239    & 37    \\ \bottomrule
\end{tabular}
\end{table}

\begin{table}[]
\caption{The characteristics of the model pre-trained on the CUB\_200\_2011 data set and fine-tuned on two contrastive classes for human evaluation.}
\label{tab:cub_model}
\begin{tabular}{@{}cccccccccc@{}}
\toprule
Spec.   & \#Train                                             & \#Test                                           & \begin{tabular}[c]{@{}c@{}}F1 \\ Train\end{tabular}                                & \begin{tabular}[c]{@{}c@{}}F1 \\ Test\end{tabular}                                 & \begin{tabular}[c]{@{}c@{}}Batch \\ Size\end{tabular} & \begin{tabular}[c]{@{}c@{}}Max. \\ Ep.\end{tabular} & \begin{tabular}[c]{@{}c@{}}Opti\\ mizer\end{tabular} & Loss                                                   \\ \midrule
Sparr.  & \begin{tabular}[c]{@{}c@{}}559, \\ 556\end{tabular} & \begin{tabular}[c]{@{}c@{}}71 \\ 70\end{tabular} & \begin{tabular}[c]{@{}c@{}}BV/SF: \\ 0.995 \\ SFBV: \\ 0.999\end{tabular}          & \begin{tabular}[c]{@{}c@{}}BV/SF: \\ 0.920 \\ SFBV: \\ 0.950\end{tabular}          & 4                                                     & 2                                                   & SGD                                                  & \begin{tabular}[c]{@{}c@{}}Cr. \\ Entr.\end{tabular} \\ \midrule
Blackb. & \begin{tabular}[c]{@{}c@{}}559, \\ 524\end{tabular} & \begin{tabular}[c]{@{}c@{}}72 \\ 73\end{tabular} & \begin{tabular}[c]{@{}c@{}}BRu/TriRe: \\ 0.996 \\ TriRe/BRu: \\ 1.000\end{tabular} & \begin{tabular}[c]{@{}c@{}}BRu/TriRe: \\ 0.966 \\ TriRe/BRu: \\ 0.960\end{tabular} & 4                                                     & 2                                                   & SGD                                                  & \begin{tabular}[c]{@{}c@{}}Cr. \\ Entr.\end{tabular} \\ \bottomrule
\end{tabular}
\end{table}

\paragraph{Study Design and Evaluation Metrics:}

As mentioned earlier, a group of 4 raters with backgrounds in ornithology and birding participated in this study. Before the rating task, they underwent a short training session that included familiarising them with the study material and the rating scheme, practising rating exercises, and clarifying any ambiguities.

Four variants of explanations were shown, being the items to be rated: (1) visual explanations in the form of heatmaps (similar to those in Figure \ref{fig:teapot_vase_explanations}), (2) verbal explanations (a template-based translation from Prolog to natural language as done by \cite{finzel2021explanation}), (3) combined explanations (CoReX) and (4) official descriptions of species.
The rating scheme was a standard rating scheme, where participants were asked to put the four explanation items in order according to their usefulness (highest rank: 4, lowest rank: 1). Each rater provided ratings independently.

Krippendorff's Alpha ($\alpha$) was employed to assess the inter-rater reliability of the ratings \cite{krippendorff2018content}. This statistical measure is suitable for studies with multiple raters and ratings with an ordinal scale. Further interesting properties, although not relevant for the outcomes of our study, is the fact that Krippendorff's $\alpha$ can handle different sample sizes and missing data.

\paragraph{Results:}

\begin{table}[]
\centering
\caption{For the measurement of rater agreement Krippendorff's $\alpha$ values were computed \cite{krippendorff2018content}. Values were collected for 32 items of the sparrow subset, 32 items of the blackbirds subset and of all 64 items. For the sparrows we find moderate agreement.}
\label{tab:alpha_values}
\begin{tabular}{@{}cll@{}}
\toprule
\multicolumn{3}{c}{Krippendorff's $\alpha$ values}                                                             \\ \midrule
Sparrows (32 items)        & \multicolumn{1}{c}{Blackbirds (32 items)} & \multicolumn{1}{c}{All (64 items)} \\ \midrule
\multicolumn{1}{l}{0.7032} & 0.1766                                    & 0.4364                             \\ \bottomrule
\end{tabular}
\end{table}

With Krippendorff's $\alpha$ we measured the extent of agreement among raters beyond chance. An $\alpha$ of 1 would indicate perfect agreement and a value of -1 perfect disagreement. Values beyond 0.8 are considered highly reliable to support agreement. Values between 0.67 and 0.79 indicate moderate agreement. Values between 0 and 0.67 indicate low agreement. An $\alpha$ of 0 indicates random rating patterns as suggested by \cite{krippendorff2018content}. Table \ref{tab:alpha_values} presents the main results. For the items from the sparrows subset we find moderate agreement. For the blackbirds subset the agreement is considerably lower among raters. For all items in total Krippendorff's $\alpha$ indicates agreement patterns somewhere between moderate agreement and slightly diverging agreement.

\begin{figure}[bt!]
        \centering
        \includegraphics[width=0.7\columnwidth]{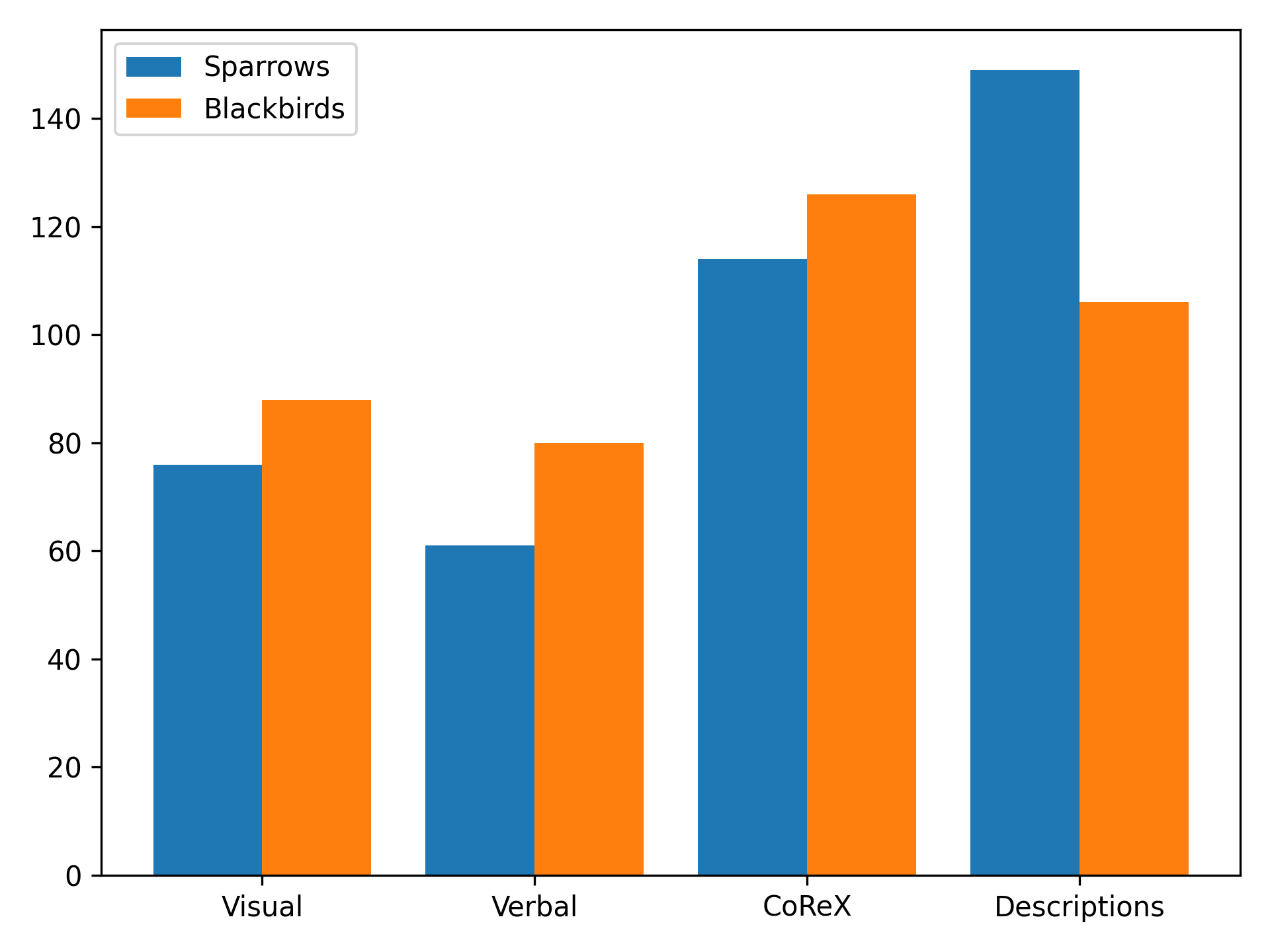} 
    \caption{Outcomes of the human evaluation: raters assigned higher scores to combined (concept-based and relational) explanations and descriptions of classes compared to only visual or verbal explanations among all study items.}
    \label{fig: human_study_birds}
\end{figure}

These findings are a precondition to robust interpretation of the usefulness of different explanations as presented in the study as they indicate that ratings did not result from chance. Figure \ref{fig: human_study_birds} shows the main result obtained from the human evaluation of experts, showing that combined explanations have been preferred over only visual or only verbal explanations.

\subsection{Explanatory Applicability of CoReX}

We enhance CoReX by exploration of sample classifications based on contrastive explanations, rule-based cluster analysis and user-defined constraints. This may serve deve\-lopers and experts to assess the quality of the underlying model during the development phase or may help novices to learn about the domain of interest (e.g., bird species as presented above) as pointed out in \cite{finzel2024human}.

\subsubsection{Contrastive Explanations}

Contrastive local explanations can be produced by evaluating an ILP model for a specific misclassified sample. This means that one can unify the background knowledge of a misclassified sample with rules from the ILP model (e.g., the rule with the highest coverage). The evaluation returns concepts and relations that are missing in the sample in order to belong to the target class. CoReX provides a component that verbalizes these failures with the help of an implemented Prolog-based prover (see code repository given in Section \ref{sec:declaration}). A demo of contrastive explanations is available via an implemented interface with examples\footnote{CoReX demonstrator: \url{https://gitlab.rz.uni-bamberg.de/cogsys/public/corex-demo}}.

\subsubsection{Cluster-based Evaluation} \label{sec:outlier}

Rule-based cluster analysis allows for outlier and prototypical example detection. In particular, we analyze, which rules jointly cover which samples by building the power set of combinations. We examine clusters that cover most of examples, samples not covered by any rule, and clusters that cover few samples with just a small set of rules. The clusters computed for the different data sets can be found in Section \ref{sec:clusters} of the Appendix.

\begin{figure}[bt!]
        \centering
        \includegraphics[width=\columnwidth]{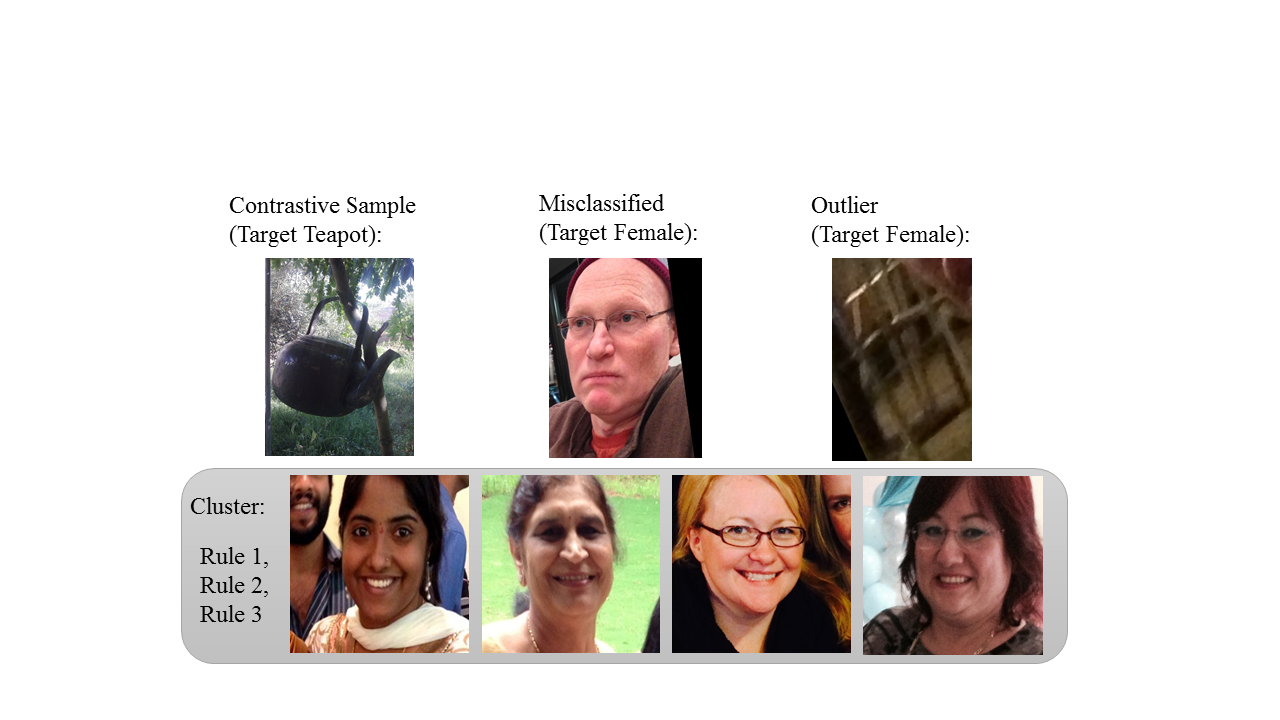} 
    \caption{Teapot (top left), false positive female (top middle), outlier female (top right) and rule cluster for smiling persons (bottom row).}
    \label{fig:contrastive_outlier}
\end{figure}

In Figure~\ref{fig:contrastive_outlier} we present a collection of interesting cases. Consider the contrastive sample from ImageNet (target: teapot), which is a teapot in the ground truth as well as in the model truth (CNN), but not in the explainer truth (ILP). When evaluating the ILP model on the sample, we can observe that a concept from the best covering rule was missing in the background knowledge: the spout. It did not belong to the most relevant concepts extracted from the CNN model for this particular sample and was therefore not included in the background knowledge and consequently also not in the rules of the ILP explainer. In such cases, the ILP explainer output can help to identify deviations from more general concepts and relations representative of a class.

Our cluster analysis on the Adience Test data set (target: female) yields interesting results as well. The most representative cluster mainly contains samples with the relation ``eyes above of smiling mouth'' (see Figure~\ref{fig:contrastive_outlier}). For more details, we refer to Figure \ref{fig:cluster_adience_test_fm} and Figure \ref{fig:rankings_adience_test_fm} in the Appendix. In particular, Figure \ref{fig:rankings_adience_test_fm} in combination with our illustrations in Figure \ref{fig:adience_explanations} shed light into the meaning of concepts extracted from the CNN and deemed important by the ILP learner. The labels for concepts were determined based on most relevant image regions as provided by CRP.

Further analysis identifies a cluster that is small with few examples covered. In this cluster all samples have in common that one concept comprises a border artefact, among others this applies to the outlier, which is displayed in the top row on the right side of Figure~\ref{fig:contrastive_outlier}. One cluster was empty (an example not covered by any rule). This example was misclassified by the CNN model as ``female'', although its ground truth was ``male''. The coverage of the sample fails in ILP due to failed unification of the background knowledge of the sample with the general rules learned by ILP. This is in line with the observation that the sample was misclassified by the CNN. Our analysis indicates that the CNN was not using any of the most relevant concepts to decide on the classification of the misclassified sample. This way, ILP as an explainer helps to identify abnormal model behavior.

\subsubsection{Constraint-based Evaluation}

As pointed out in Figure \ref{fig:overview} and Algorithm \ref{alg:mainAlg} the CoReX framework allows for putting constraints on learned concepts and relations. As supported by the results of the ablation study in Section \ref{sec:results}, constraining concepts in the CNN by masking them can lead to changes in the predictive outcomes of such an image classifier, since specific concepts are excluded from being used in the prediction process. This is desirable, whenever the used concepts have been incorrect from a human expert's point of view or where a more concise model is needed to save computational resources (in such a case, masking can be complemented by pruning and re-training parts of a model).

Concept masking is applied globally, meaning that the predictive process changes for all input data that is processed by a constrained model. However, in the case of exceptional cases, such as the outlier or individual misclassifications as presented before, it may be helpful to first explore the effect of constraints in the explainer (the interpretable surrogate model).

To complement the qualitative evaluation of our CoReX framework, we integrated user-defined constraints on concepts and their relations as provided by the Aleph ILP framework \cite{srinivasan2007aleph}. The following example shall demonstrate the usefulness of such constraints in combination with contrastive and clustered rule-based explanations.

Consider again the outlier in the top right corner of Figure \ref{fig:contrastive_outlier}, where a border artifact was included in an explanatory rule learned by the ILP explainer and identified by rule-based cluster analysis. With the help of a user-defined constraint, it is possible to exclude the border artifact and to update the learned explanatory ILP model.

By doing this, we were able to revise the interpretable ILP-based surrogate model. New rules were learned and covered all previous examples, except for the outlier with the border artifact, which is a desirable outcome. Consequently, the surrogate model found alternative concepts and relations, which did not reduce the fidelity of the rules in the ILP model with respect to the CNN model, except for the classification of the outlier. This showcases that including user-defined constraints on concepts and relations can help to handle outliers and misclassifications and to identify alternative, equally representative concepts and relations with the help of a constrained interpretable surrogate model without compromising global applicability of the concepts in the already learned CNN model.

\section{Conclusion and Future Work}\label{sec:conclusion}

We presented CoReX, a novel XAI approach that explains classifications of CNNs through an interpretable surrogate model. It relies on identifying human-understandable concepts in intermediate layers of a CNN and on inducing relations between such concepts. Concept relevance is combined with logic rules constructed from learned ILP theories to provide expressive explanations.

An evaluation with different data sets shows that the CoReX explainer is highly faithful to the examined CNN models. This is demonstrated by computed fidelity values as well as by a concept ablation study. The ablation study shows high performance drops for concepts included in the rules of the ILP explainer compared to ablations of irrelevant concepts, which had no effect or marginal effects in comparison.
An evaluation performed in a human study indicates the usefulness of CoReX explanations compared to only visual and verbal explanations. Our results show that experts clearly preferred combined explanations given positive inter-rater reliability. We further showcased that CoReX  provides meaningful explanations for correctly and incorrectly classified instances using contrastive explanations and rule-based cluster analysis.

For making CoReX applicable to XIML use cases, we provide a framework to interact with explanations and the model respectively by including constraints in the algorithm that allow for giving the information about what concept must be suppressed or kept by a model or explainer. CoReX may thus be pathing the way for regularization and revision of the model to align with the provided corrective feedback \cite{bontempelli2022debugging,dash2022review,finzel2024human}.

In the future, the evaluation of our approach and similar techniques would highly benefit from a larger availability of contrastive data sets, where the classification of an image depends on the relations between concepts and not just their presence. As already laid out in the motivation, the instances of many domains in the real world can be explained in a way that aids humans by contrasting them with similar instances. We also showed that it is possible to extend CoReX by further logic-based systems like Popper \cite{cropper2021learning} and motivated the use of probabilistic methods \cite{fadja2021learning}. Future work will aim at evaluating the already integrated, alternative ILP systems to provide a comprehensive analysis of the generalizability of our findings.

\backmatter

\section*{Declarations}\label{sec:declaration}

\subsection{Funding}

This work was partially funded by the German Federal Ministry of Education and Research under grant FKZ 01IS18056~B, BMBF ML-3 (TraMeExCo, 2018-2021) and partially supported by the German Research Foundation under grant DFG 405630557 (PainFaceReader, 2021-2023).

\subsection{Ethics approval}

Not applicable.

\subsection{Consent to participate}

Not applicable.

\subsection{Consent for publication}

The authors affirm that human research participants provided informed consent for publication of the images in Figures \ref{fig:example_images} and \ref{fig:contrastive_outlier}. In particular, the Adience data set \cite{eidinger2014age}, consists only of material, including human faces, which has been deliberately provided under a Creative Commons (CC) licence on Flickr\footnote{Licence declaration of the Adience data set: \url{https://talhassner.github.io/home/projects/Adience/Adience-data.html}}. The Picasso data set \cite{rabold2020expressive} contains completely retouched faces. Moreover, facial features (eyes, nose, mouth) are randomly taken from a pool of features to generate artificial faces, which render persons unrecognizable.

\subsection{Competing interests}

The authors have no conflicts of interest to declare that are relevant to the content of this article.

\subsection{Availability of data and materials}

See code availability.

\subsection{Code availability}

The code basis of our CoReX approach and scripts for evaluation purposes are available at the linked repository\footnote{Repository containing the complete code base and evaluation scripts: \url{https://gitlab.rz.uni-bamberg.de/cogsys/public/corex}}. 

\subsection{Authors' contributions}

All authors contributed to this work. B.F. initiated the paper, wrote a first complete draft and contributed to all sections for the submitted version. B.F. and P.H. developed the concept behind the methodology. P.H. contributed to the implementation of the methodology. B.F. revised and extended the implementation. P.H., J.R. and B.F. planned and carried out the experimental evaluations and revised the first draft. J.R. provided substantial input for related work on concept analysis and fidelity measures. U.S. contributed with revisions of the first draft and substantial input for related work on cognitive concepts. All authors have read and approved the submitted version of this paper.

\subsection{Acknowledgments}

We would like to thank Philippe Crackau for preparing experiments based on Probabilistic Inductive Logic Programming and Luisa Schneider for preparing the computation of concepts for our human evaluation. We further thank Reduan Achtibat who provided support for the configuration of Concept Relevance Propagation and Sebastian Lapuschkin for fruitful discussions on relevance-based feature extraction. We further thank the experts in ornithology and birding for participating in the preparation and conduction of the human evaluation study.

\newpage

\bibliographystyle{sn-basic}
\bibliography{sn-bibliography}

\newpage

\begin{appendices}

\section{Spatial Relations}\label{sec:spatial}

See Table \ref{tbl:relations}.

\begin{table}[h!]
\centering
\caption{Overview of all supported relations. Relations are organized in unary and binary sets and apply w.r.t. a sample. Negative relations represent a concept's relevance scores sign. Sets like the SimpleAlignment, Distance and Surrounding are implemented based on the geometric interfaces of the python shapely library \protect\cite{gilley2022shapely}. CompassAlignment is loosely based on a direction-based spatial navigation model of \protect\cite{Sriharee15}. NineIntersectionModel is based on the DE-9IM developed by Clementini, et al. \protect\cite{clementini1996DE9im,ClementiniFO93} with the omission of the intersect-relation
}
\label{tbl:relations}
\scalebox{0.75}{

\begin{tabular}{p{3cm} l l l p{4cm}}
\toprule
\textbf{Set} & \textbf{relation} & \textbf{definition} & \textbf{semantic} \\
\midrule
\multicolumn{2}{l}{unary relations} \\
\midrule
Existence            & \textit{$has\_a(A)$} & $neg(A) \lor pos(A)$ & has A\\
Negativity           & \textit{$neg(A)$} & $R(A) < 0$ & A, that shouldn't be there\\
    & \textit{$pos(A)$} & $R(A) > 0$ & A, that should be there\\
\midrule
\multicolumn{2}{l}{binary relations} \\
\midrule
SimpleAlignment     & \textit{$above\_of(A,B)$} & $A.centroid.y < B.centroid.y$ & A is above of B\\
					 & \textit{$below\_of(A,B)$} & $A.centroid.y > B.centroid.y$ & A is below of B\\
					 & \textit{$left\_of(A,B)$} & $A.centroid.x < B.centroid.x$ & A is left of B\\
					 & \textit{$right\_of(A,B)$} & $A.centroid.x > B.centroid.x$ & A is right of B\\
CompassAlignment    		 & \textit{$center(A,B)$} & $A.centroid.buffer.intersects(B)$ & B is centered on A\\
					 & \textit{$middle\_right(A,B)$} & $\triangle(A,l_{\frac{15}{8}\pi},l_{\frac{1}{8}\pi}).intersects(B)$
					 & B is middle right to A\\
					 & \textit{$bottom\_right(A,B)$} & $\triangle(A,l_{\frac{1}{8}\pi},l_{\frac{3}{8}\pi}).intersects(B)$ & B is bottom right to A\\
					 & \textit{$bottom\_middle(A,B)$} & $\triangle(A,l_{\frac{3}{8}\pi},l_{\frac{5}{8}\pi}).intersects(B)$ & B is middle bottom to A\\
					 & \textit{$bottom\_left(A,B)$} & $\triangle(A,l_{\frac{5}{8}\pi},l_{\frac{7}{8}\pi}).intersects(B)$ & B is bottom left to A\\
					 & \textit{$middle\_left(A,B)$} & $\triangle(A,l_{\frac{7}{8}\pi},l_{\frac{9}{8}\pi}).intersects(B)$ & B is middle left to A\\
					 & \textit{$top\_left(A,B)$} &		 $\triangle(A,l_{\frac{9}{8}\pi},l_{\frac{11}{8}\pi}).intersects(B)$ & B is top left to A\\
					 & \textit{$top\_middle(A,B)$} & $\triangle(A,l_{\frac{11}{8}\pi},l_{\frac{13}{8}\pi}).intersects(B)$ & B is middle top to A\\
					 & \textit{$top\_right(A,B)$} & $\triangle(A,l_{\frac{13}{8}\pi},l_{\frac{15}{8}\pi}).intersects(B)$ & B is top right to A\\
NineIntersectionModel& \textit{$disjoint(A,B)$} & $A.disjoint(B)$ & A is disjoint of B\\
					 & \textit{$equals(A,B)$} & $A.equals(B)$ & A equals B\\
                     & \textit{$touches(A,B)$} & $A.touches(B)$ & A touches B\\
					 & \textit{$overlaps(A,B)$} & $A.overlaps(B)$ & A overlaps B\\
					 & \textit{$covers(A,B)$} & $A.covers(B)$ & A covers B\\
                     & \textit{$contains(A,B)$} & $A.contains(B)$ & A contains B\\
                     & \textit{$covered\_by(A,B)$} & $A.covered\_by(B)$ & A is covered by B\\
                     & \textit{$within(A,B)$} & $A.within(B)$ & A is within B\\
Distance             & \textit{$close\_to(A,B)$} & $A.distance(B) < range$ & A is close to B\\
\midrule
\multicolumn{2}{l}{special binary relation} \\
\midrule					
Surrounding          & \textit{$amid\_x(A,B)$} & $ B.between(A_1.centroid.x, A_2.centroid.x) $ & B is horizontally \\& & & surrounded by A\\

                     & \textit{$amid\_y(A,B)$} & $ B.between(A_1.centroid.y, A_2.centroid.y) $ 
                     & B is vertically \\ & & & surrounded by A\\
\bottomrule
\end{tabular}
}
\end{table}

\newpage

\section{Number of Concepts}\label{sec:number}

\begin{figure}[!h]
    \centering
    \includegraphics[width=.9\textwidth]{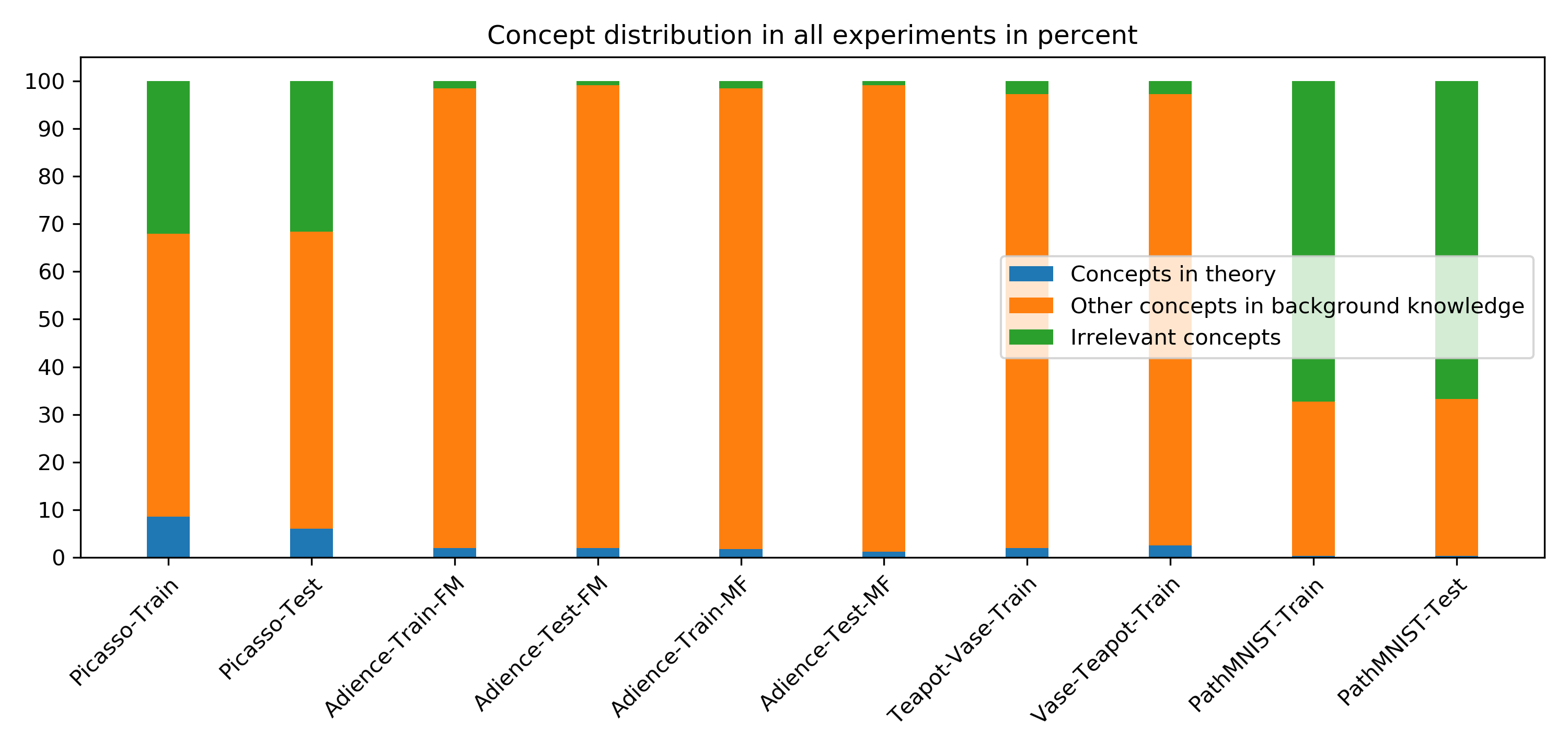}
    \caption{The distribution of concepts (given in percent) in the theory, in the background knowledge (excluding the theory concepts), and irrelevant concepts for the Aleph experiments. See Table~\ref{tab:number_of_concepts} for a detailed breakdown}
    \label{fig:number_of_concepts}
\end{figure}

Figure~\ref{fig:number_of_concepts} shows the distribution of concepts for all experiments conducted with Aleph, split by the concepts occurring in the rules of a learned ILP theory, the other concepts in the background knowledge, and the irrelevant concepts. Table~\ref{tab:number_of_concepts} additionally gives the absolute values and percentages presented in Figure~\ref{fig:number_of_concepts}.

\begin{table*}[!h]
    \centering
    \caption{The number of concepts in the theory, in the background knowledge (excluding the theory concepts), and irrelevant concepts for the Aleph experiments}
    \label{tab:number_of_concepts}
    \begin{tabular}{l|ll|ll|ll}
    \toprule
        \textbf{Experiment} & \# Rule & \% Rule & \# BK & \% BK & \# Irrel. & \% Irrel.\\
        \midrule
        Picasso-Train & 44 & 8.59 & 304 & 59.38 & 164 & 32.03\\
        Picasso-Test & 31 & 6.05 & 319 & 62.30 & 162 & 31.64\\
        Adience-Train-FM & 10 & 1.95 & 494 & 96.48 & 8 & 1.56\\
        Adience-Test-FM & 10 & 1.95 & 497 & 97.07 & 5 & 0.98\\
        Adience-Train-MF & 9 & 1.76 & 495 & 96.68 & 8 & 1.56\\
        Adience-Test-MF & 6 & 1.17 & 501 & 97.85 & 5 & 0.98\\
        Teapot-Vase-Train & 10 & 1.95 & 488 & 95.31 & 14 & 2.73\\
        Vase-Teapot-Train & 13 & 2.54 & 485 & 94.73 & 14 & 2.73\\
        PathMNIST-Train & 6 & 0.29 & 664 & 32.42 & 1378 & 67.29\\
        PathMNIST-Test & 7 & 0.34 & 674 & 32.91 & 1367 & 66.75\\
    \bottomrule
    \end{tabular}
\end{table*}

\begin{figure}[!h]
    \centering
    \includegraphics[width=.9\textwidth]{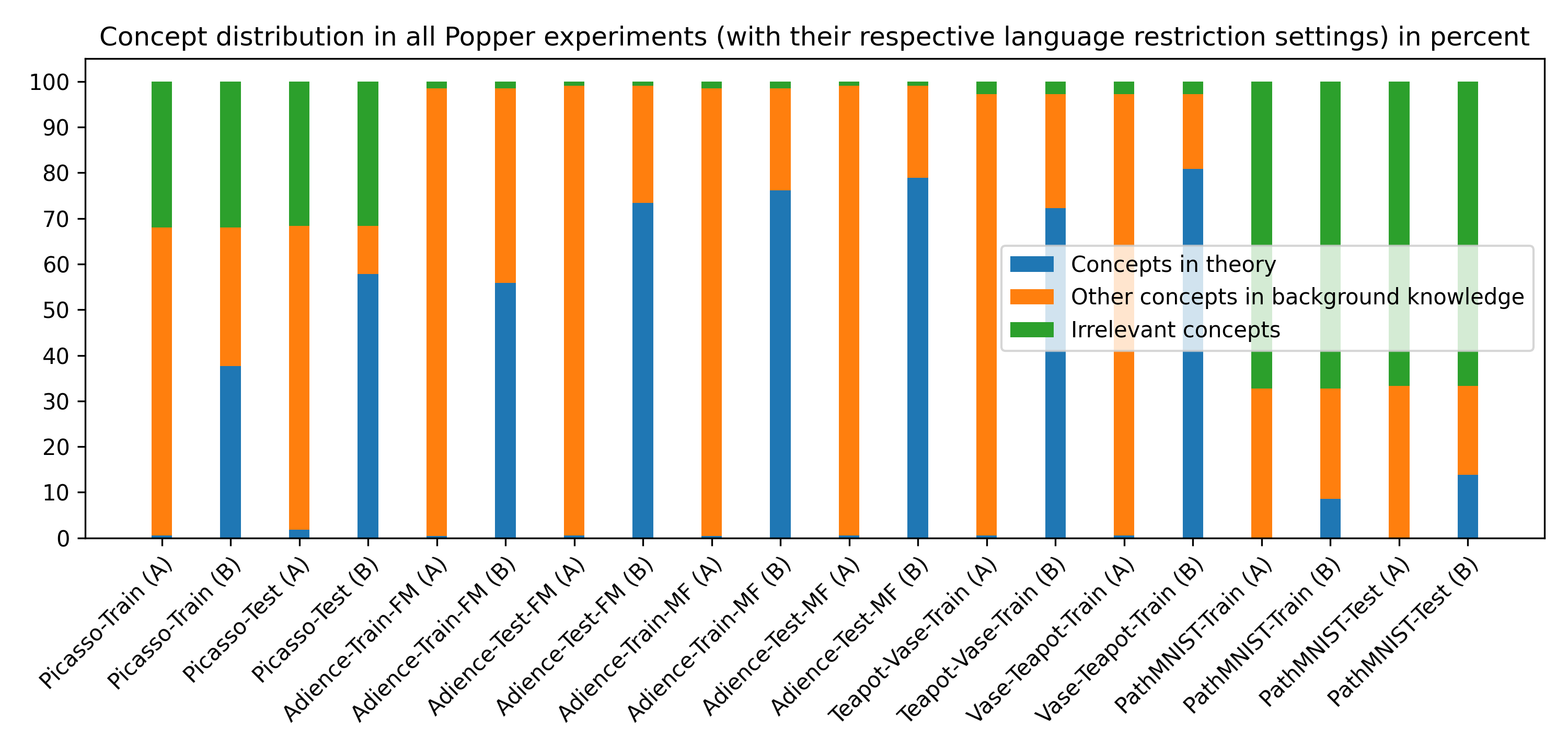}
    \caption{The distribution of concepts (given in percent) in the theory, in the background knowledge (excluding the theory concepts), and irrelevant concepts for the Popper experiments. The two settings \emph{A} and \emph{B} from above are stated separately. See Table~\ref{tab:number_of_concepts_popper} for a detailed breakdown}
    \label{fig:number_of_concepts_popper}
\end{figure}

Figure~\ref{fig:number_of_concepts_popper} shows the distribution of concepts for all experiments conducted with Popper, split by the concepts occurring in the rules of a learned ILP theory, the other concepts in the background knowledge, and the irrelevant concepts. Two approaches on guiding the search for clauses were used: 
In setting \emph{A}, positive and negative concepts need to be found in the background knowledge, before any relations can use them and be included into a clause. In setting \emph{B}, first, relations need to be found in the background knowledge, before fitting positive and negative concepts are introduced into the clause. We let all experiments run for a maximum of 30 minutes before collecting the resulting theories. Table~\ref{tab:number_of_concepts_popper} additionally gives the absolute values and percentages presented in Figure~\ref{fig:number_of_concepts_popper}. 


\begin{table*}[!h]
    \centering
    \caption{The number of concepts in the theory, in the background knowledge (excluding the theory concepts), and irrelevant concepts for the Popper experiments with their respective language constraint settings}
    \label{tab:number_of_concepts_popper}
    \begin{tabular}{l|ll|ll|ll}
        \toprule
        \textbf{Experiment} & \# Rule & \% Rule & \# BK & \% BK & \# Irrel. & \% Irrel.\\
        \midrule
        Picasso-Train (A) & 3 & 0.59 & 345 & 67.38 & 164 & 32.03\\
        Picasso-Train (B) & 193 & 37.70 & 155 & 30.27 & 164 & 32.03\\
        Picasso-Test (A) & 9 & 1.76 & 341 & 66.60 & 162 & 31.64\\
        Picasso-Test (B) & 296 & 57.81 & 54 & 10.55 & 162 & 31.64\\
        Adience-Train-FM (A) & 2 & 0.39 & 502 & 98.05 & 8 & 1.56\\
        Adience-Train-FM (B) & 286 & 55.86 & 218 & 42.58 & 8 & 1.56\\
        Adience-Test-FM (A) & 3 & 0.59 & 504 & 98.44 & 5 & 0.98\\
        Adience-Test-FM (B) & 376 & 73.44 & 131 & 25.59 & 5 & 0.98\\
        Adience-Train-MF (A) & 2 & 0.39 & 502 & 98.05 & 8 & 1.56\\
        Adience-Train-MF (B) & 390 & 76.17 & 114 & 22.27 & 8 & 1.56\\
        Adience-Test-MF (A) & 3 & 0.59 & 504 & 98.44 & 5 & 0.98\\
        Adience-Test-MF (B) & 404 & 78.91 & 103 & 20.12 & 5 & 0.98\\
        Teapot-Vase-Train (A) & 3 & 0.59 & 495 & 96.68 & 14 & 2.73\\
        Teapot-Vase-Train (B) & 370 & 72.27 & 128 & 25.00 & 14 & 2.73\\
        Vase-Teapot-Train (A) & 3 & 0.59 & 495 & 96.68 & 14 & 2.73\\
        Vase-Teapot-Train (B) & 414 & 80.86 & 84 & 16.41 & 14 & 2.73\\
        PathMNIST-Train (A) & 3 & 0.15 & 667 & 32.57 & 1378 & 67.29\\
        PathMNIST-Train (B) & 176 & 8.59 & 494 & 24.12 & 1378 & 67.29\\
        PathMNIST-Test (A) & 3 & 0.15 & 678 & 33.11 & 1367 & 66.75\\
        PathMNIST-Test (B) & 284 & 13.87 & 397 & 19.38 & 1367 & 66.75\\
        \bottomrule
    \end{tabular}
\end{table*}

\newpage

\section{Explainer Fidelity in Experiments with Popper}\label{sec:fidelity_other_ilp}

Table~\ref{tab:fidelity_popper} shows the explainer fidelity for the experiments conducted with Popper.

\begin{table*}[!h]
    \centering
    \caption{The explainer fidelity for all experiments conducted with Popper for the two language constraint settings described in Section~\ref{sec:number} of the Appendix}
    \label{tab:fidelity_popper}
    \begin{tabular}{l|l}
    \toprule
    \textbf{Experiment} & Explainer Fidelity\\
    \midrule
    Picasso-Train (A) & 0.8745\\
    Picasso-Train (B) & 0.8663\\
    Picasso-Test (A) & 0.8900\\
    Picasso-Test (B) & 0.8460\\
    Adience-Train-FM (A) & 0.9975\\
    Adience Train-FM (B) & 0.9975\\
    Adience-Test-FM (A) & 1.0000\\
    Adience-Test-FM (B) & 0.9950\\
    Adience-Train-MF (A) & 1.0000\\
    Adience-Train-MF (B) & 1.0000\\
    Adience-Test-MF (A) & 1.0000\\
    Adience-Test-MF (B) & 0.9975\\
    Teapot-Vase-Train (A) & 1.0000\\
    Teapot-Vase-Train (B) & 1.0000\\
    Vase-Teapot-Train (A) & 0.9970\\
    Vase-Teapot-Train (B) & 0.8852\\
    PathMNIST-Train (A) & 1.0000\\
    PathMNIST-Train (B) & 1.0000\\
    PathMNIST-Test (A) & 1.0000\\
    PathMNIST-Test (B) & 0.9940\\
    \bottomrule
    \end{tabular}
\end{table*}

\newpage

\section{Rule clusters that cover most samples}\label{sec:clusters}

\begin{figure}[!h]
	\centering
	\includegraphics[width=1\textwidth]{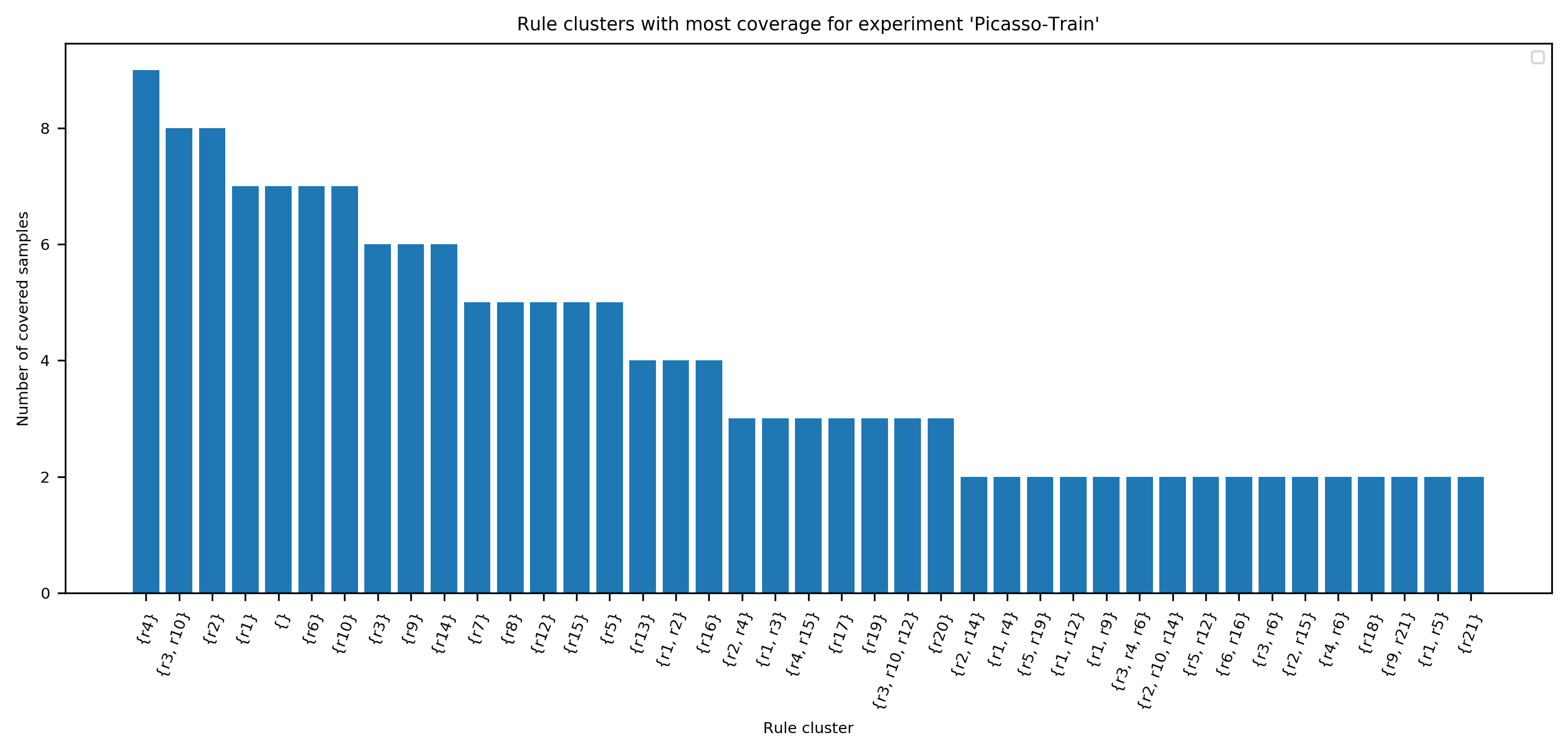}
	\caption{Rule clusters with most coverage for experiment 'Picasso-Train'. Clusters with a frequency of 1 are omitted for better readability. There were 89 such clusters}
	\label{fig:cluster_picasso_train}
\end{figure}

\begin{figure}[!h]
	\centering
	\includegraphics[width=1\textwidth]{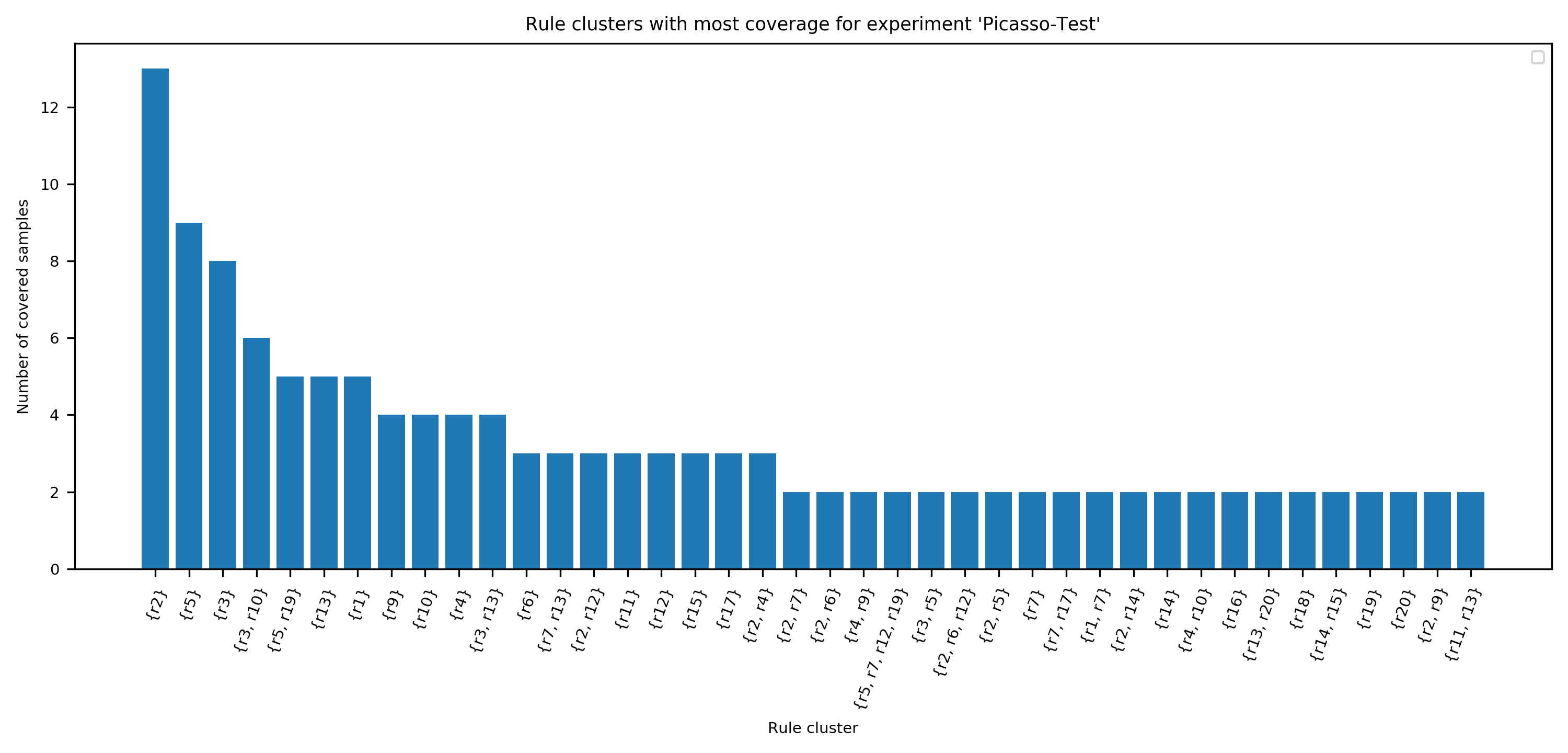}
	\caption{Rule clusters with most coverage for experiment 'Picasso-Test'. Clusters with a frequency of 1 are omitted for better readability. There were 117 such clusters}
	\label{fig:cluster_picasso_test}
\end{figure}

\begin{figure}[!h]
	\centering
	\includegraphics[width=1\textwidth]{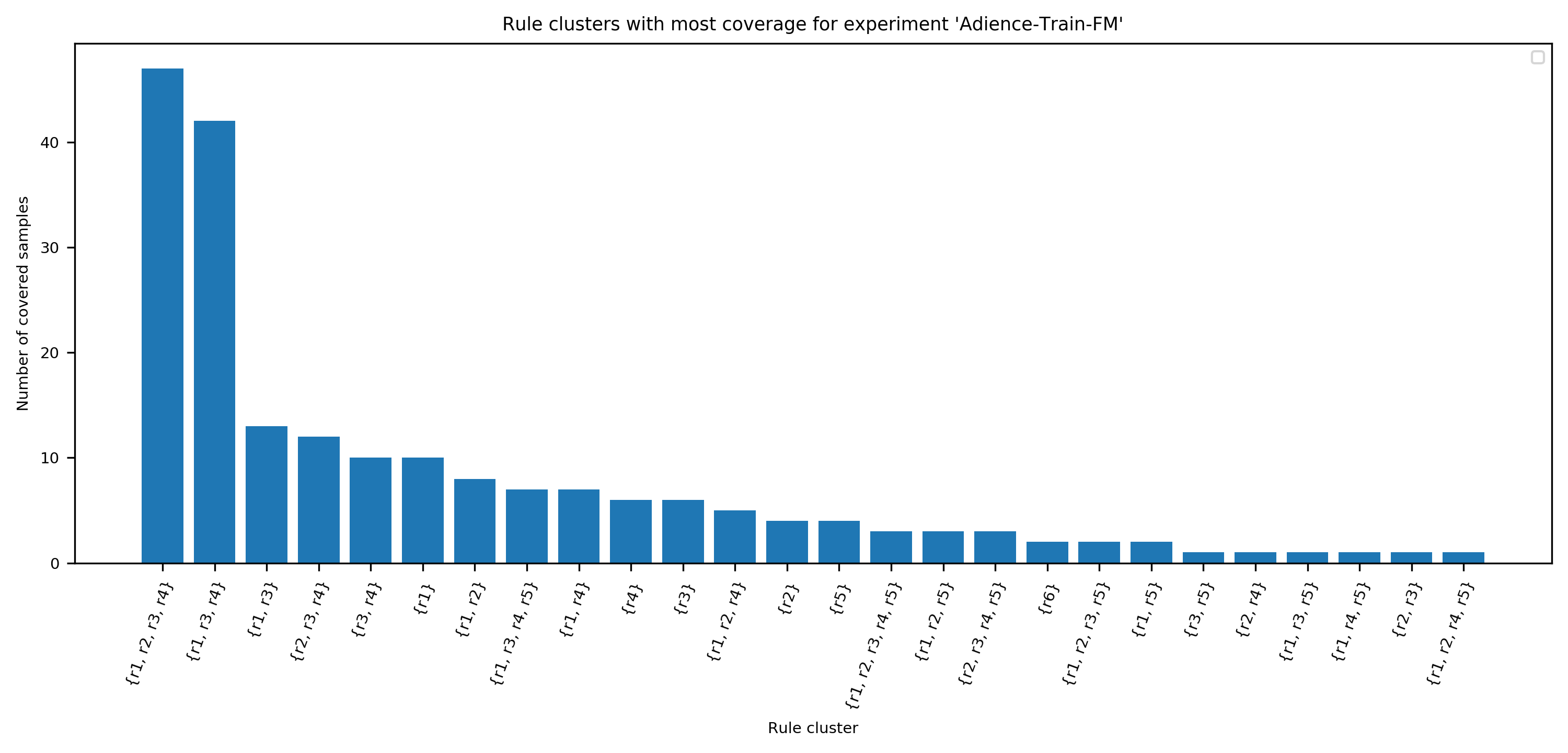}
	\caption{Rule clusters with most coverage for experiment 'Adience-Train-FM'}
	\label{fig:cluster_adience_train_fm}
\end{figure}

\begin{figure}[!h]
	\centering
	\includegraphics[width=1\textwidth]{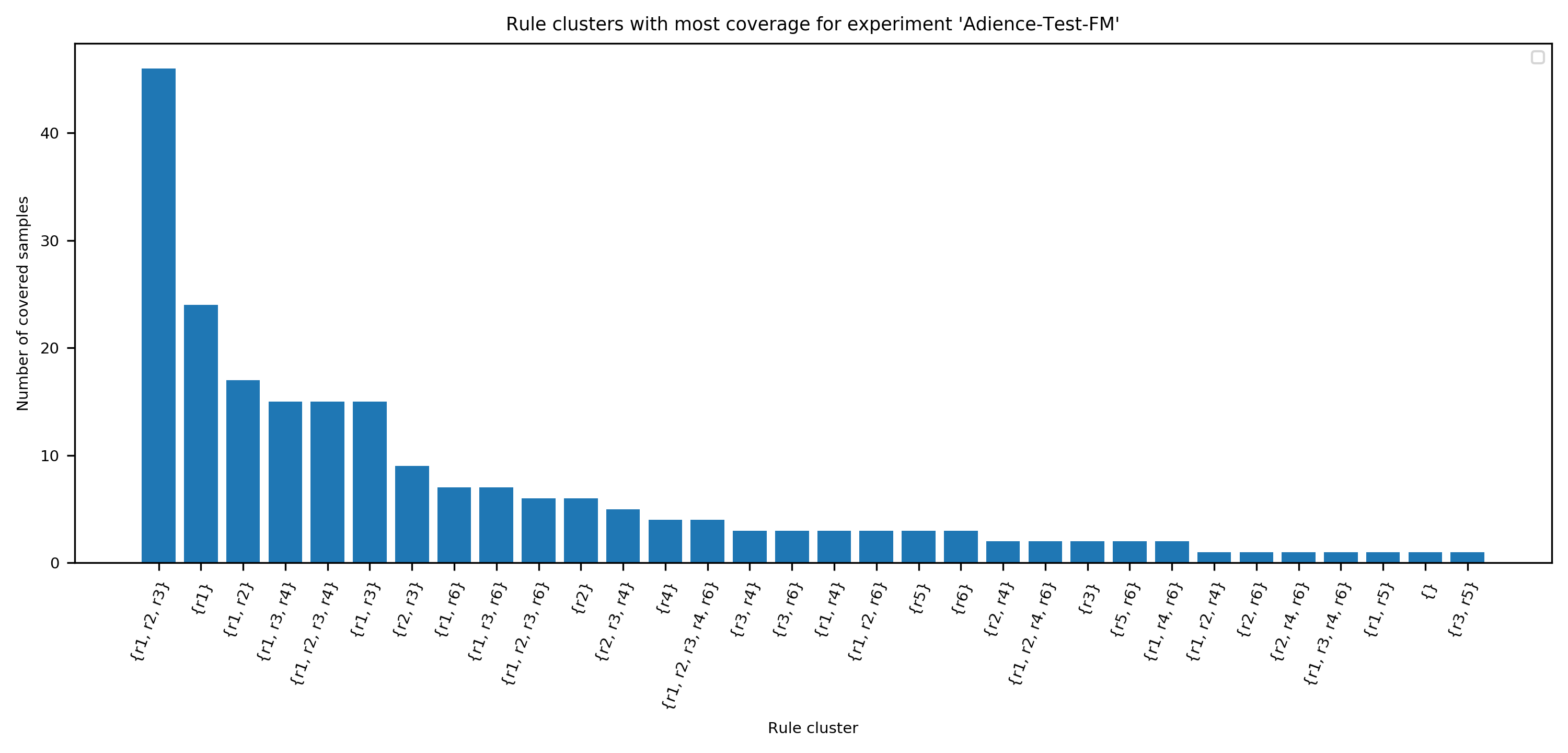}
	\caption{Rule clusters with most coverage for experiment 'Adience-Test-FM'}
	\label{fig:cluster_adience_test_fm}
\end{figure}

\begin{figure}[!h]
	\centering
	\includegraphics[width=1\textwidth]{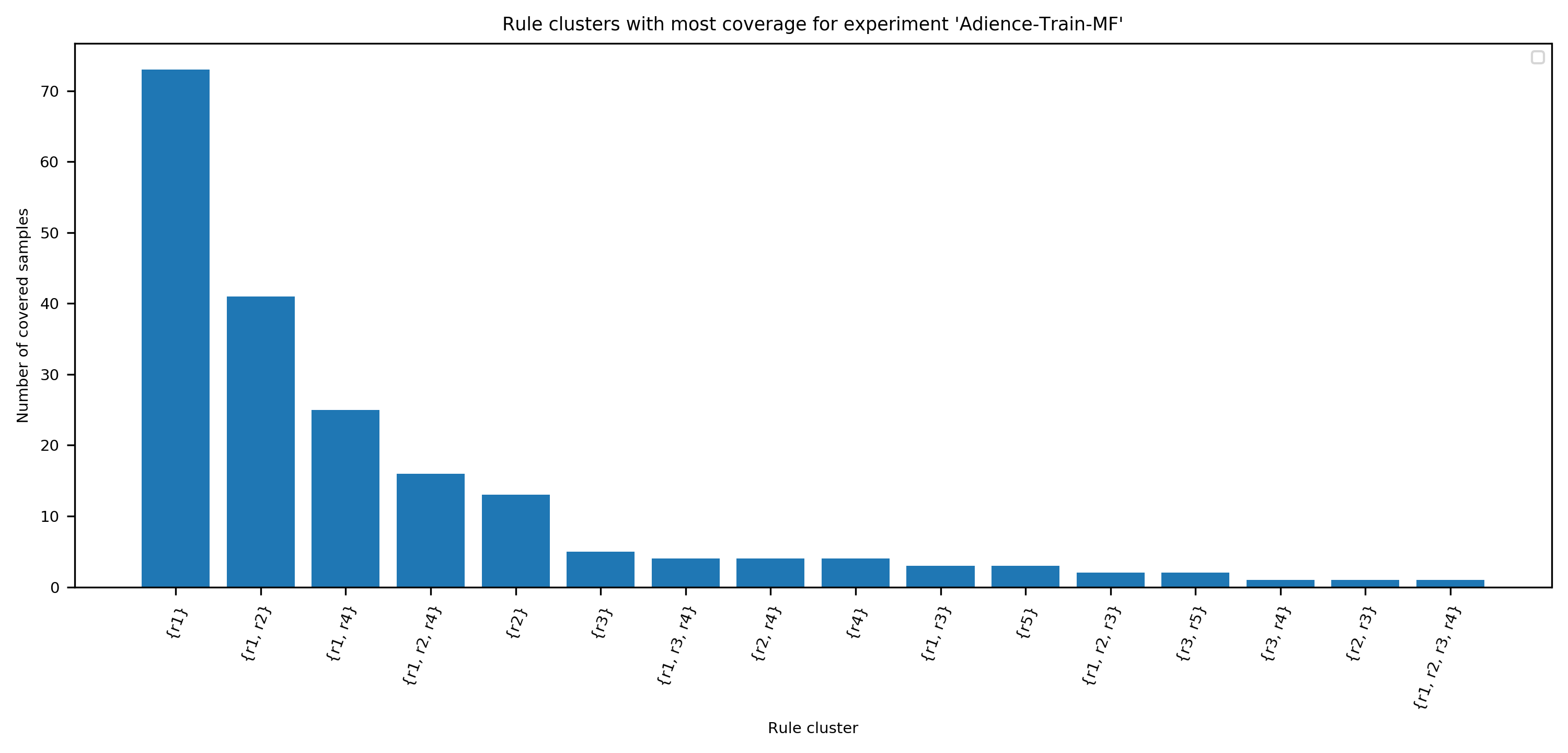}
	\caption{Rule clusters with most coverage for experiment 'Adience-Train-MF'}
	\label{fig:cluster_adience_train_mf}
\end{figure}

\begin{figure}[!h]
	\centering
	\includegraphics[width=1\textwidth]{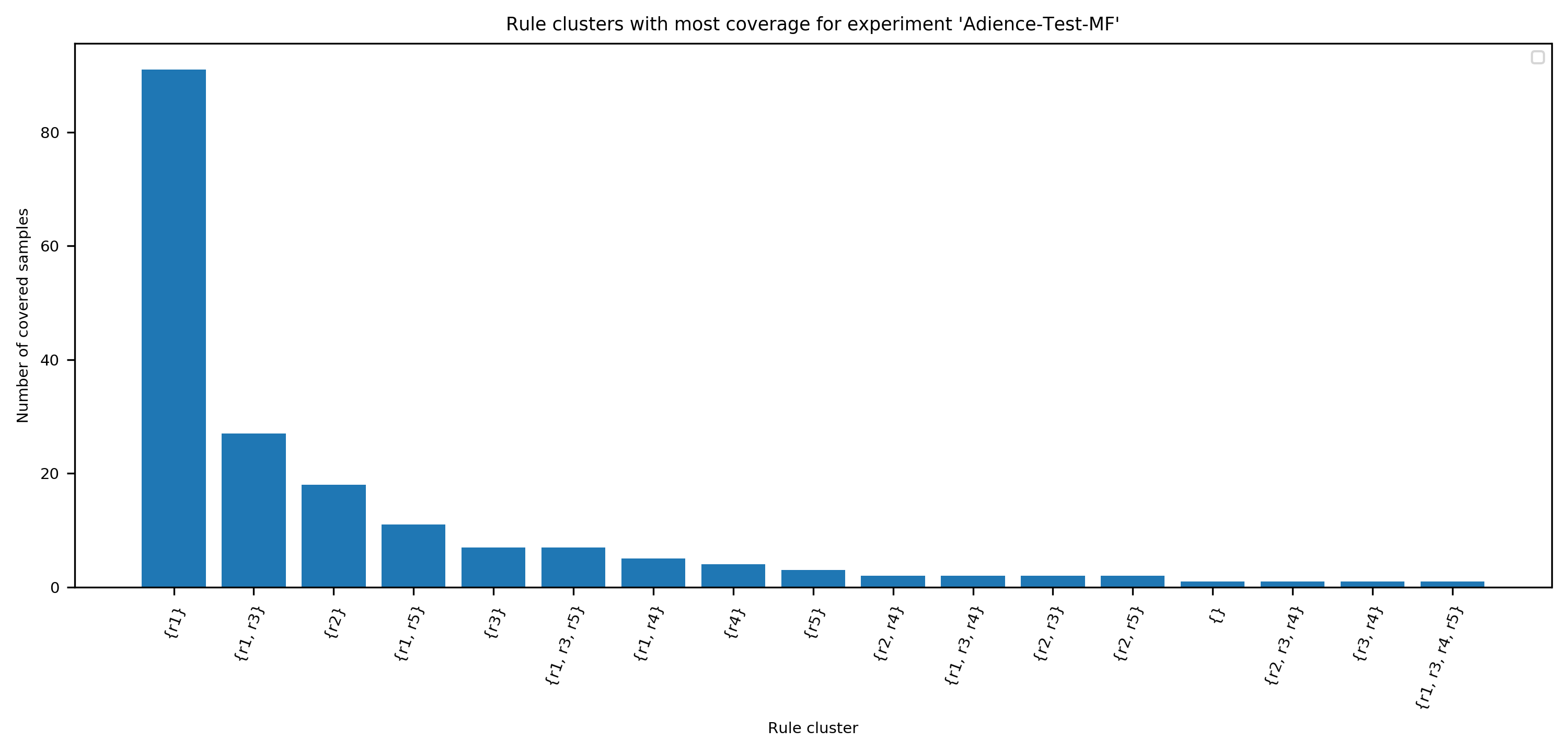}
	\caption{Rule clusters with most coverage for experiment 'Adience-Test-MF'}
	\label{fig:cluster_adience_test_mf}
\end{figure}

\begin{figure}[!h]
	\centering
	\includegraphics[width=1\textwidth]{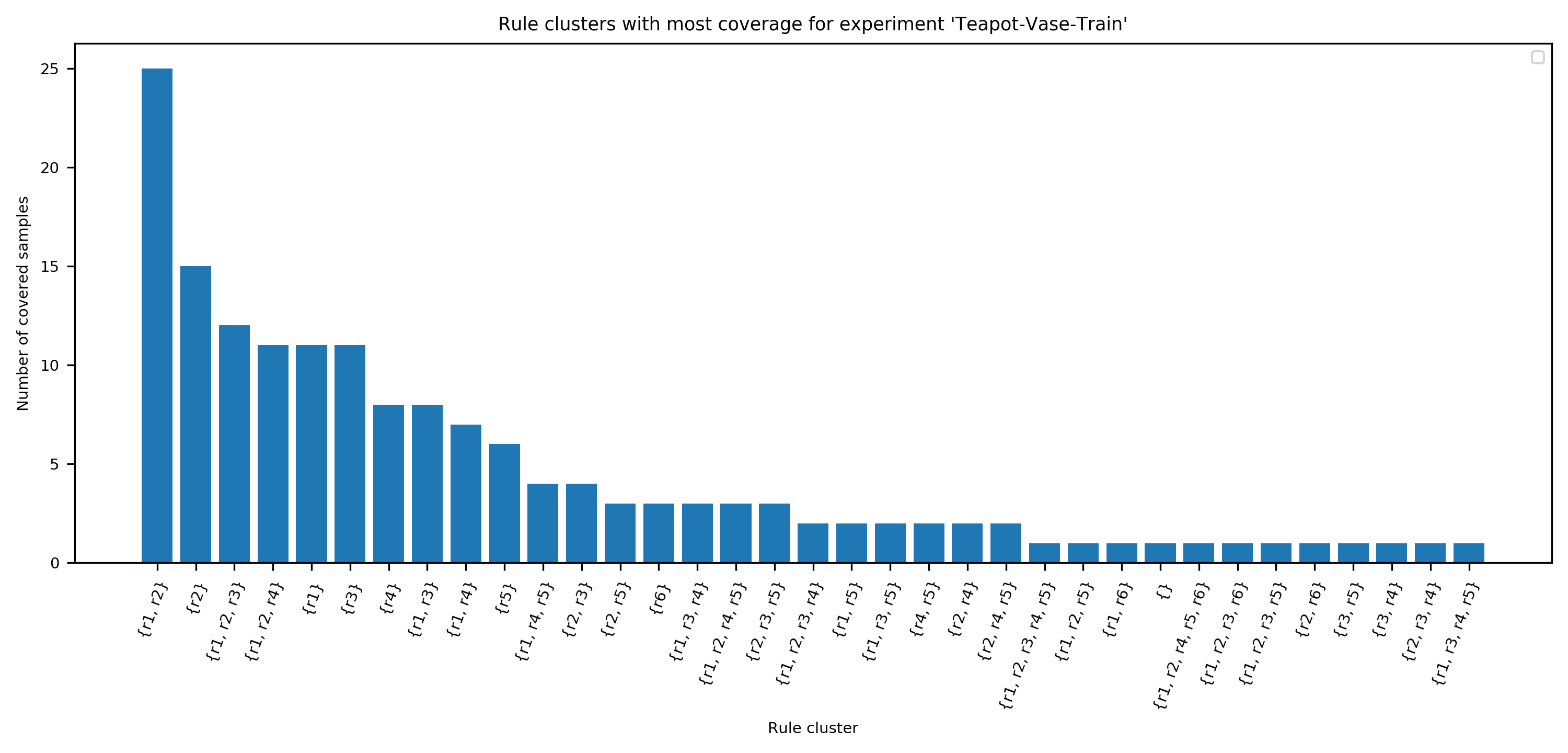}
	\caption{Rule clusters with most coverage for experiment 'Teapot-Vase-Train'}
	\label{fig:cluster_teapot}
\end{figure}

\begin{figure}[!h]
	\centering
	\includegraphics[width=1\textwidth]{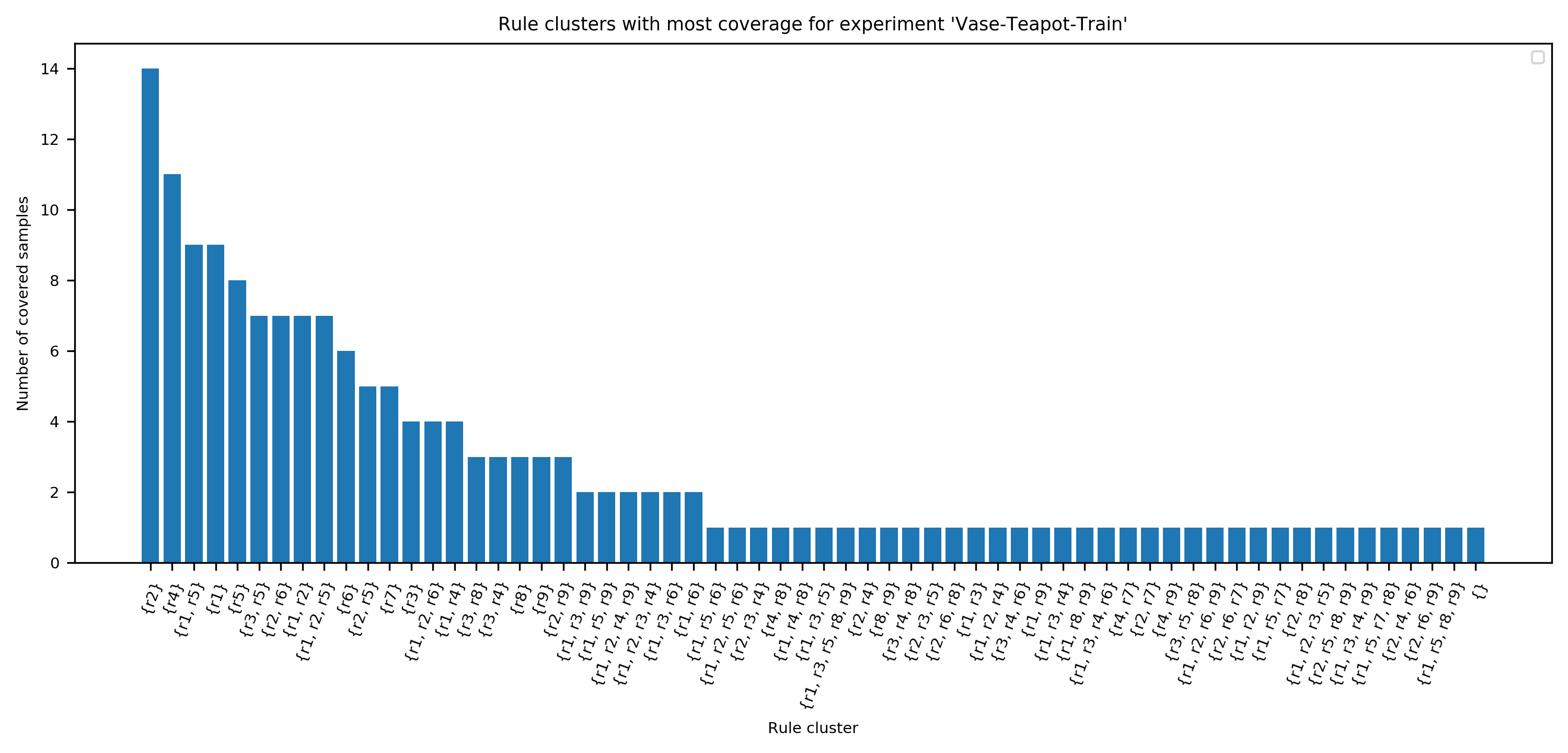}
	\caption{Rule clusters with most coverage for experiment 'Vase-Teapot-Train'}
	\label{fig:cluster_vase}
\end{figure}

\begin{figure}[!h]
	\centering
	\includegraphics[width=1\textwidth]{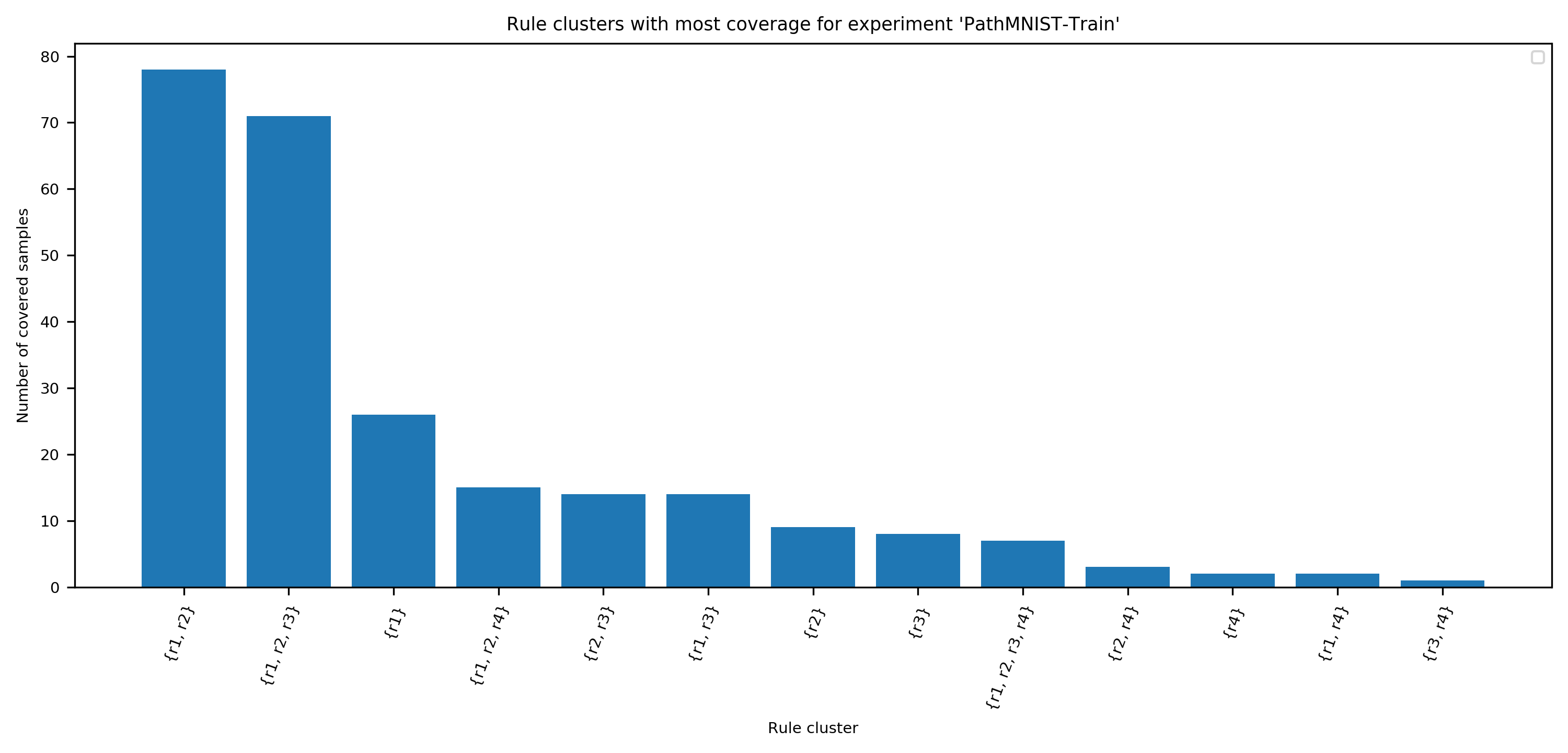}
	\caption{Rule clusters with most coverage for experiment 'PathMNIST-Train'}
	\label{fig:cluster_pathmnist_train}
\end{figure}

\begin{figure}[!h]
	\centering
	\includegraphics[width=1\textwidth]{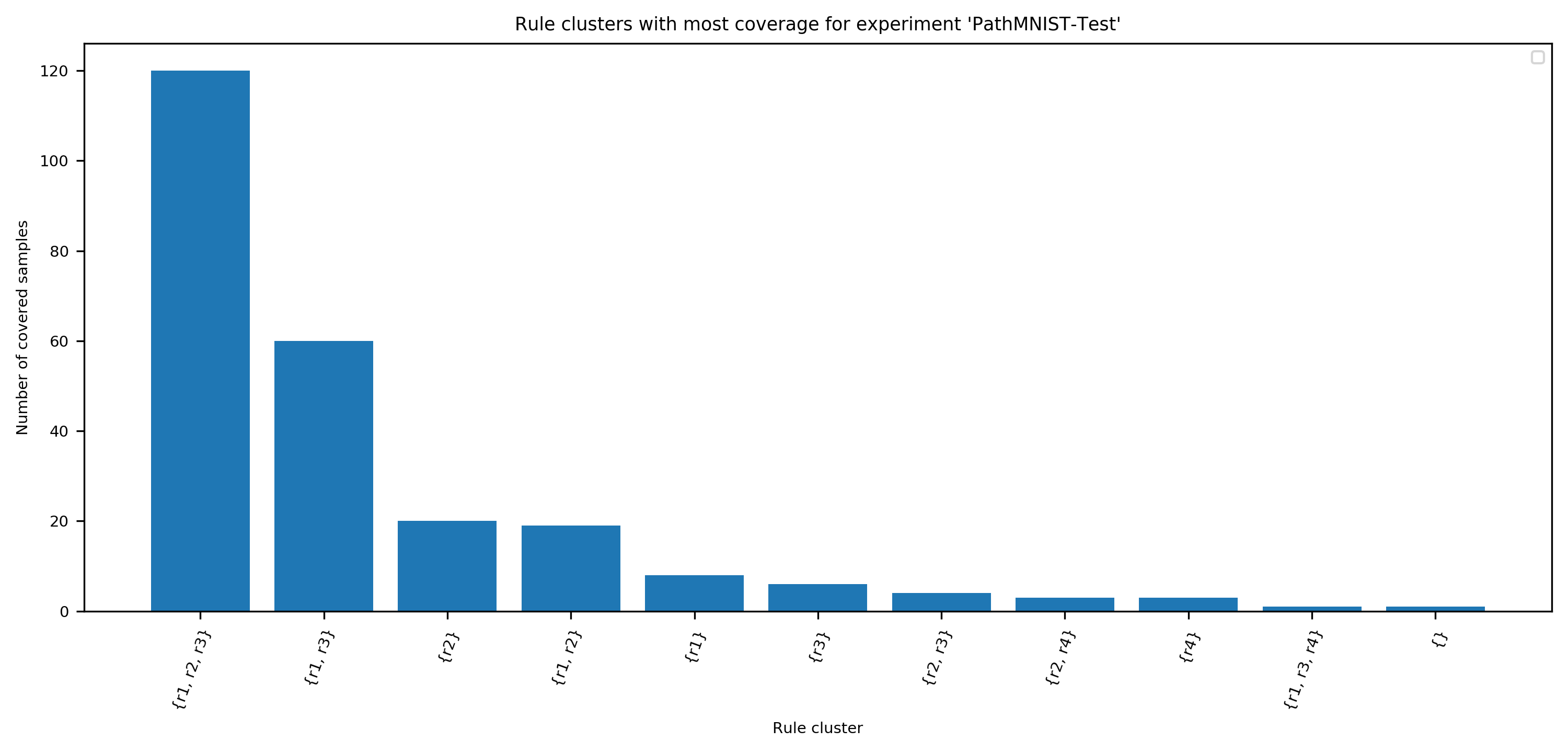}
	\caption{Rule clusters with most coverage for experiment 'PathMNIST-Test'}
	\label{fig:cluster_pathmnist_test}
\end{figure}

In Figures \ref{fig:cluster_picasso_train}, \ref{fig:cluster_picasso_test}, \ref{fig:cluster_adience_train_fm}, \ref{fig:cluster_adience_test_fm}, \ref{fig:cluster_adience_train_mf}, \ref{fig:cluster_adience_test_mf}, \ref{fig:cluster_teapot}, \ref{fig:cluster_vase}, \ref{fig:cluster_pathmnist_train}, \ref{fig:cluster_pathmnist_test}, for all experiments, we listed the rule clusters that covered the most number of samples for the Aleph system. The x axis lists all elements from the power set of rules from the learned theory that had a sample coverage greater than zero. $ri$ refers to the $i$th rule in the theory. The empty set indicates samples not covered by any rule.

\clearpage

\newpage

\section{Frequency of concepts from top-3 rules per concept rank}\label{sec:rankings}

\begin{figure}[!h]
    \centering

    \begin{subfigure}{0.49\linewidth} 
         \centering
         \includegraphics[width=\linewidth]{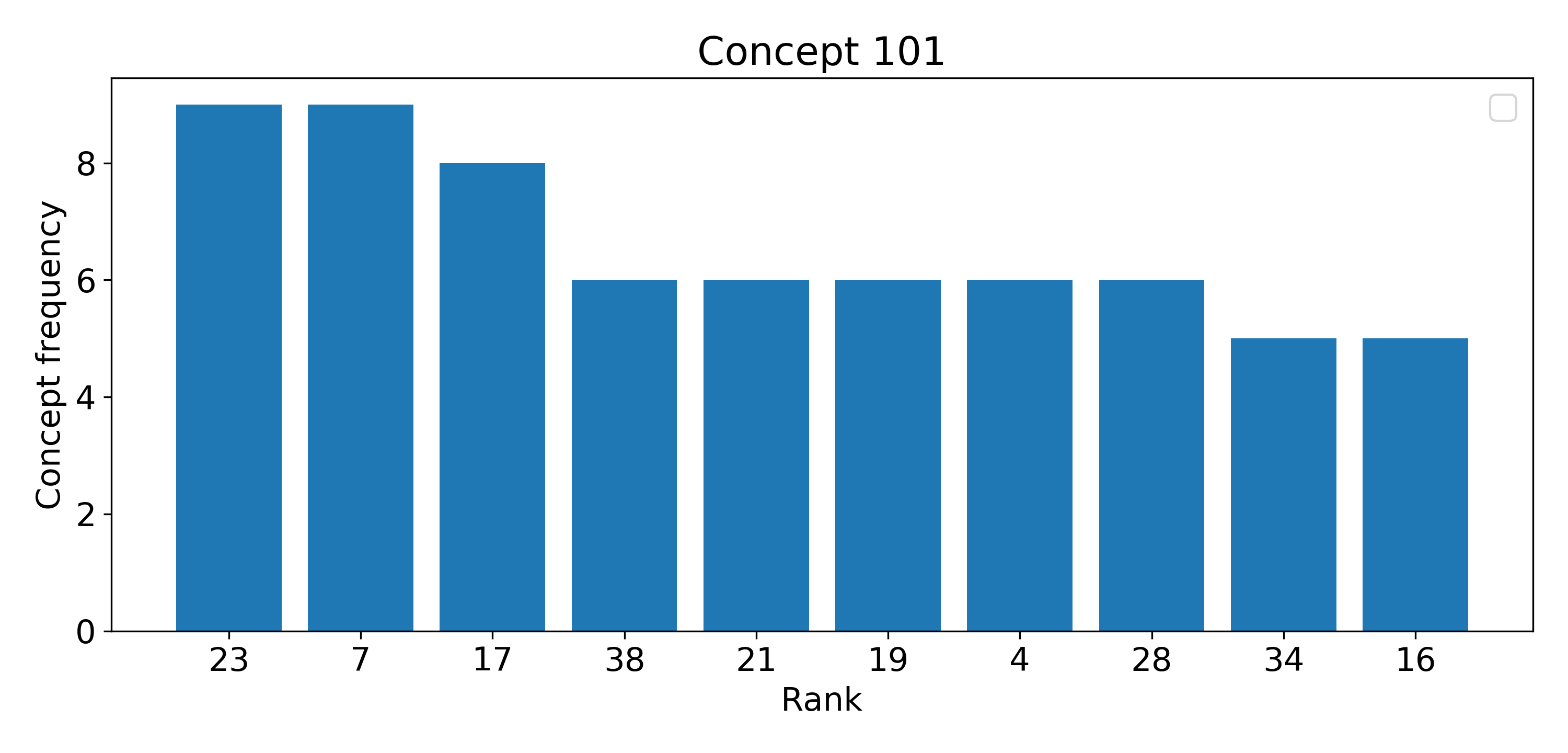}
     \end{subfigure}
    \hfill
    \begin{subfigure}{0.49\linewidth} 
         \centering
         \includegraphics[width=\linewidth]{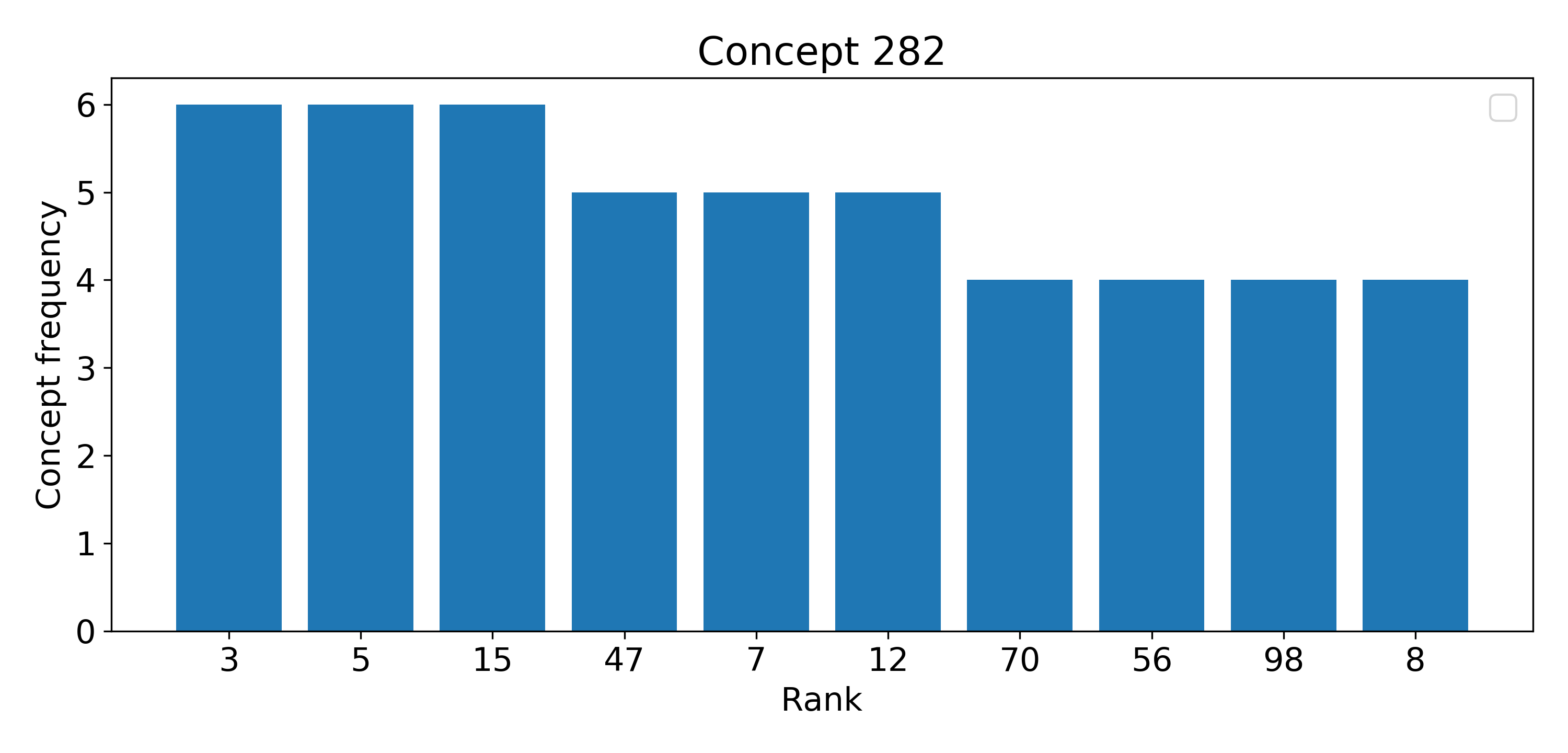}
     \end{subfigure}
     
    \begin{subfigure}{0.49\linewidth} 
         \centering
         \includegraphics[width=\linewidth]{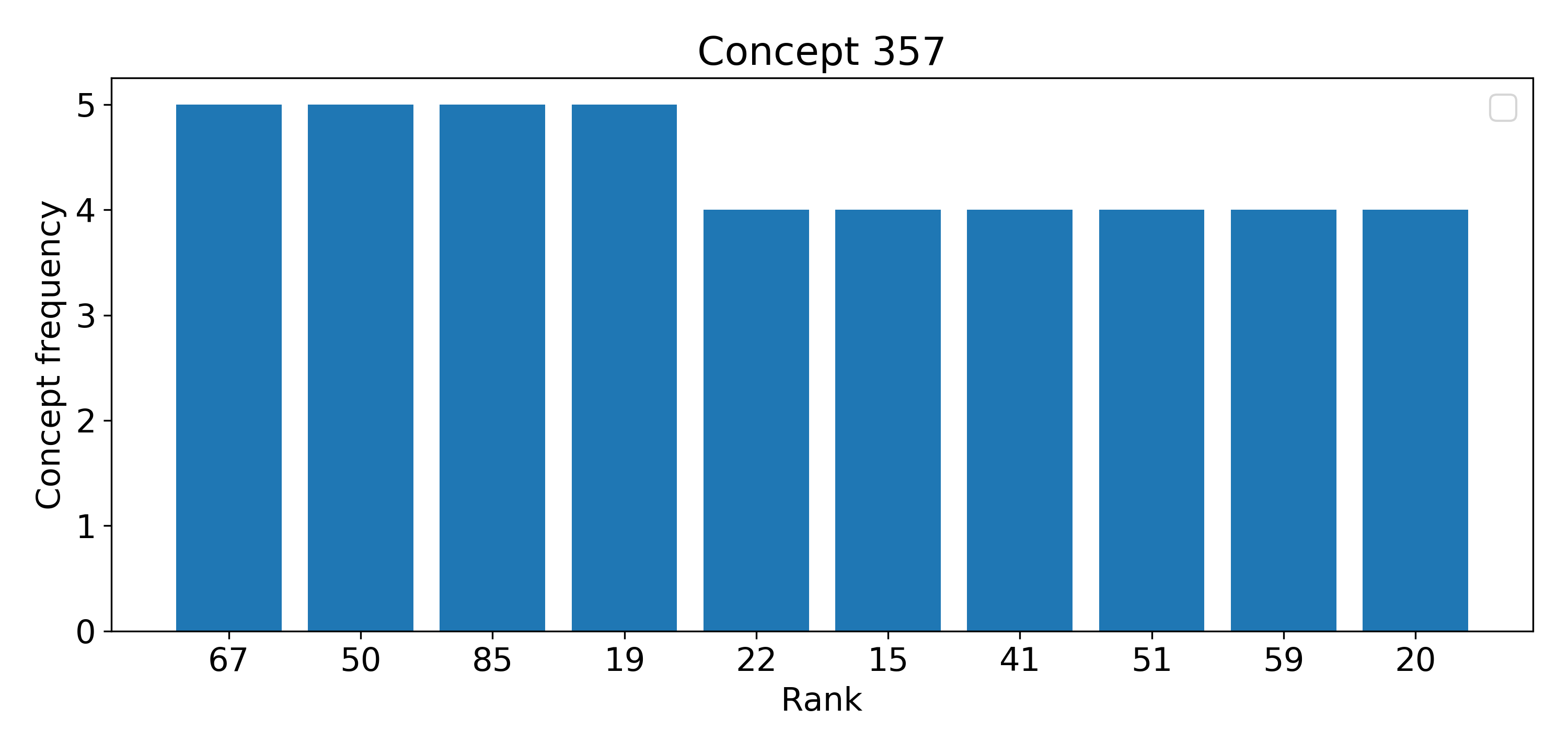}
     \end{subfigure}
    \hfill
    \begin{subfigure}{0.49\linewidth} 
         \centering
         \includegraphics[width=\linewidth]{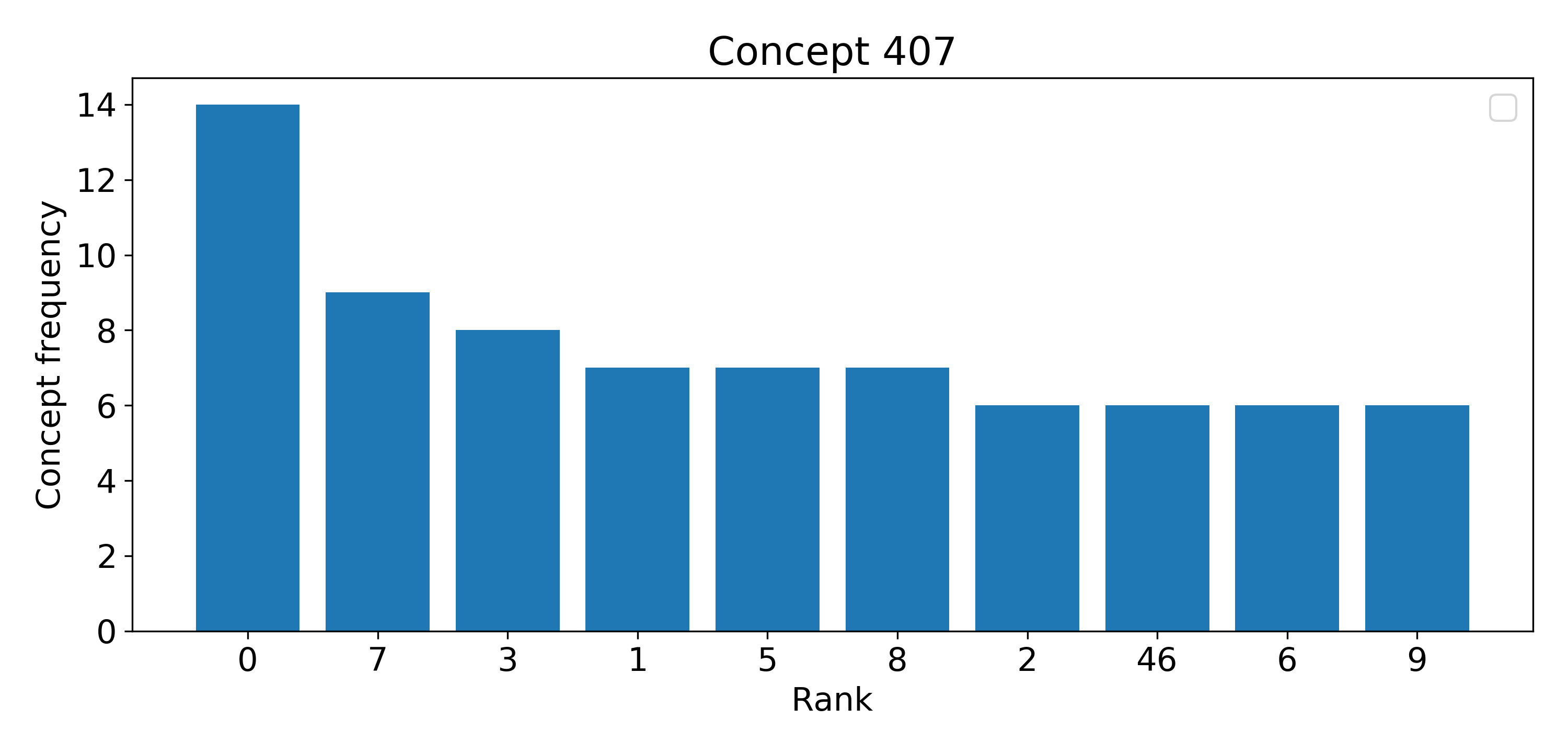}
     \end{subfigure}

    \begin{subfigure}{0.49\linewidth} 
         \centering
         \includegraphics[width=\linewidth]{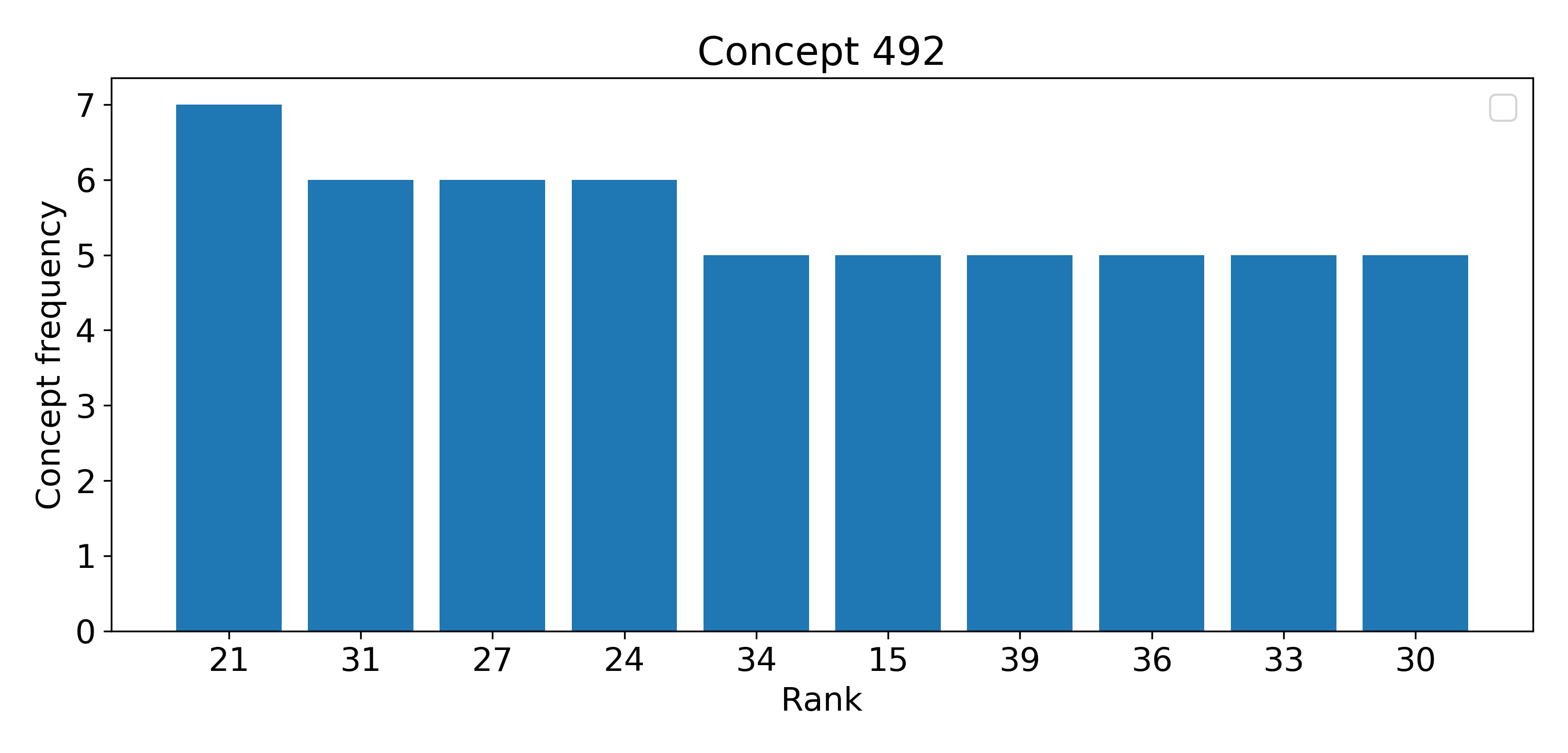}
     \end{subfigure}
    
    \caption{Ranks that are reached the most over all samples for the top-3 rule concepts for experiment 'Picasso-Train'. The 3 rules with the most covered samples (sample count in \textbf{bold}) are:\\
            (\textbf{44}) Face, if concept 357 above of concept 407\\
            (\textbf{43}) Face, if concept 407 is right of concept 101\\
            (\textbf{40}) Face, if concept 492 is below concept 282}
    \label{fig:rankings_picasso_train}
\end{figure}

\begin{figure}[!h]
    \centering

    \begin{subfigure}{0.49\linewidth} 
         \centering
         \includegraphics[width=\linewidth]{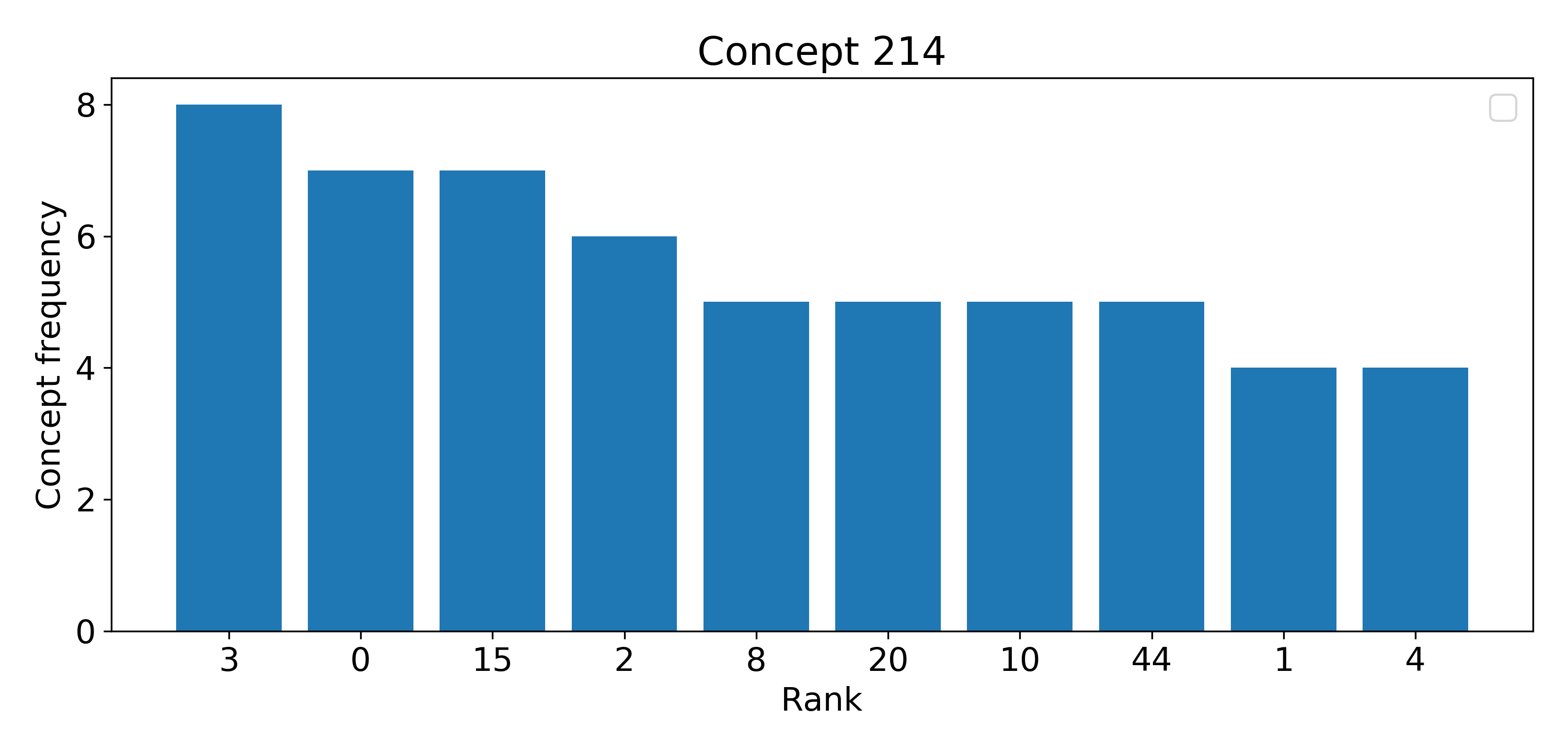}
     \end{subfigure}
    \hfill
    \begin{subfigure}{0.49\linewidth} 
         \centering
         \includegraphics[width=\linewidth]{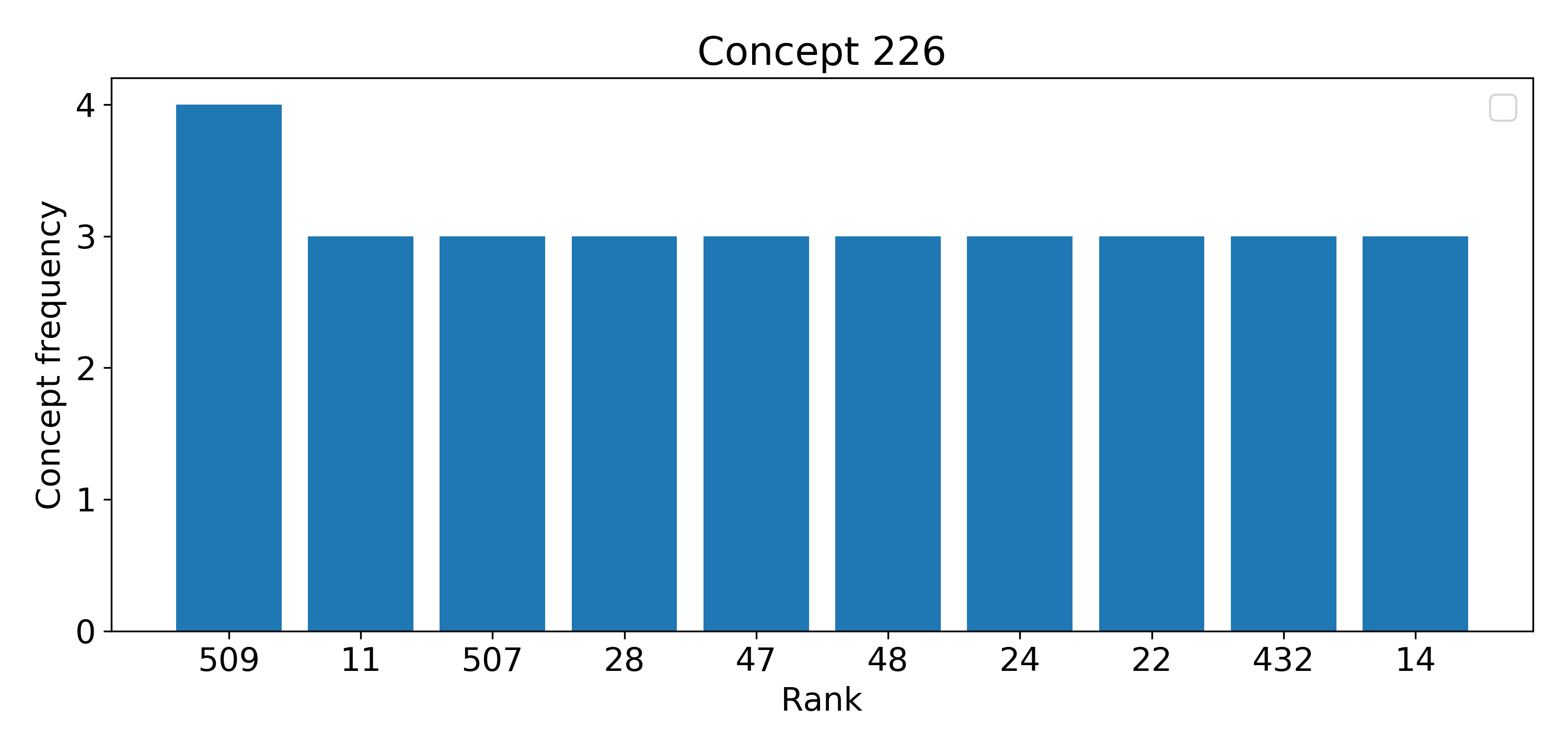}
     \end{subfigure}
     
    \begin{subfigure}{0.49\linewidth} 
         \centering
         \includegraphics[width=\linewidth]{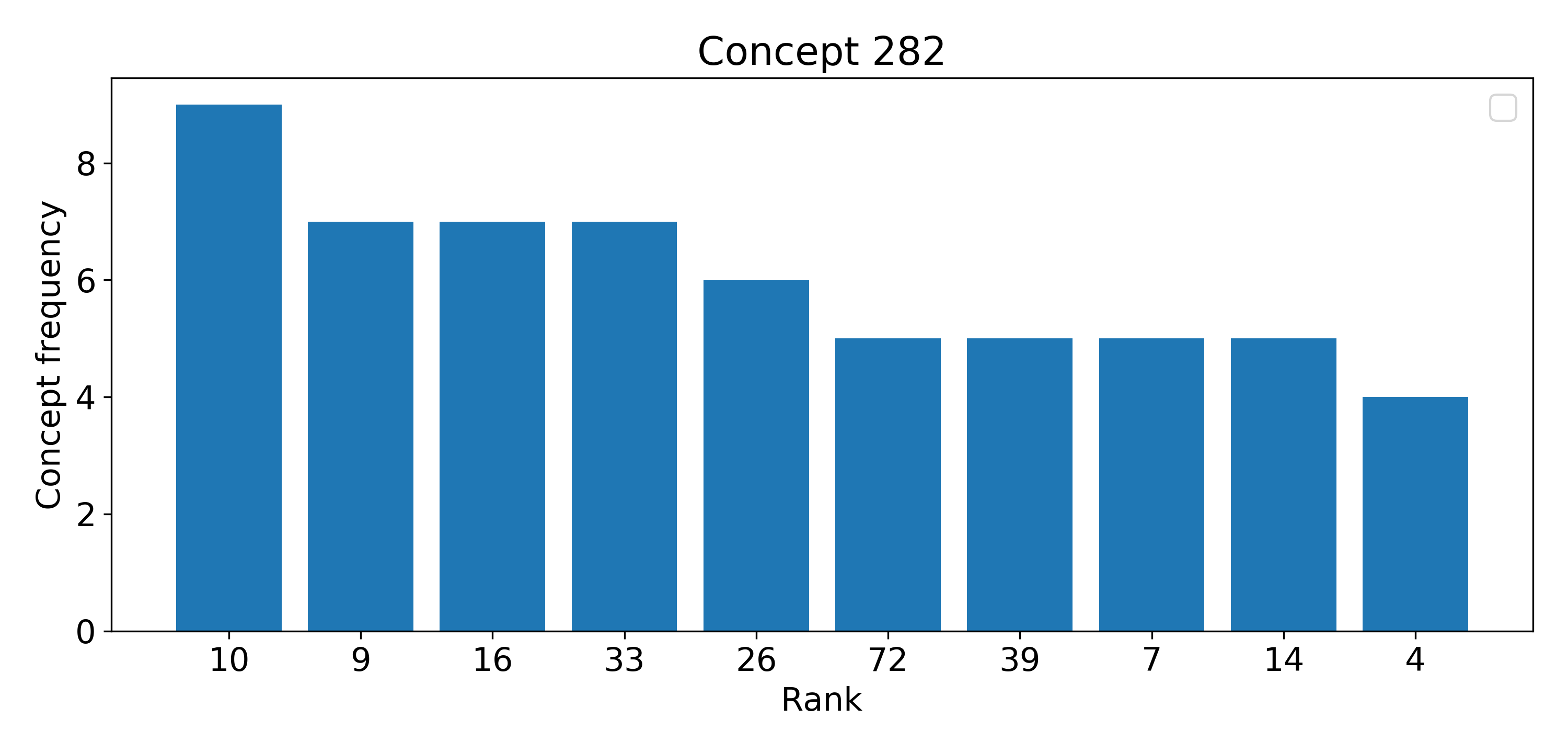}
     \end{subfigure}
    \hfill
    \begin{subfigure}{0.49\linewidth} 
         \centering
         \includegraphics[width=\linewidth]{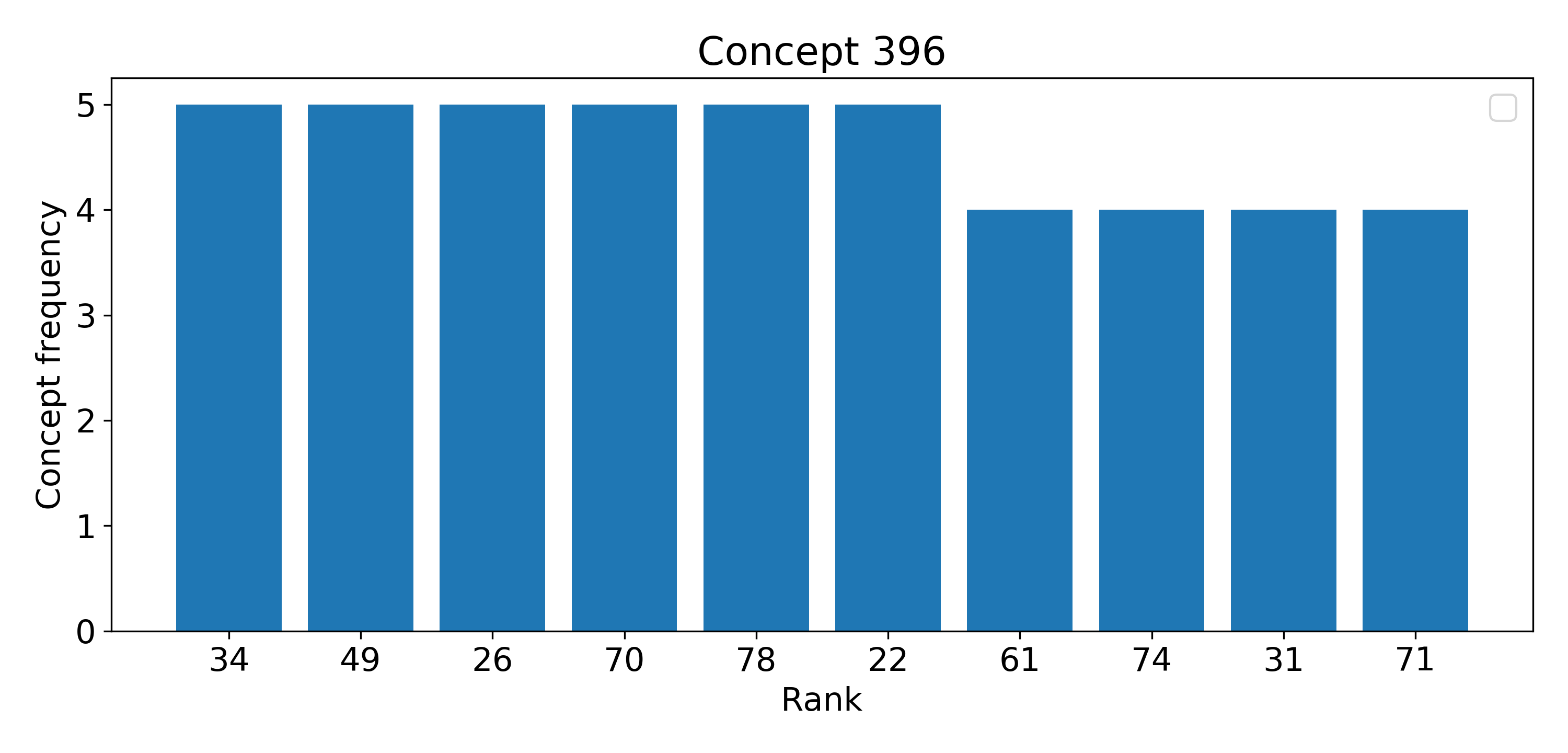}
     \end{subfigure}

    \begin{subfigure}{0.49\linewidth} 
         \centering
         \includegraphics[width=\linewidth]{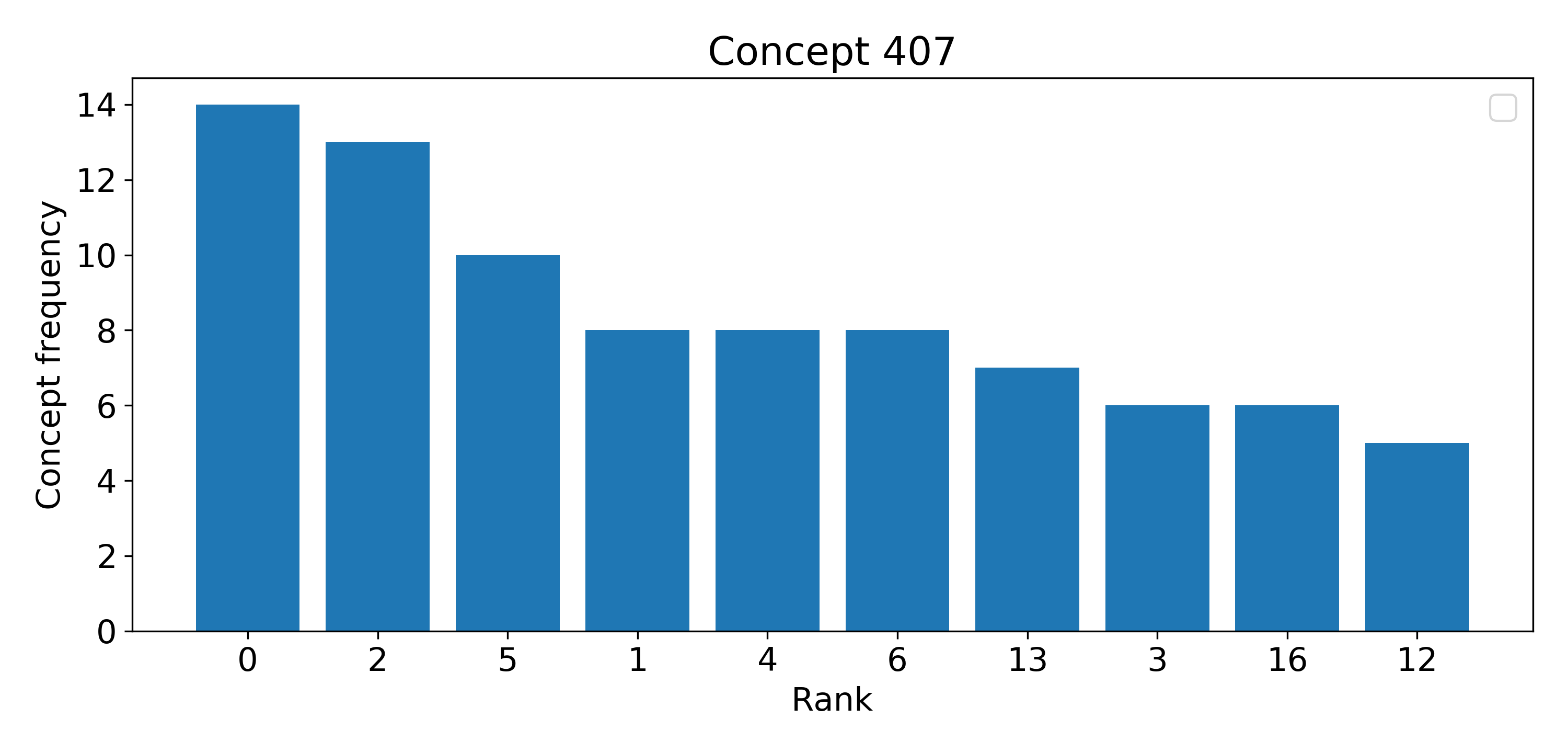}
     \end{subfigure}
     \hfill
    \begin{subfigure}{0.49\linewidth} 
         \centering
         \includegraphics[width=\linewidth]{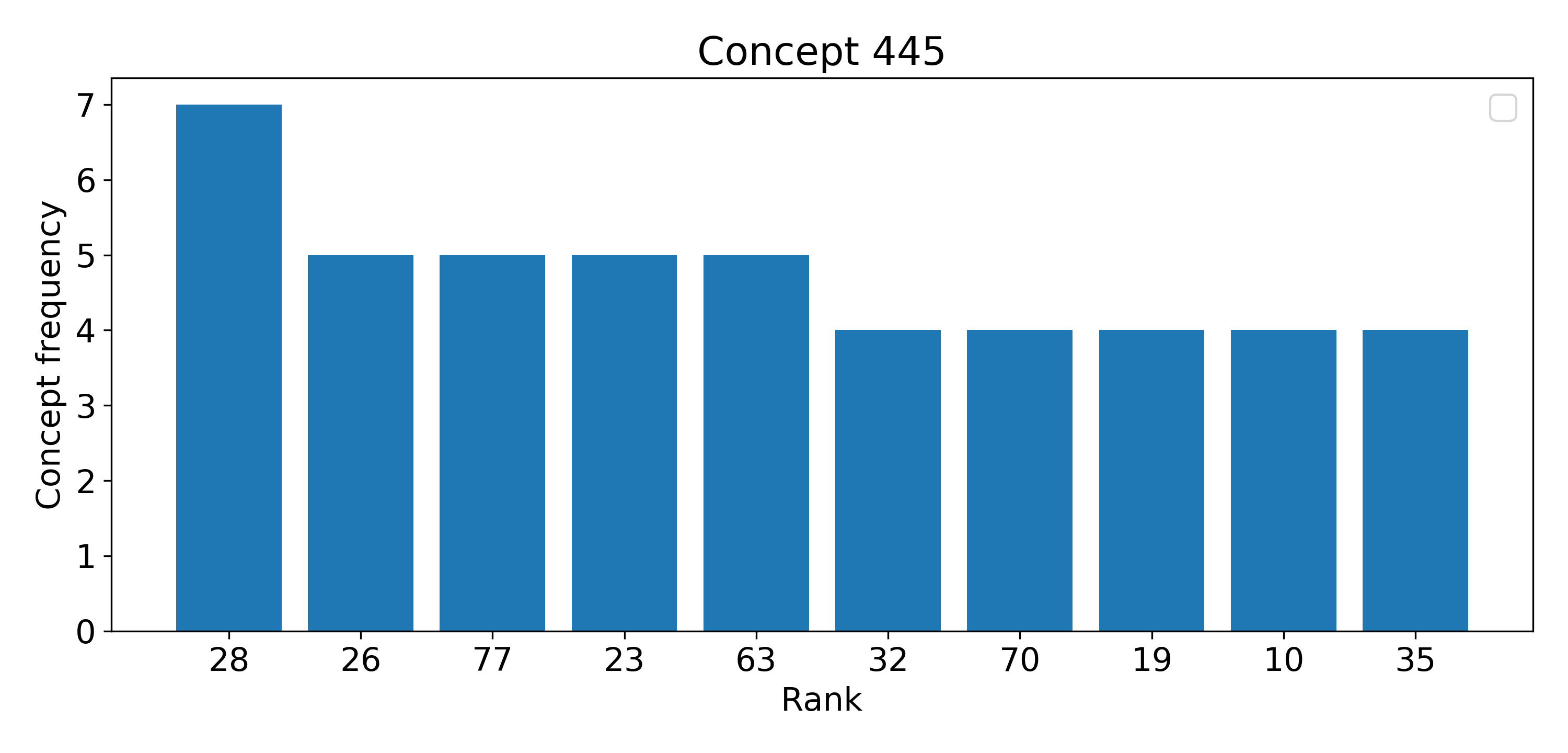}
     \end{subfigure}
    
    \caption{Ranks that are reached the most over all samples for the top-3 rule concepts for experiment 'Picasso-Test'. The 3 rules with the most covered samples (sample count in \textbf{bold}) are:\\
            (\textbf{62}) Face, if concept 282 is above of concept 396\\
            (\textbf{52}) Face, if concept 226 is above of concept 407\\
            (\textbf{40}) Face, if concept 214 is below concept 445}
    \label{fig:rankings_picasso_test}
\end{figure}

\begin{figure}[!h]
    \centering

    \begin{subfigure}{0.49\linewidth} 
         \centering
         \includegraphics[width=\linewidth]{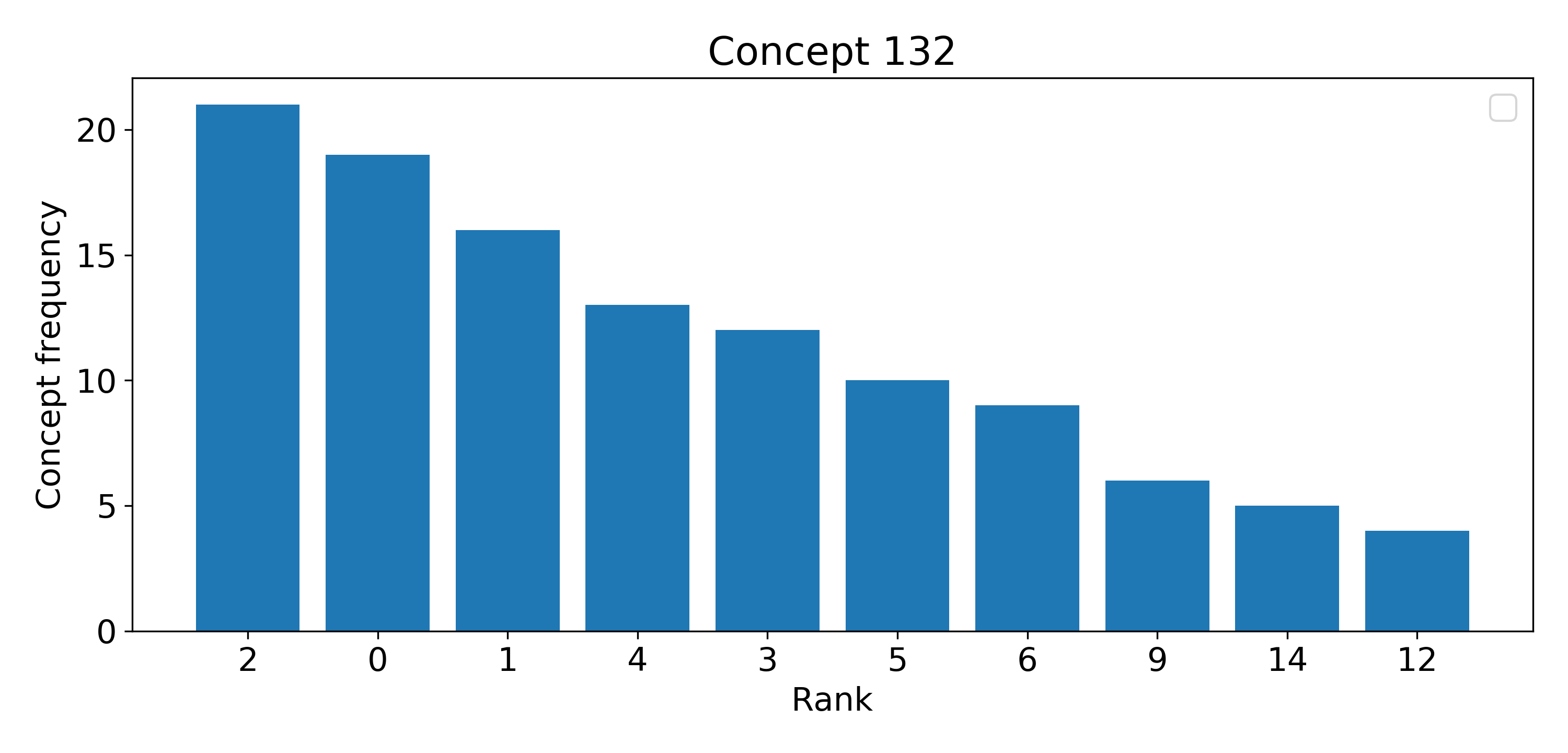}
     \end{subfigure}
    \hfill
    \begin{subfigure}{0.49\linewidth} 
         \centering
         \includegraphics[width=\linewidth]{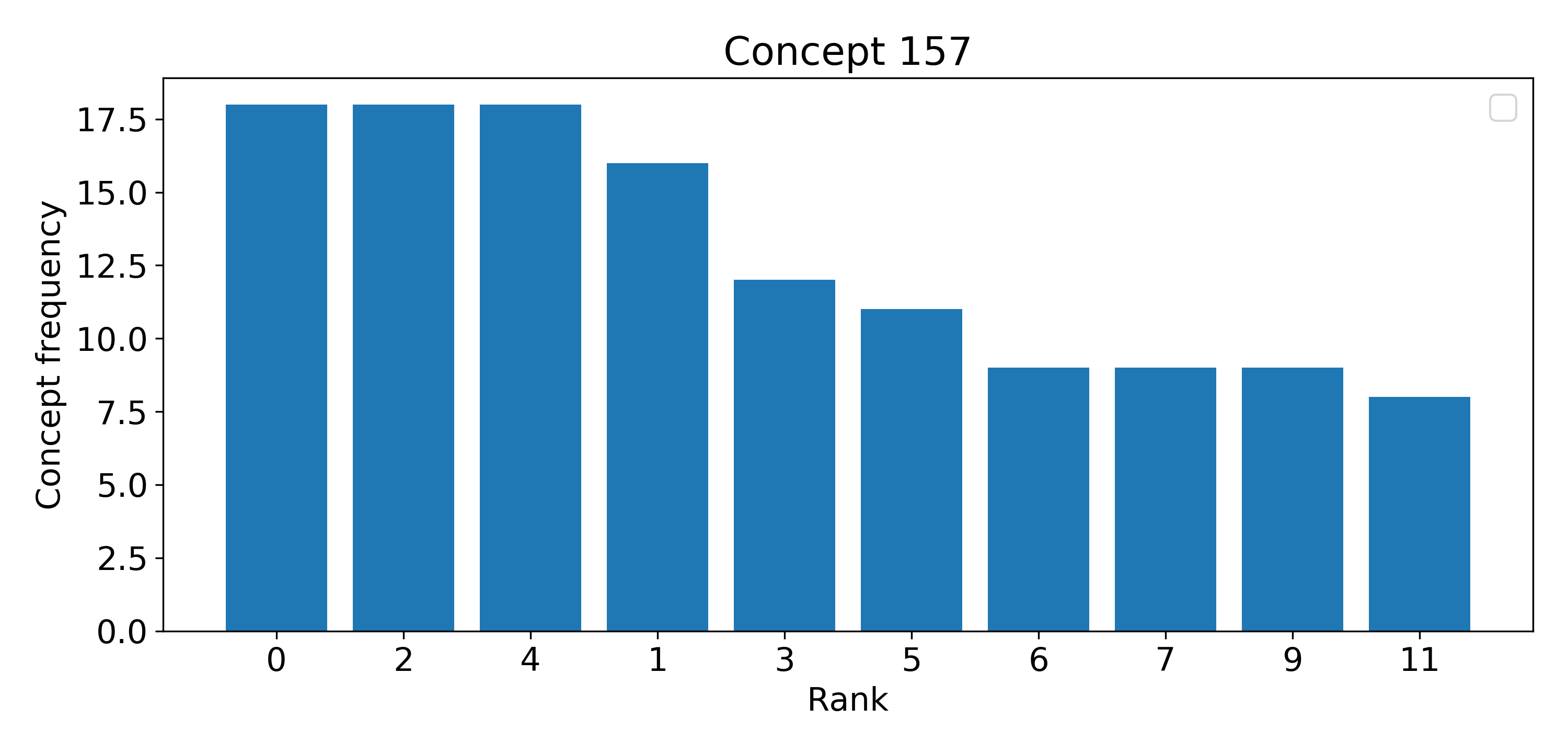}
     \end{subfigure}
     
    \begin{subfigure}{0.49\linewidth} 
         \centering
         \includegraphics[width=\linewidth]{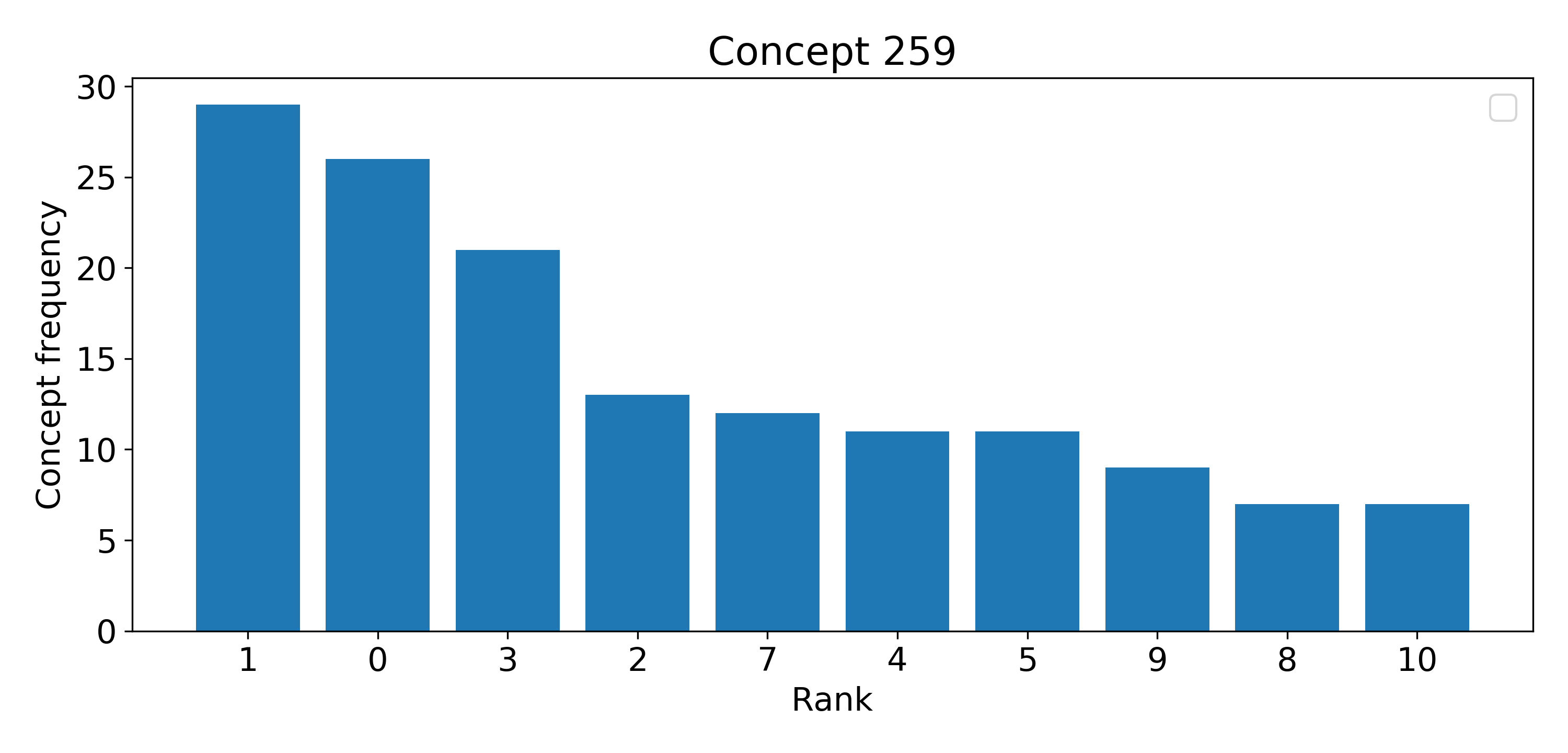}
     \end{subfigure}
    \hfill
    \begin{subfigure}{0.49\linewidth} 
         \centering
         \includegraphics[width=\linewidth]{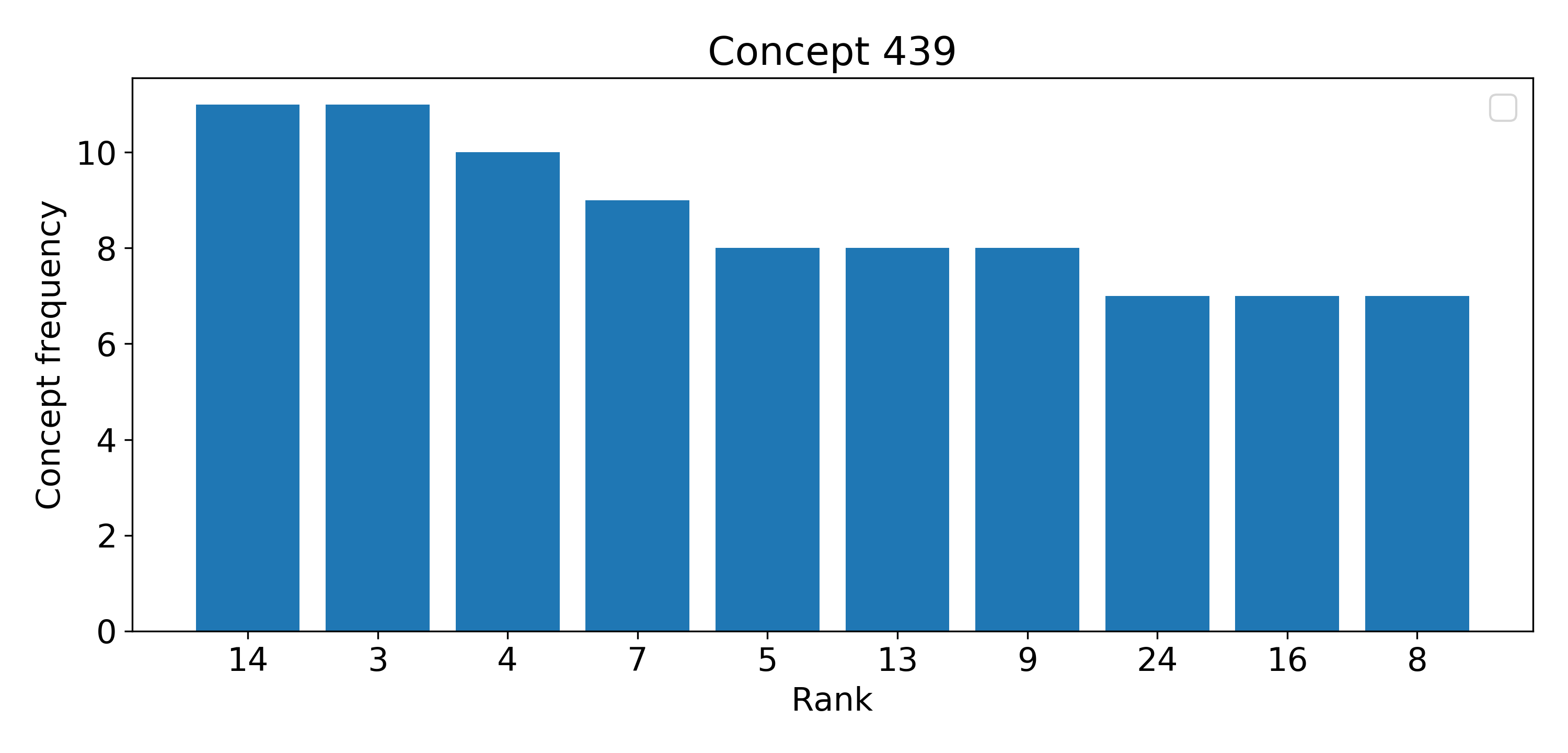}
     \end{subfigure}

    \begin{subfigure}{0.49\linewidth} 
         \centering
         \includegraphics[width=\linewidth]{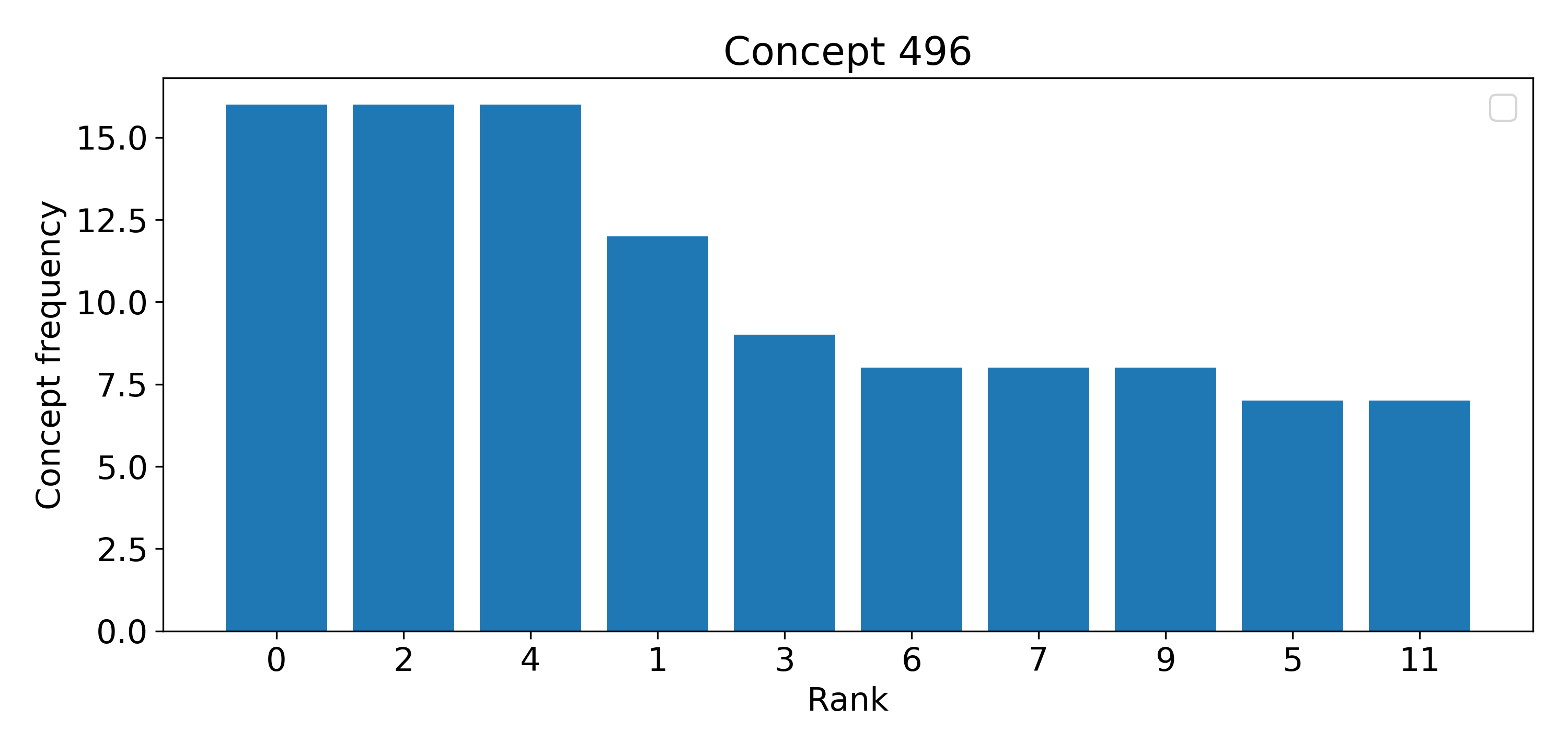}
     \end{subfigure}
    
    \caption{Ranks that are reached the most over all samples for the top-3 rule concepts for experiment 'Adience-Train-FM'. The 3 rules with the most covered samples (sample count in \textbf{bold}) are:\\
            (\textbf{152}) Female, if concept 259 is above of concept 439\\
            (\textbf{148}) Female, if concept 132 is above of concept 496\\
            (\textbf{145}) Female, if concept 157 is above of concept 496}
    \label{fig:rankings_adience_train_fm}
\end{figure}

\begin{figure}[!h]
    \centering

    \begin{subfigure}{0.49\linewidth} 
         \centering
         \includegraphics[width=\linewidth]{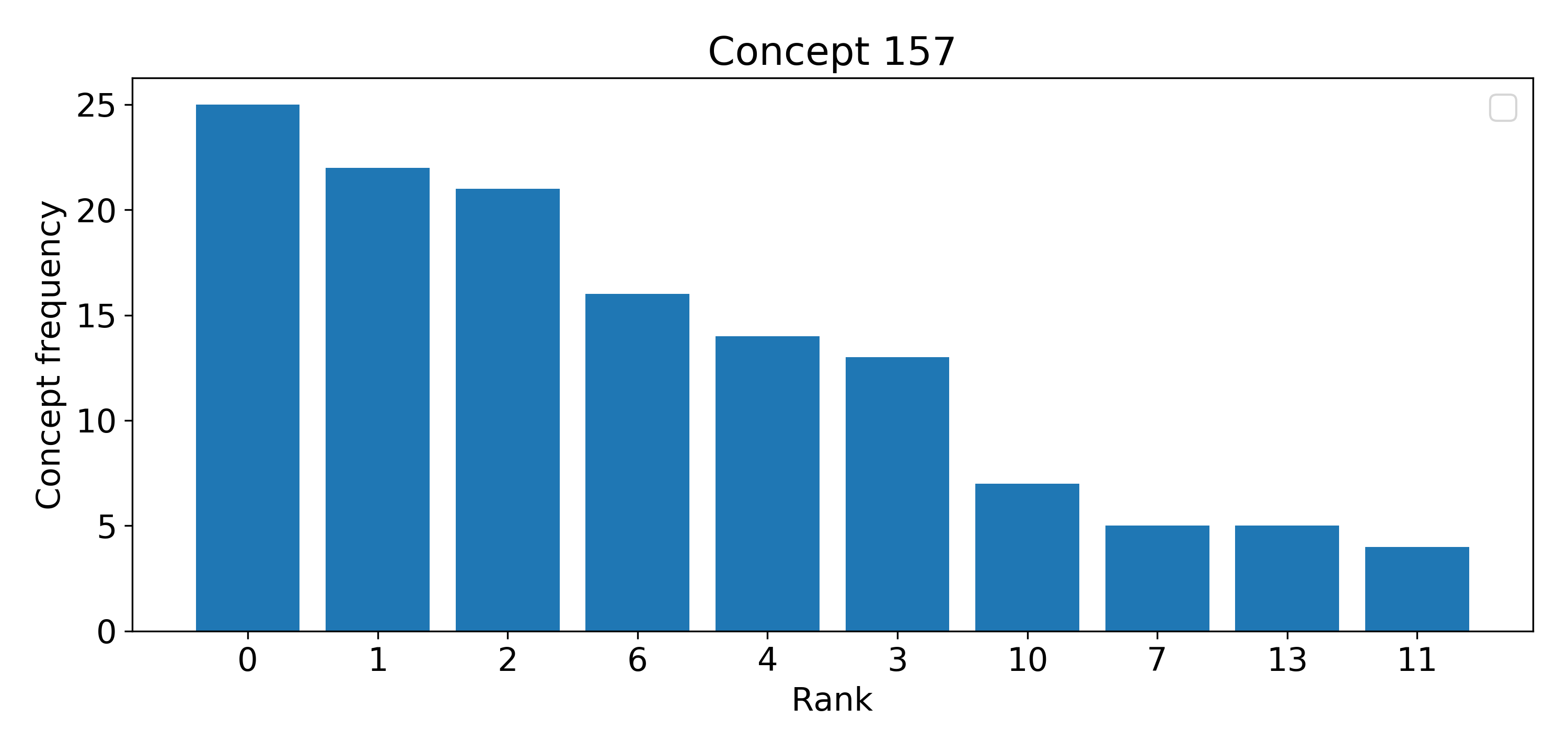}
     \end{subfigure}
    \hfill
    \begin{subfigure}{0.49\linewidth} 
         \centering
         \includegraphics[width=\linewidth]{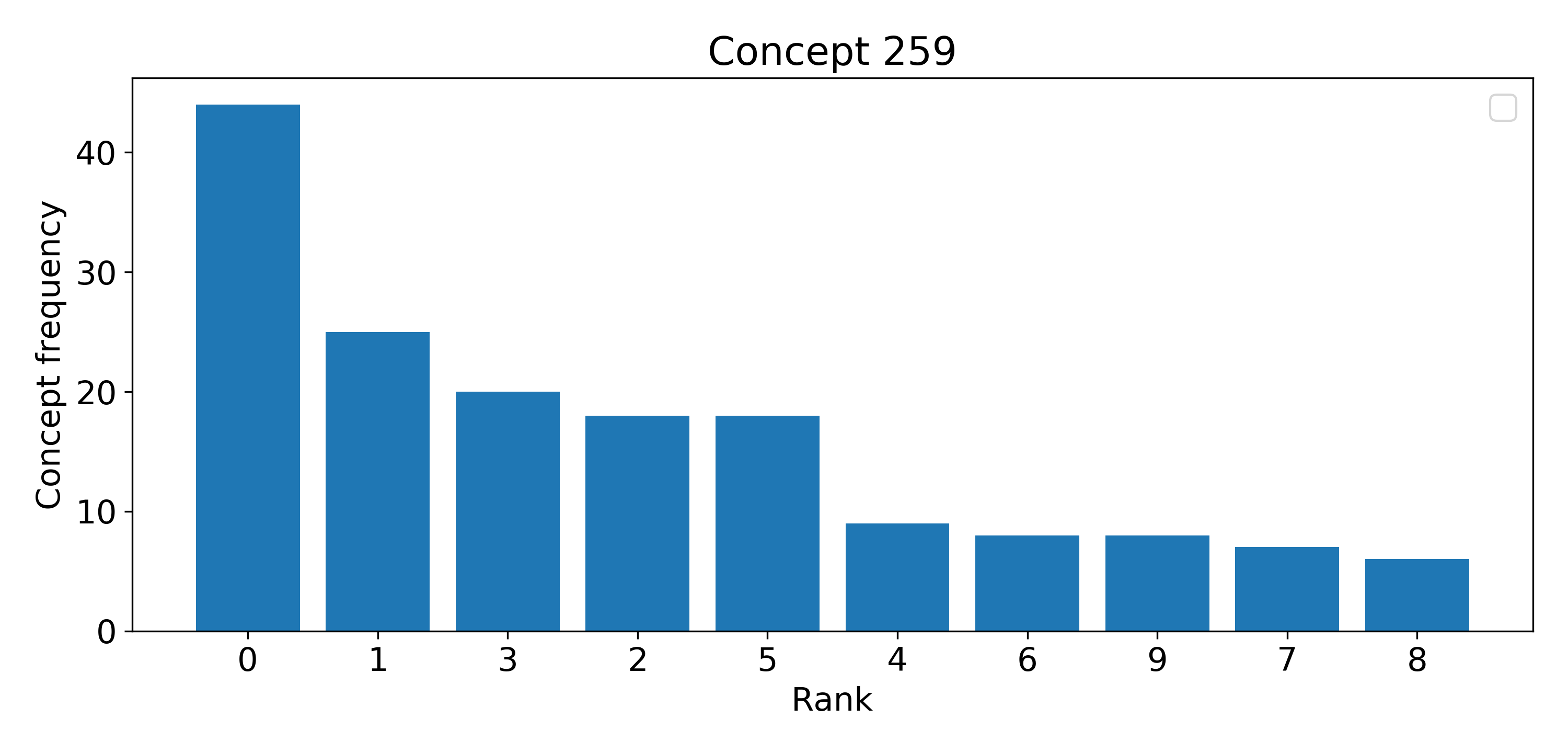}
     \end{subfigure}
     
    \begin{subfigure}{0.49\linewidth} 
         \centering
         \includegraphics[width=\linewidth]{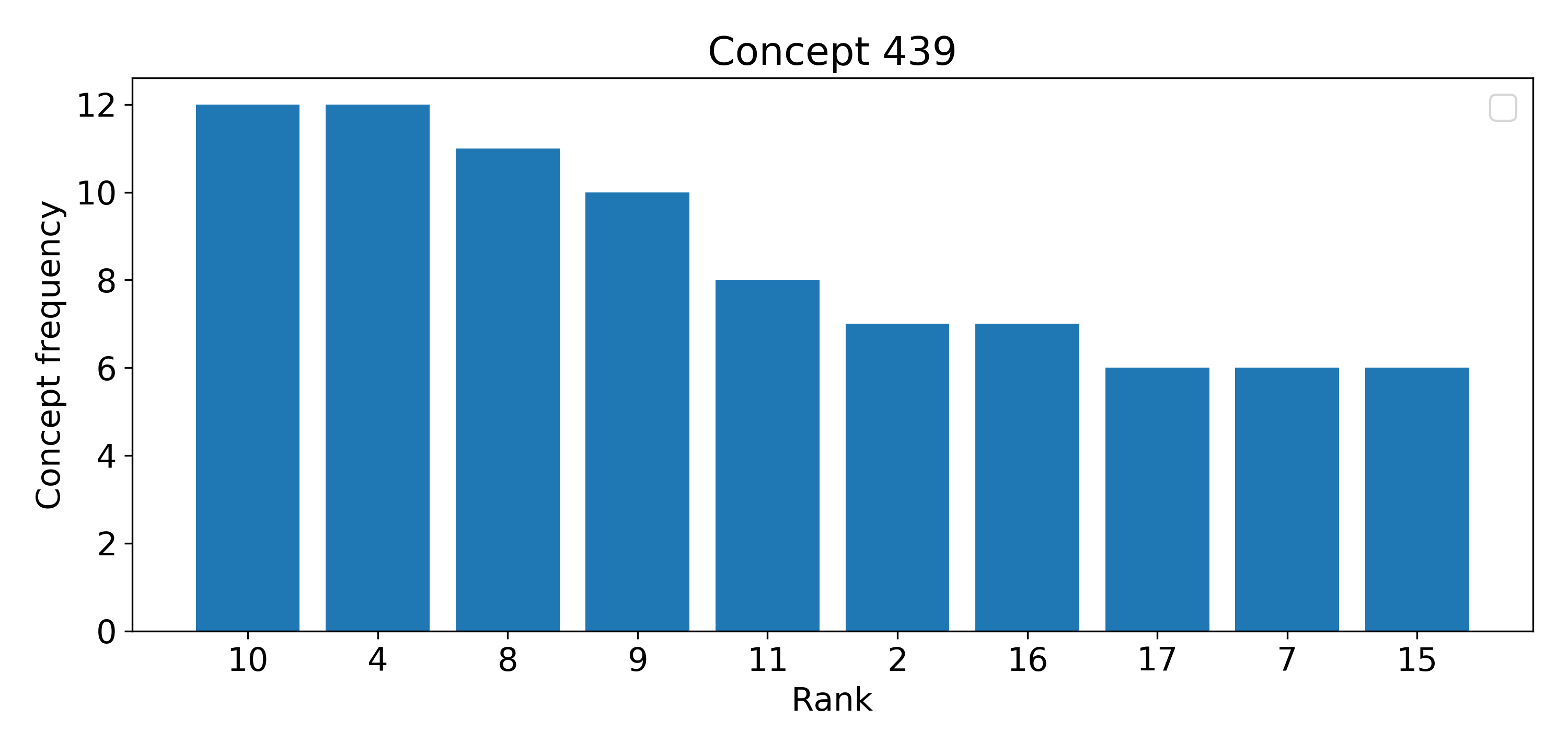}
     \end{subfigure}
    \hfill
    \begin{subfigure}{0.49\linewidth} 
         \centering
         \includegraphics[width=\linewidth]{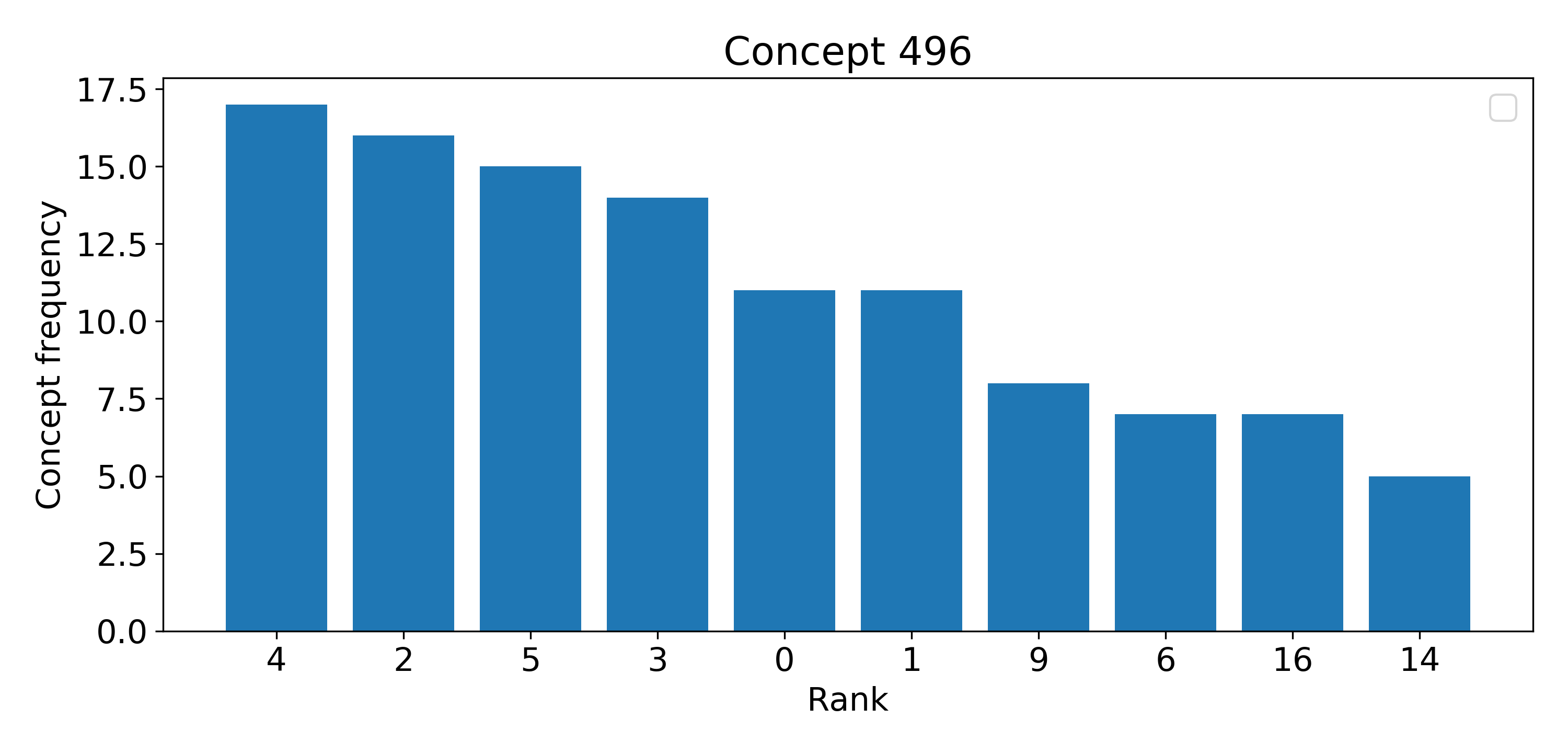}
     \end{subfigure}
    
    \caption{Ranks that are reached the most over all samples for the top-3 rule concepts for experiment 'Adience-Test-FM'. The 3 rules with the most covered samples (sample count in \textbf{bold}) are:\\
            (\textbf{169}) Female, if concept 259 is above of concept 496\\
            (\textbf{132}) Female, if concept 157 is left of concept 259\\
            (\textbf{118}) Female, if concept 157 is above of concept 439}
    \label{fig:rankings_adience_test_fm}
\end{figure}

\begin{figure}[!h]
    \centering

    \begin{subfigure}{0.49\linewidth} 
         \centering
         \includegraphics[width=\linewidth]{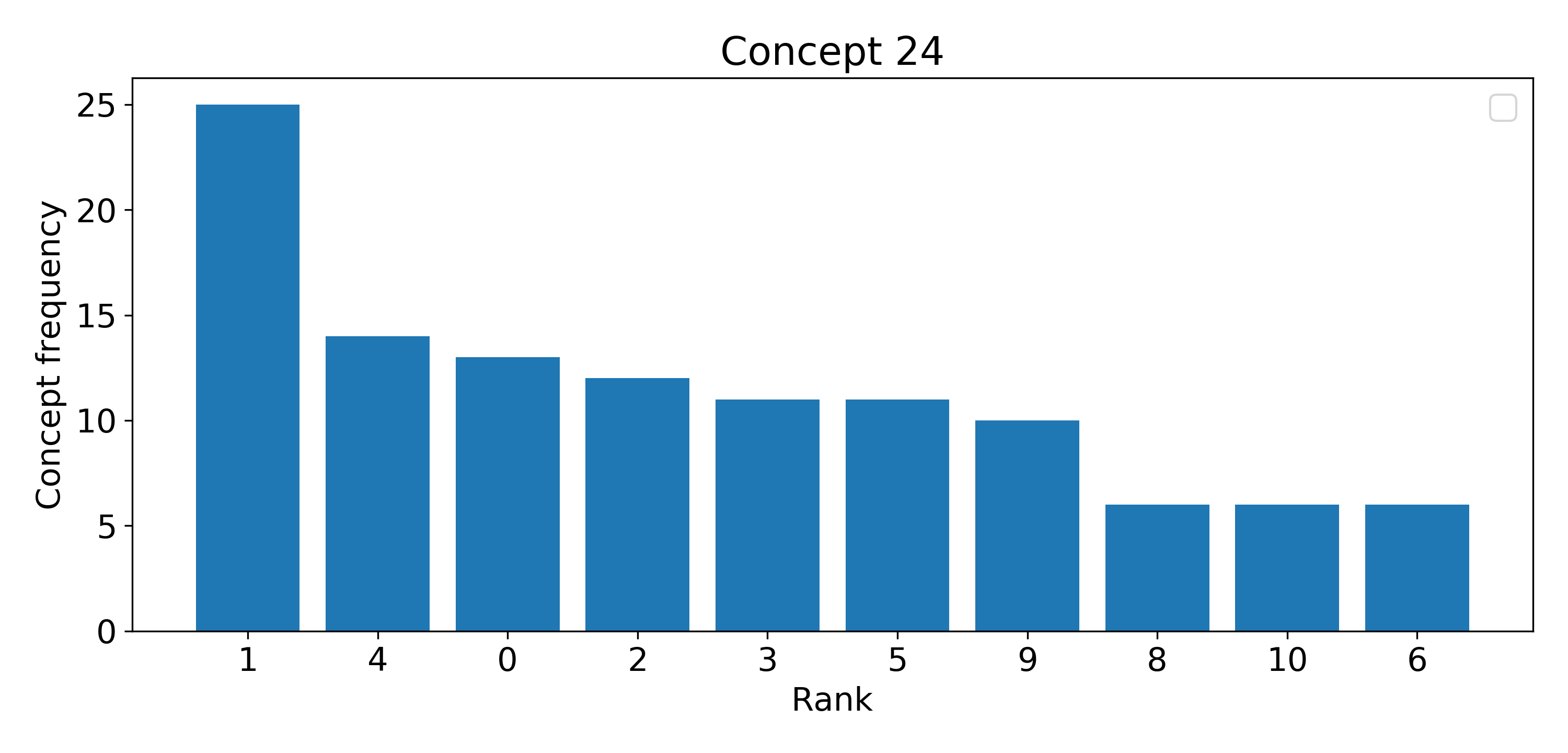}
     \end{subfigure}
    \hfill
    \begin{subfigure}{0.49\linewidth} 
         \centering
         \includegraphics[width=\linewidth]{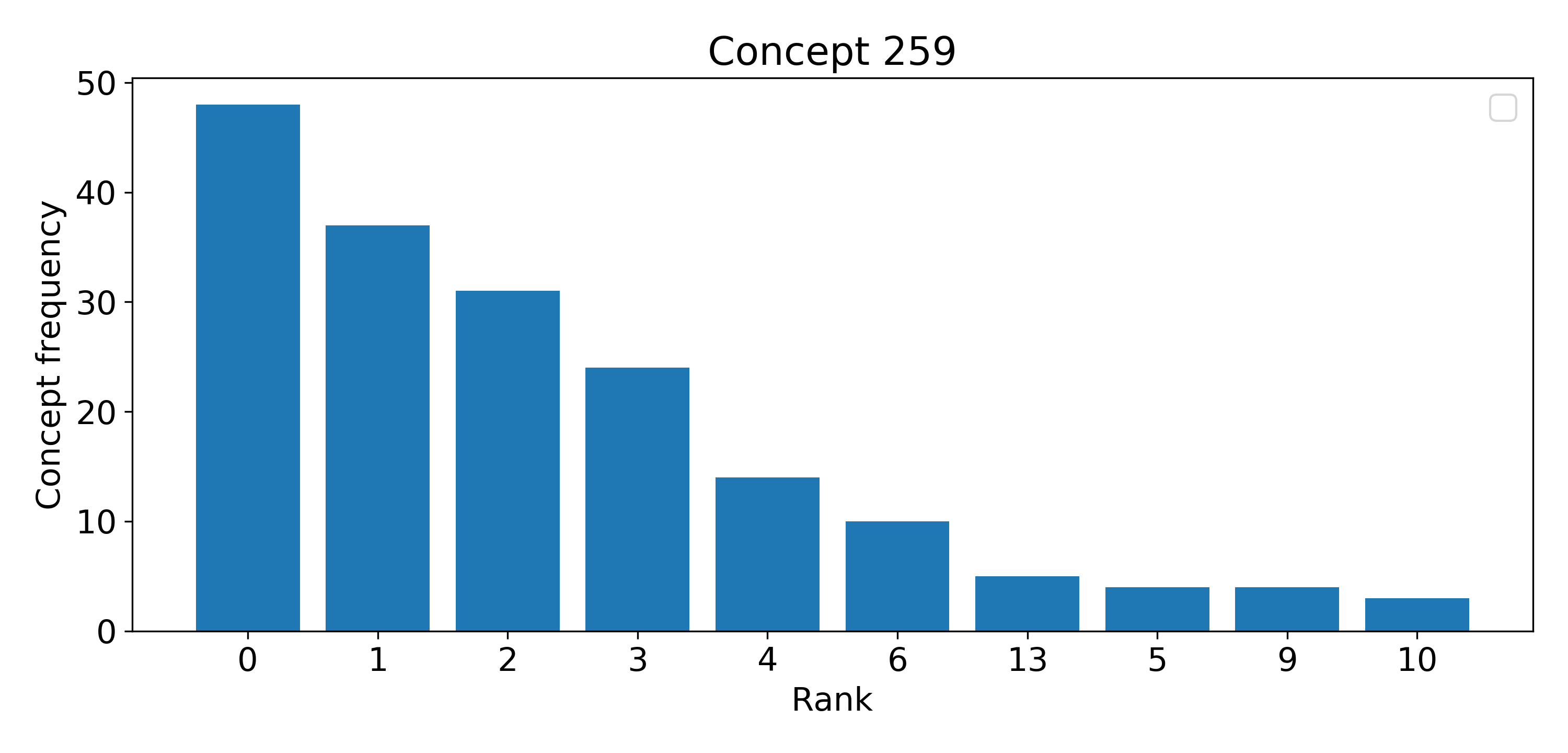}
     \end{subfigure}
     
    \begin{subfigure}{0.49\linewidth} 
         \centering
         \includegraphics[width=\linewidth]{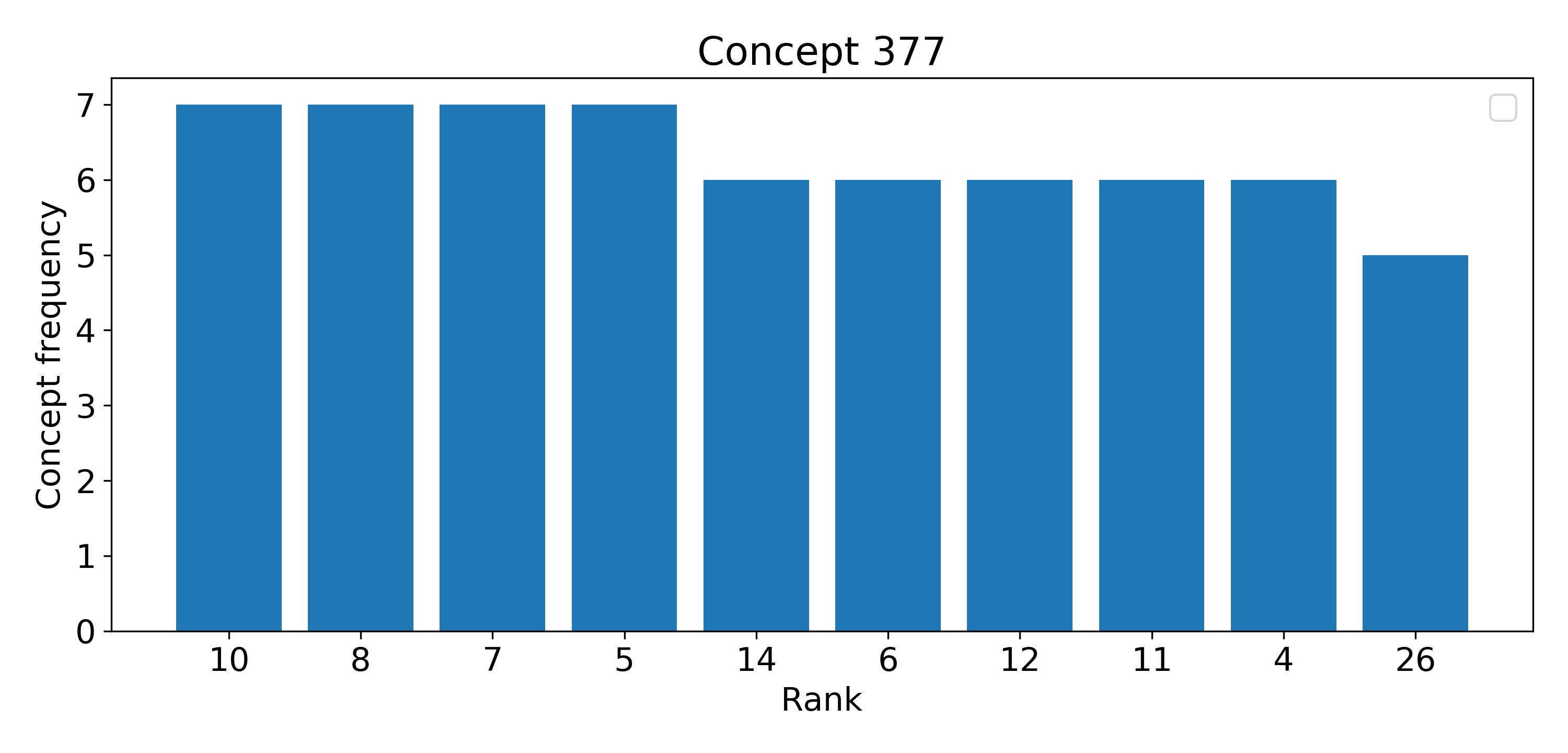}
     \end{subfigure}
    \hfill
    \begin{subfigure}{0.49\linewidth} 
         \centering
         \includegraphics[width=\linewidth]{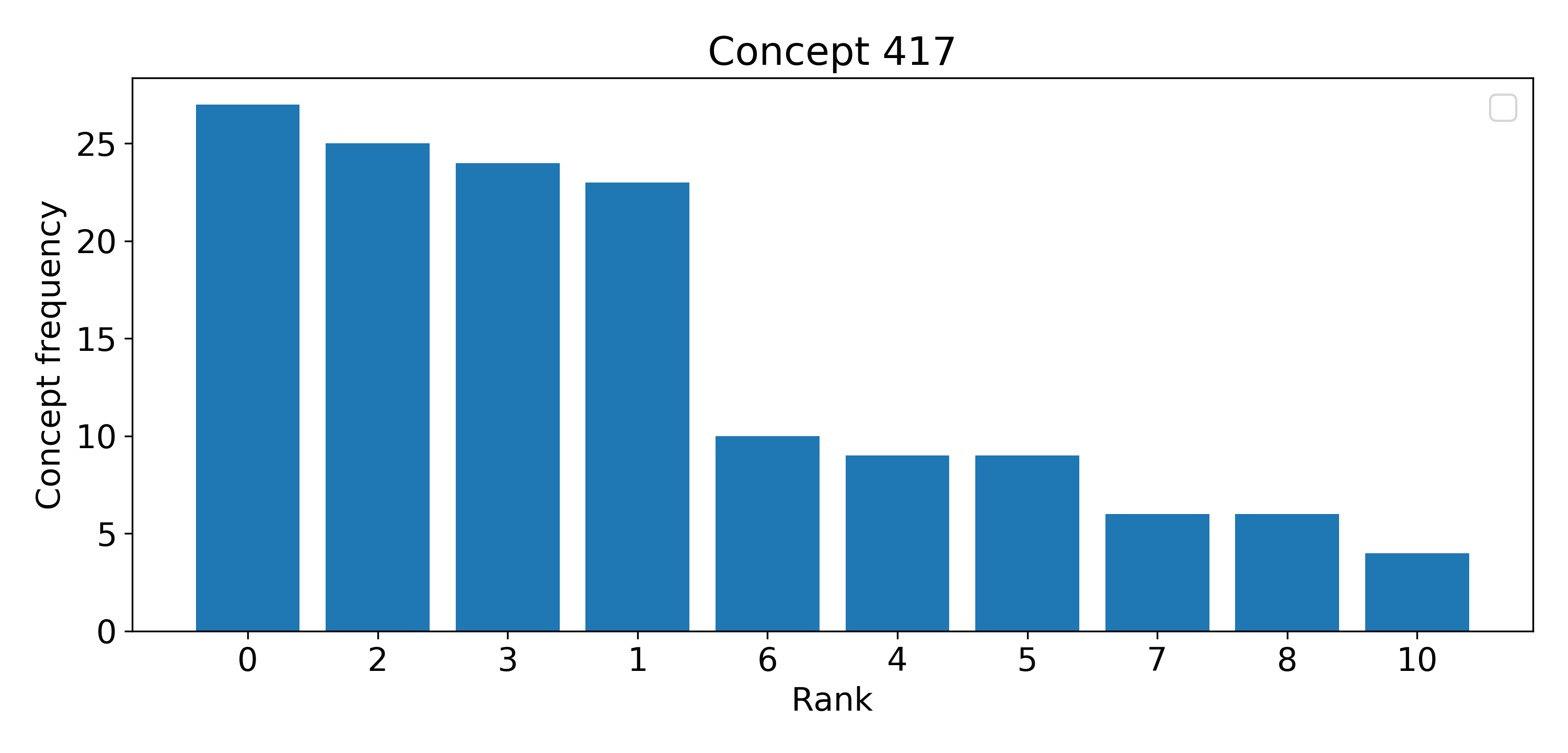}
     \end{subfigure}

    \begin{subfigure}{0.49\linewidth} 
         \centering
         \includegraphics[width=\linewidth]{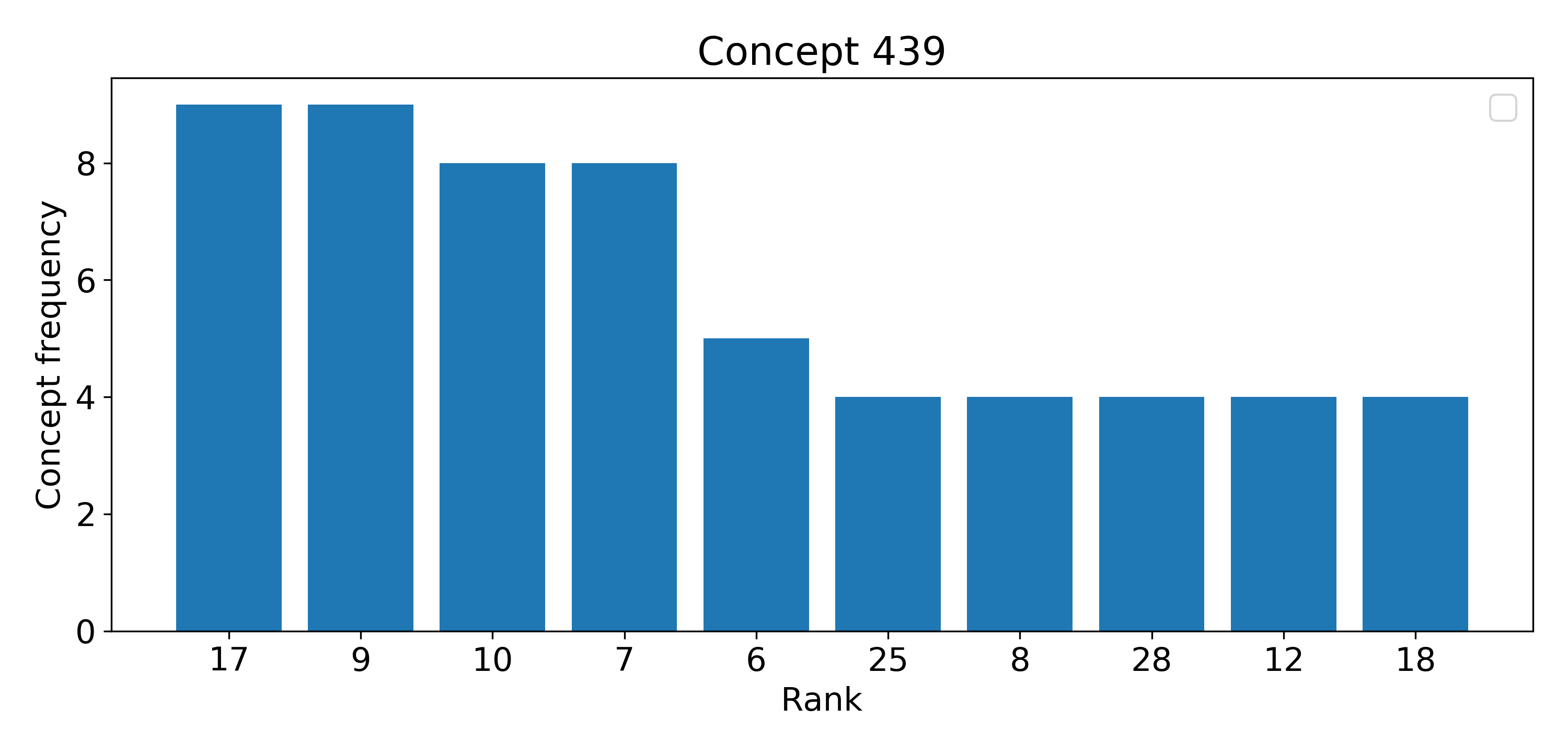}
     \end{subfigure}
    
    \caption{Ranks that are reached the most over all samples for the top-3 rule concepts for experiment 'Adience-Train-MF'. The 3 rules with the most covered samples (sample count in \textbf{bold}) are:\\
            (\textbf{165}) Male, if concept 259 is above of concept 417\\
            (\textbf{78}) Male, if concept 259 is left of concept 377\\
            (\textbf{55}) Male, if concept 24 is above of concept 439}
    \label{fig:rankings_adience_train_mf}
\end{figure}

\begin{figure}[!h]
    \centering

    \begin{subfigure}{0.49\linewidth} 
         \centering
         \includegraphics[width=\linewidth]{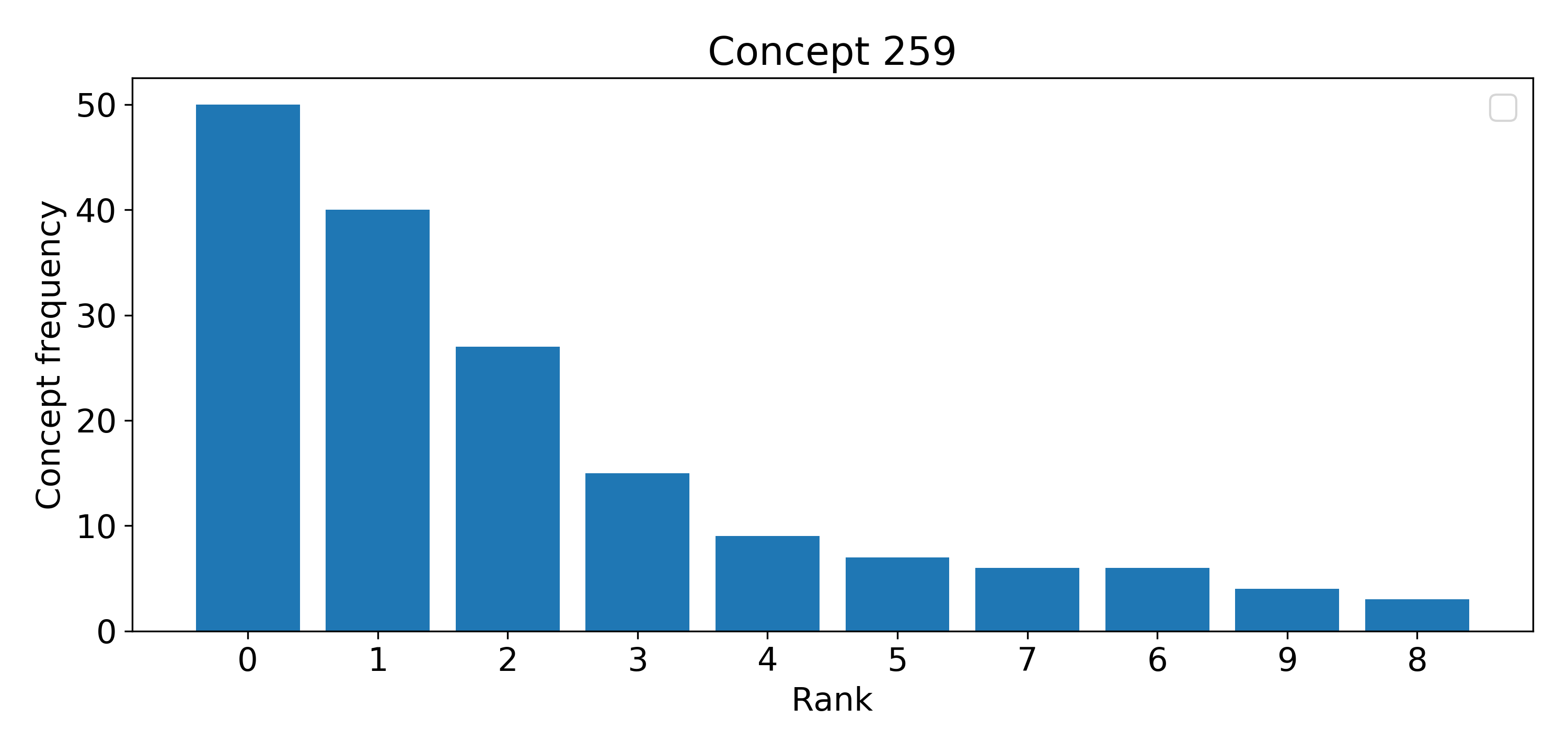}
     \end{subfigure}
    \hfill
    \begin{subfigure}{0.49\linewidth} 
         \centering
         \includegraphics[width=\linewidth]{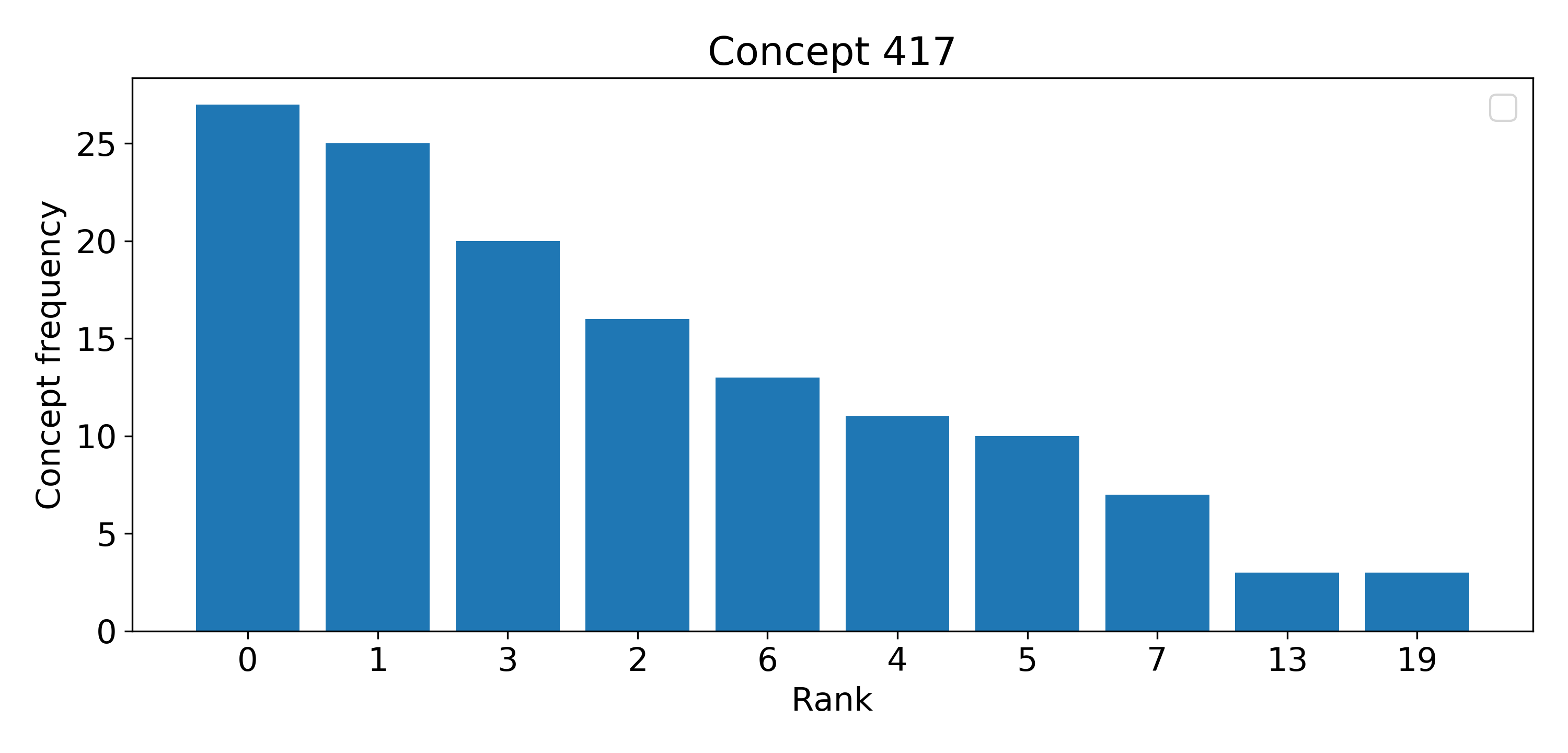}
     \end{subfigure}
     
    \begin{subfigure}{0.49\linewidth} 
         \centering
         \includegraphics[width=\linewidth]{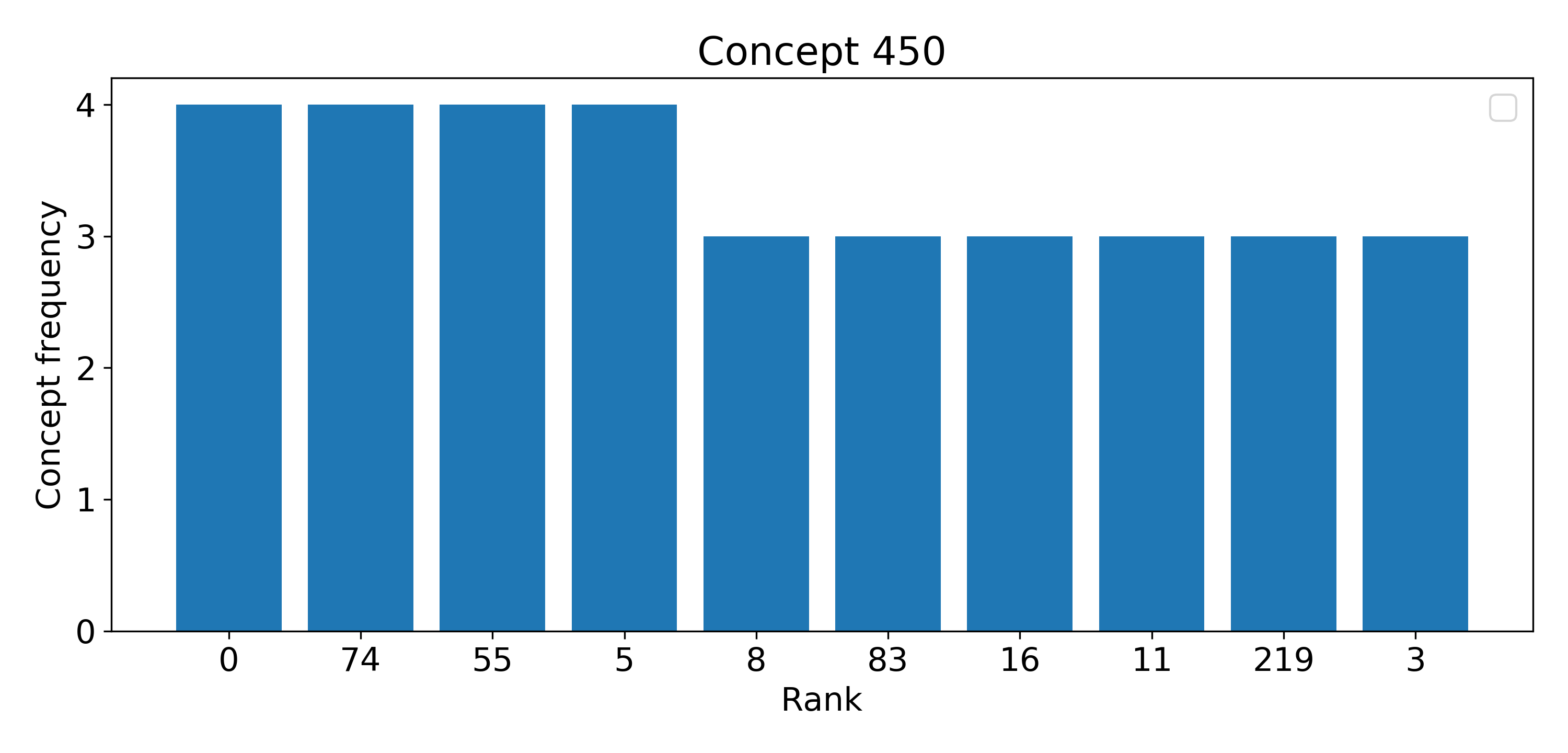}
     \end{subfigure}
    
    \caption{Ranks that are reached the most over all samples for the top-3 rule concepts for experiment 'Adience-Test-MF'. The 3 rules with the most covered samples (sample count in \textbf{bold}) are:\\
            (\textbf{144}) Male, if concept 259 is above of concept 417\\
            (\textbf{48}) Male, if concept 259 is above of concept 450\\
            (\textbf{25}) Male, if concept 259 is below concept 417}
    \label{fig:rankings_adience_test_mf}
\end{figure}

\begin{figure}[!h]
    \centering

    \begin{subfigure}{0.49\linewidth} 
         \centering
         \includegraphics[width=\linewidth]{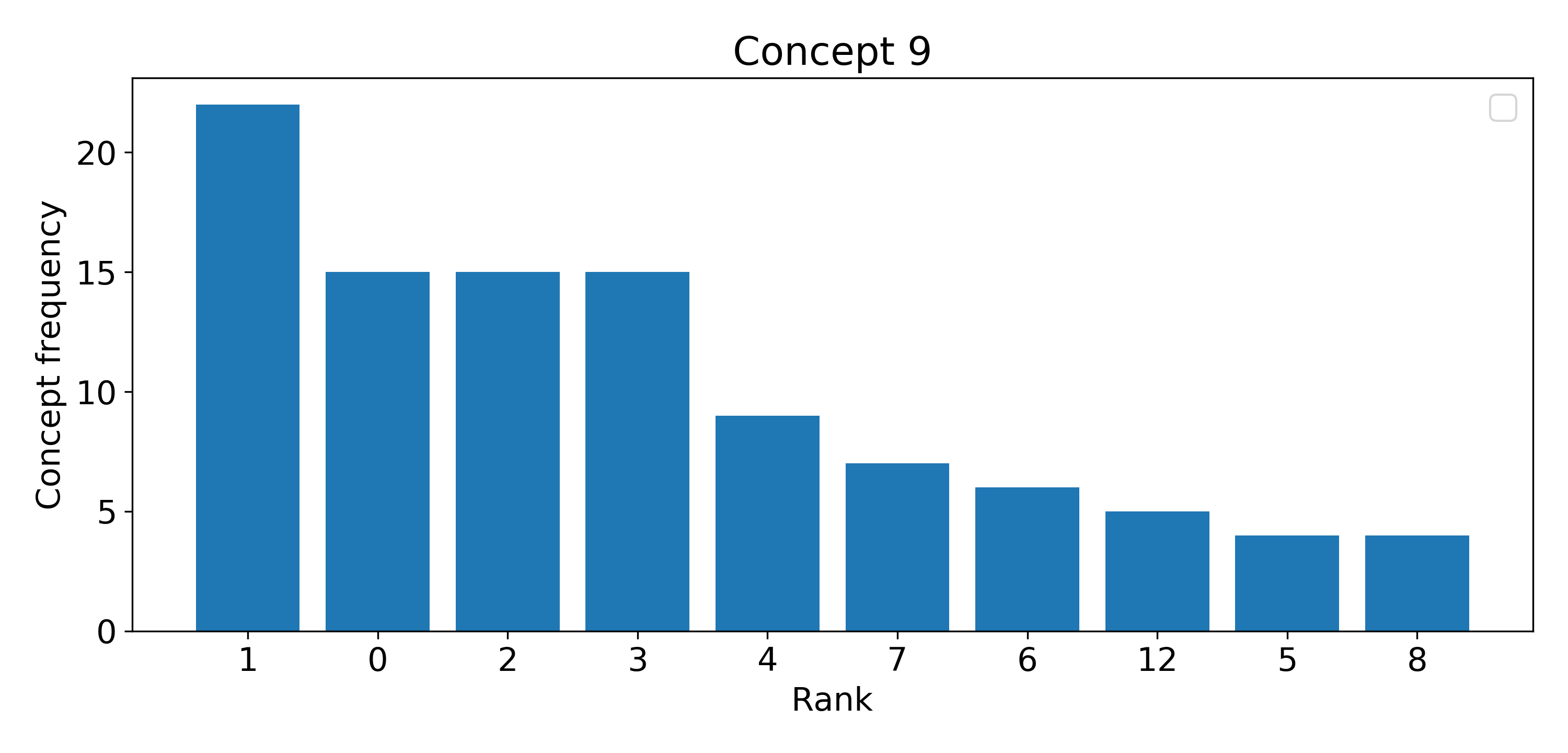}
     \end{subfigure}
    \hfill
    \begin{subfigure}{0.49\linewidth} 
         \centering
         \includegraphics[width=\linewidth]{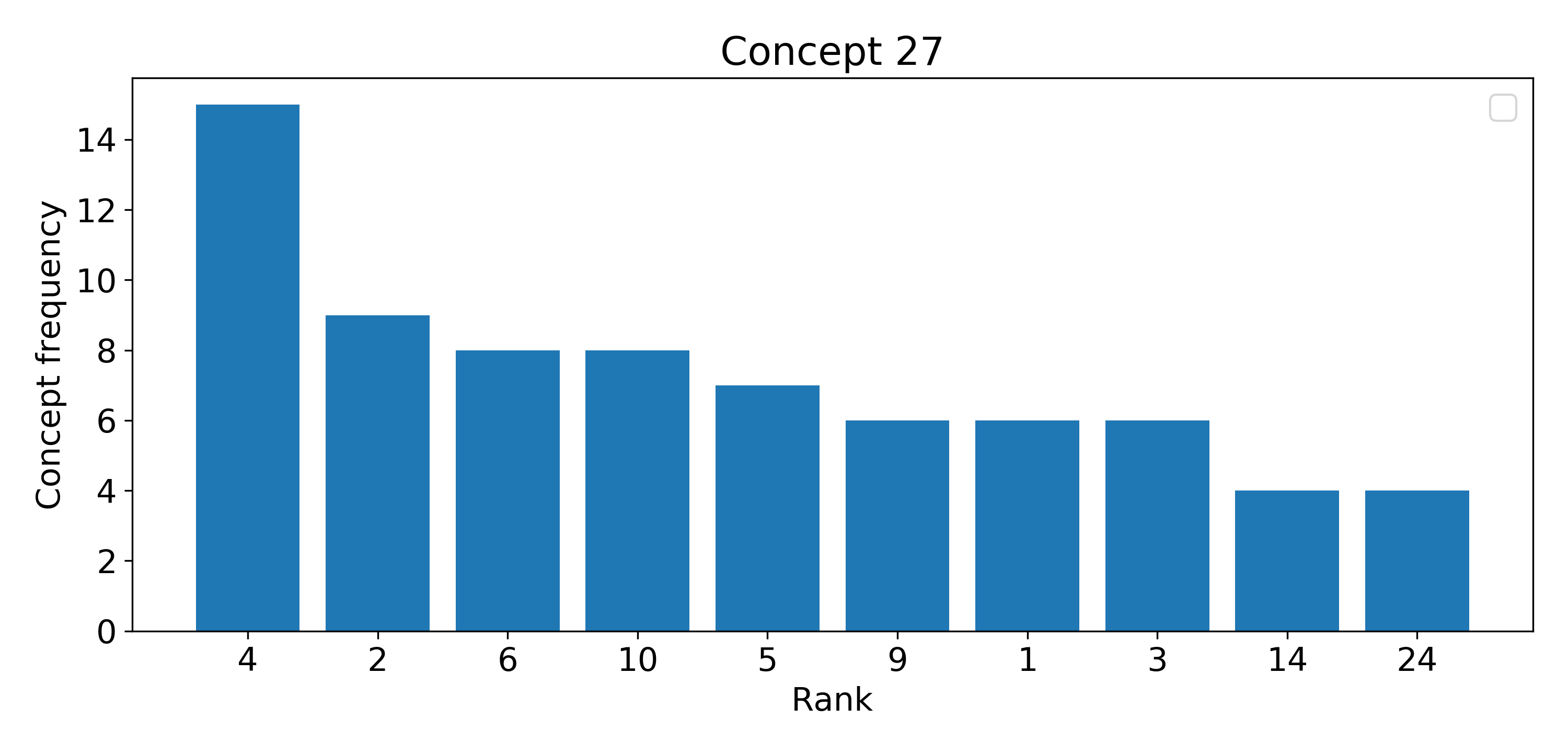}
     \end{subfigure}
     
    \begin{subfigure}{0.49\linewidth} 
         \centering
         \includegraphics[width=\linewidth]{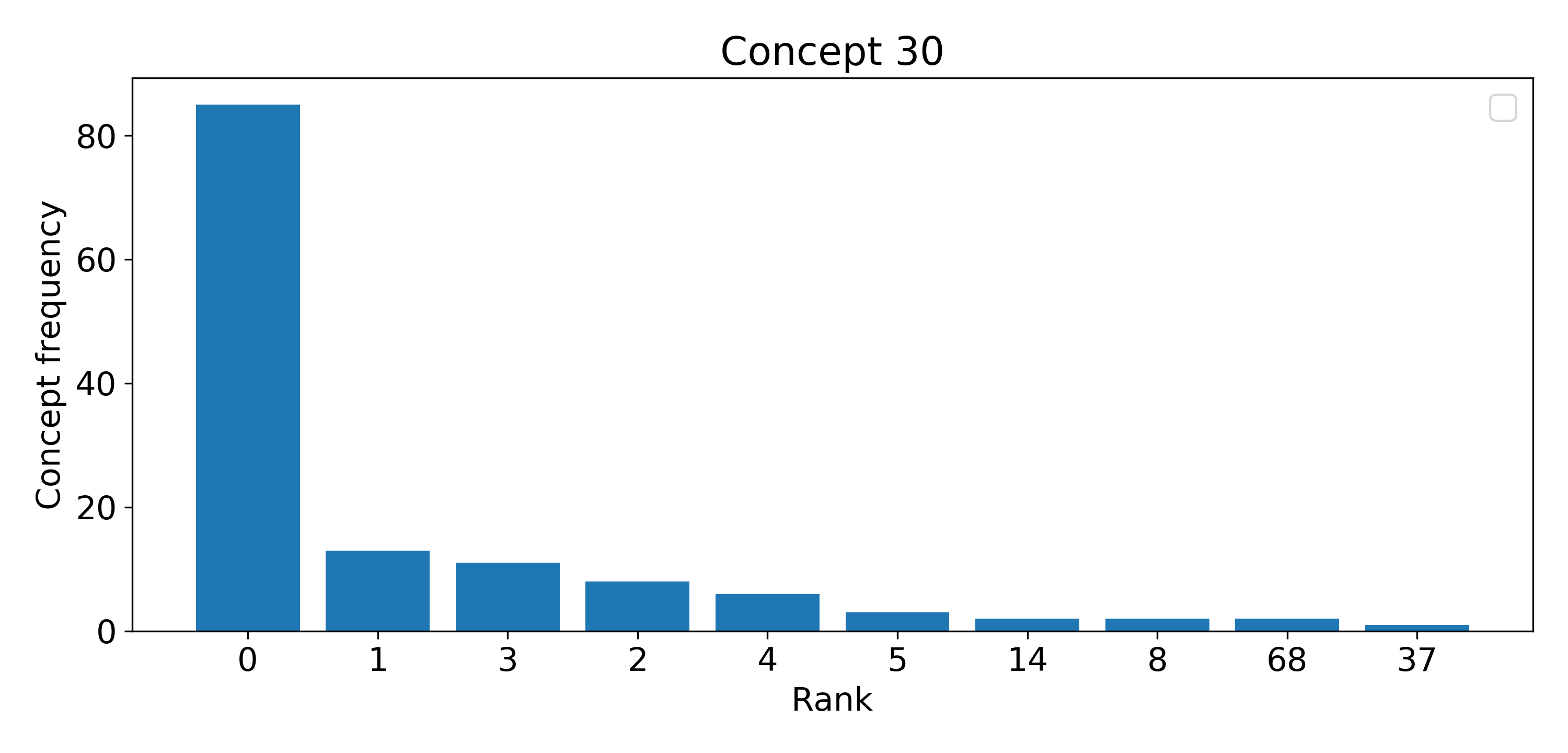}
     \end{subfigure}
    \hfill
    \begin{subfigure}{0.49\linewidth} 
         \centering
         \includegraphics[width=\linewidth]{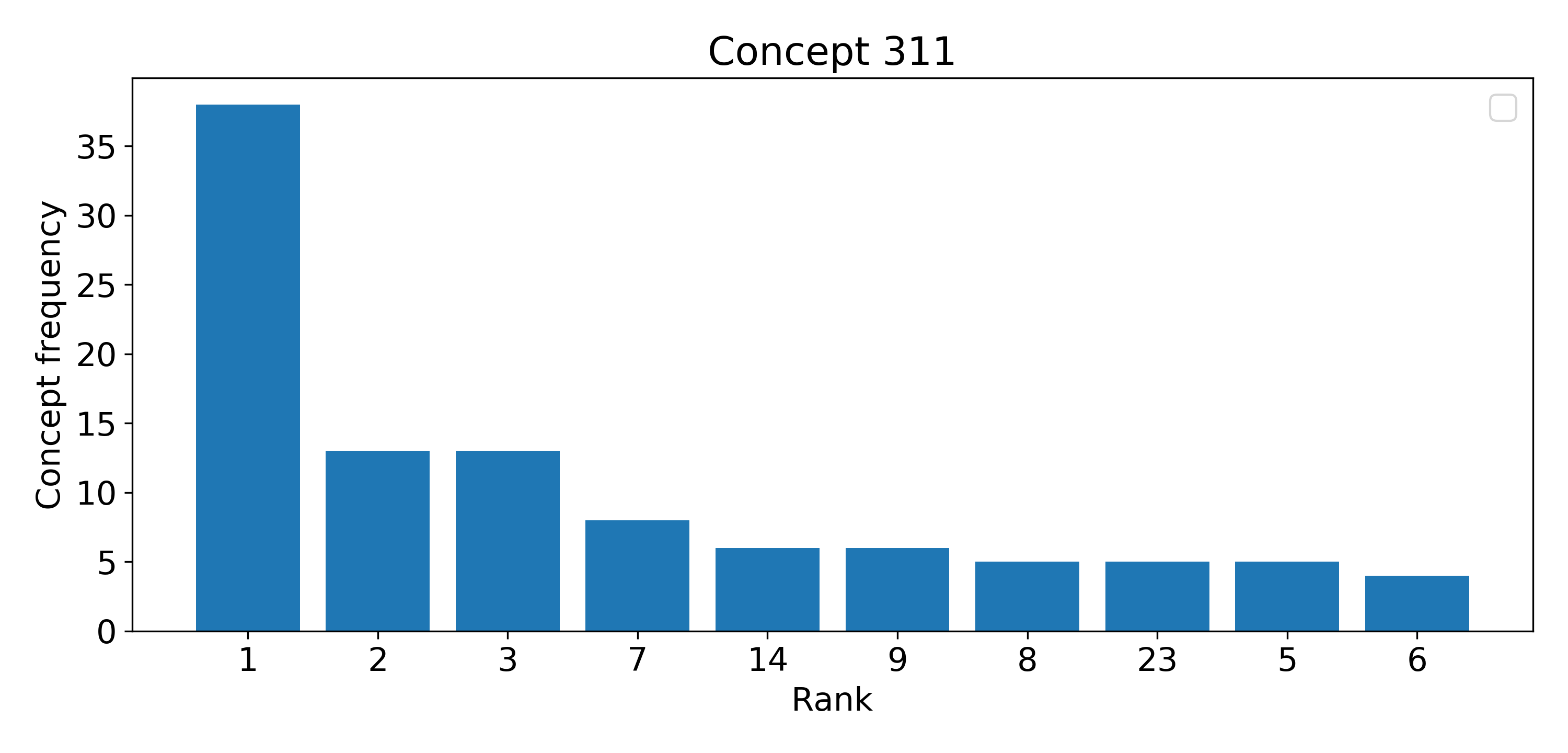}
     \end{subfigure}
    
    \caption{Ranks that are reached the most over all samples for the top-3 rule concepts for experiment 'Teapot-Vase-Train'. The 3 rules with the most covered samples (sample count in \textbf{bold}) are:\\
        (\textbf{97}) Teapot, if concept 27 is above of concept 311\\
        (\textbf{89}) Teapot, if concept 30 is middle right of concept 9\\
        (\textbf{52}) Teapot, if concept 30 is right of concept 9}
    \label{fig:rankings_teapot}
\end{figure}

\begin{figure}[!h]
    \centering

    \begin{subfigure}{0.49\linewidth} 
         \centering
         \includegraphics[width=\linewidth]{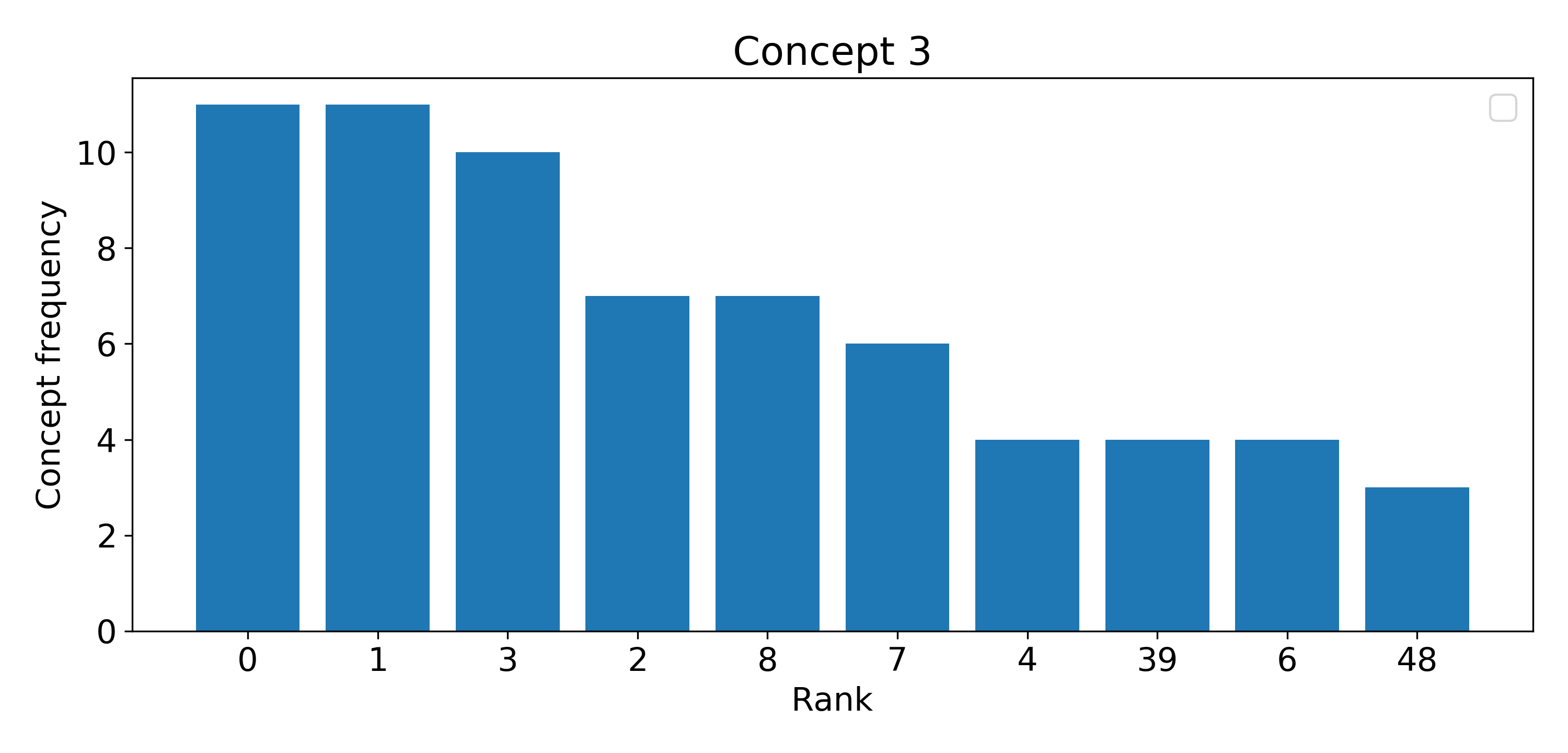}
     \end{subfigure}
    \hfill
    \begin{subfigure}{0.49\linewidth} 
         \centering
         \includegraphics[width=\linewidth]{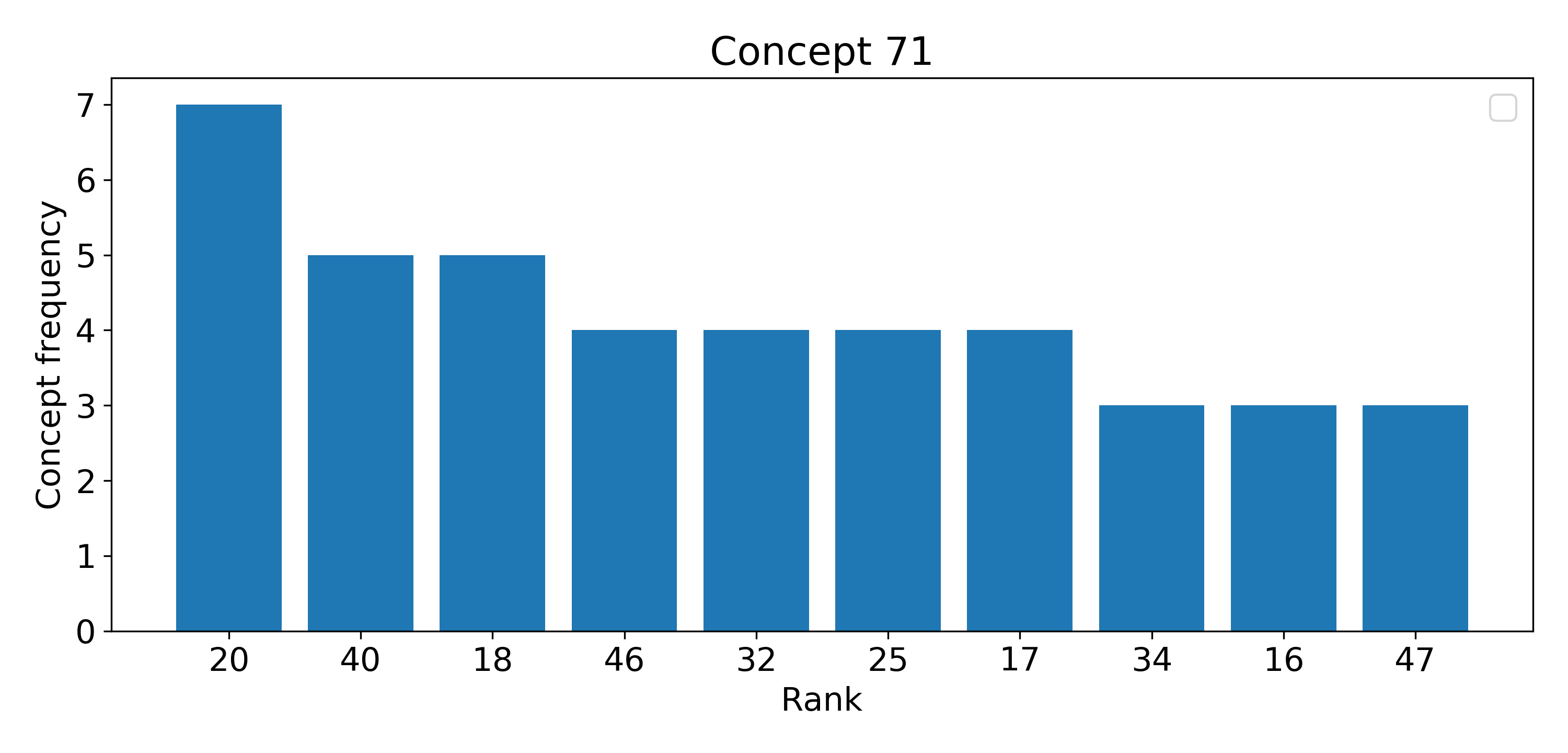}
     \end{subfigure}
     
    \begin{subfigure}{0.49\linewidth} 
         \centering
         \includegraphics[width=\linewidth]{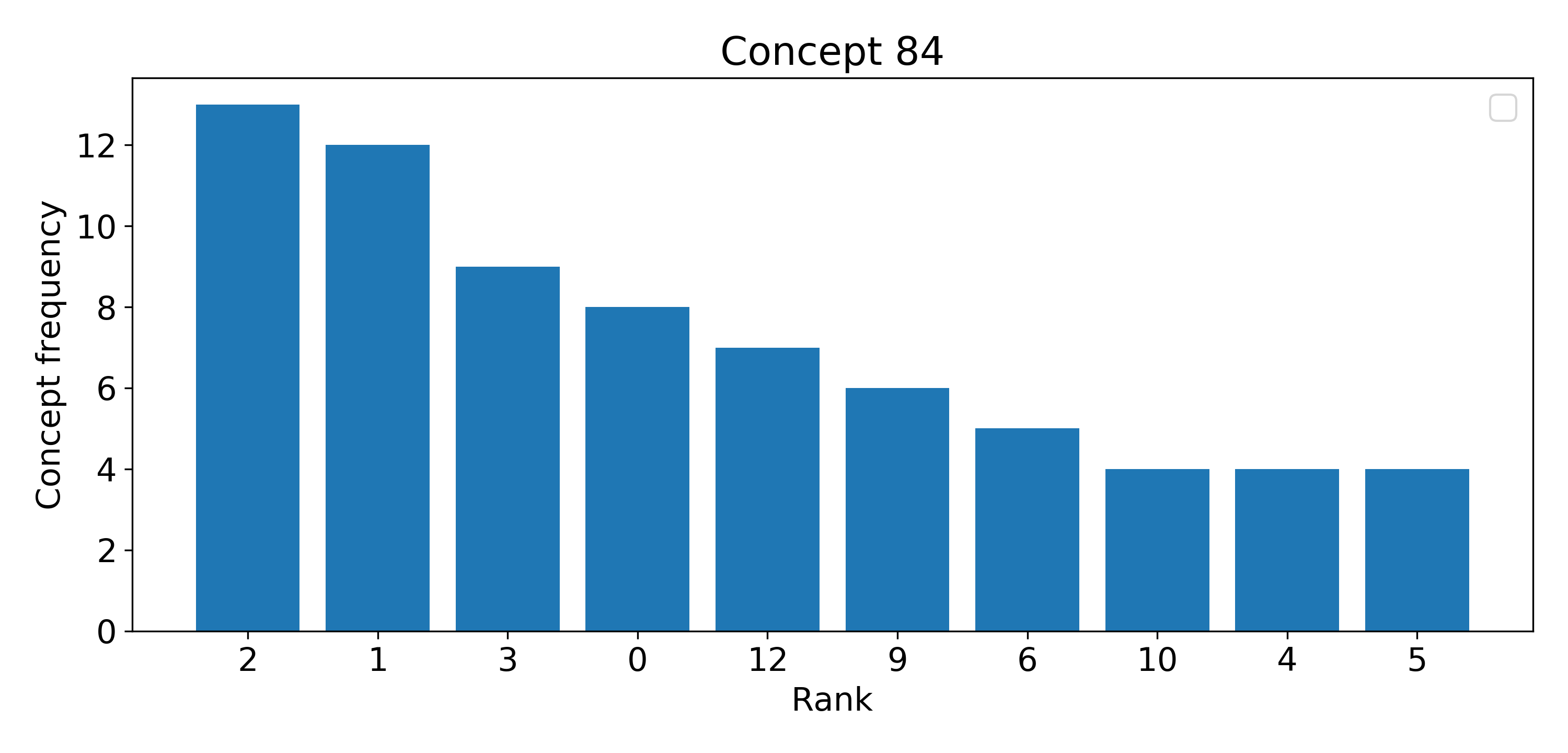}
     \end{subfigure}
    \hfill
    \begin{subfigure}{0.49\linewidth} 
         \centering
         \includegraphics[width=\linewidth]{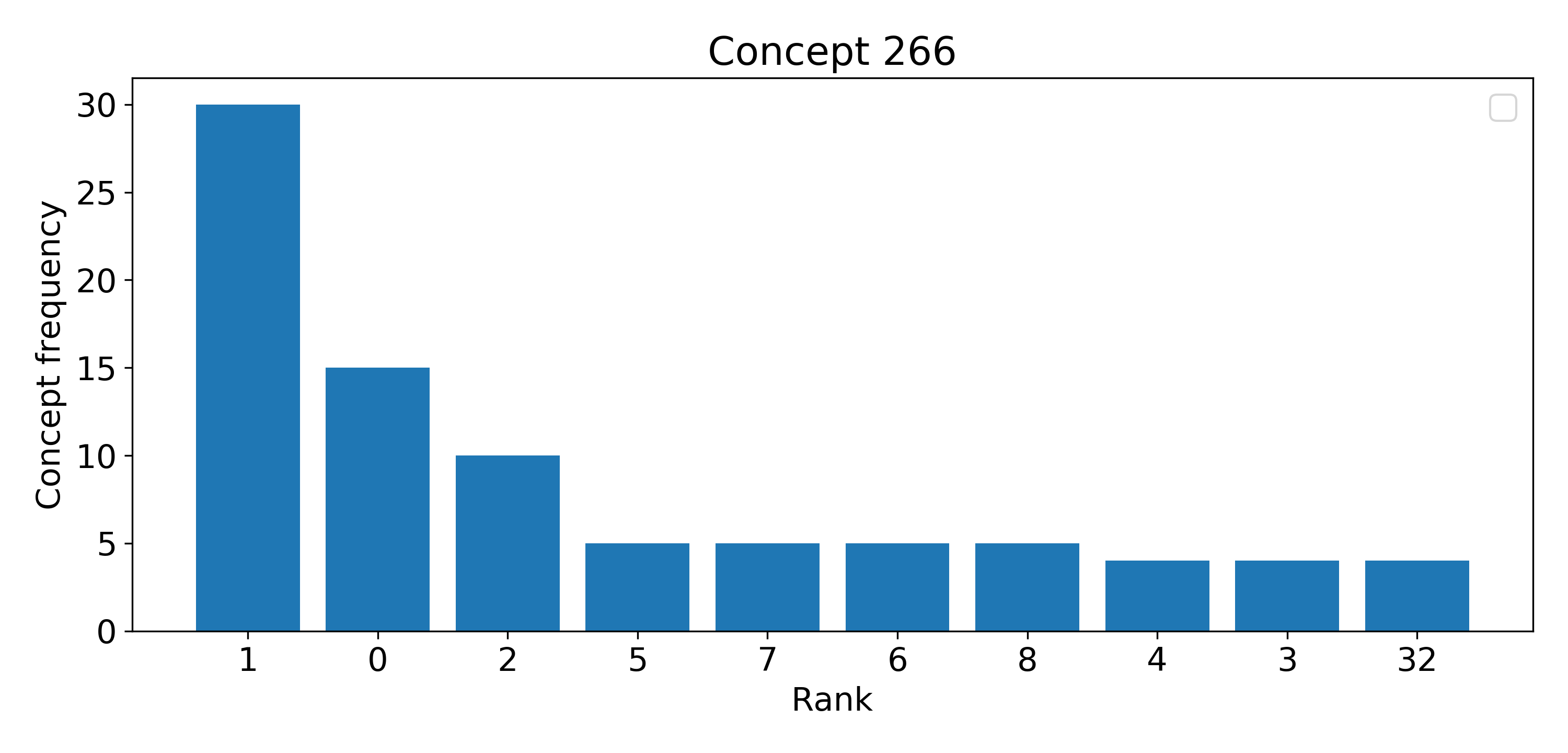}
     \end{subfigure}

    \begin{subfigure}{0.49\linewidth} 
         \centering
         \includegraphics[width=\linewidth]{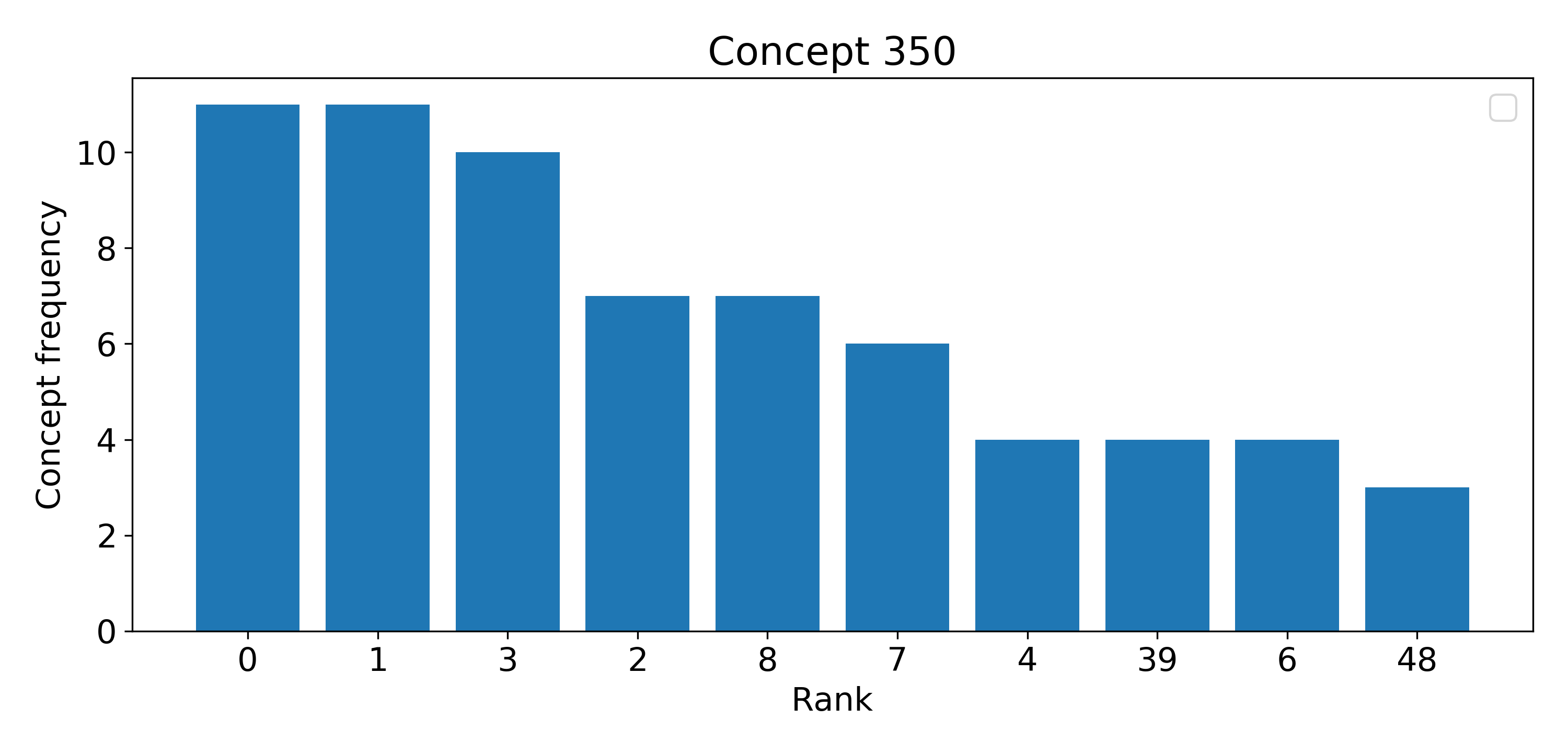}
     \end{subfigure}
    \hfill
    \begin{subfigure}{0.49\linewidth} 
         \centering
         \includegraphics[width=\linewidth]{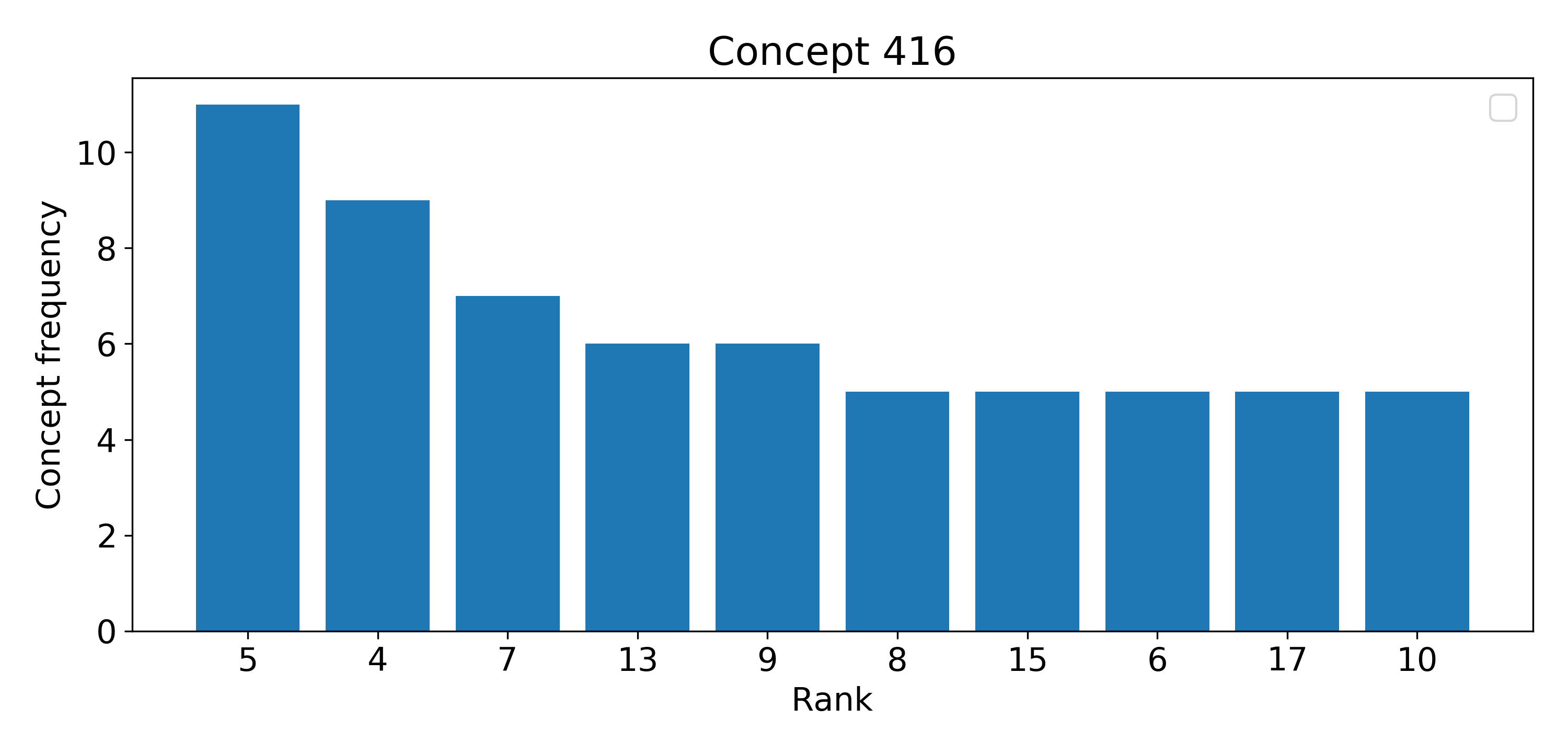}
     \end{subfigure}
    
    \caption{Ranks that are reached the most over all samples for the top-3 rule concepts for experiment 'Vase-Teapot-Train'. The 3 rules with the most covered samples (sample count in \textbf{bold}) are:\\
            (\textbf{70}) Vase, if concept 350 is below concept 266\\
            (\textbf{66}) Vase, if concept 71 is below concept 416\\
            (\textbf{49}) Vase, if concept 84 is right of concept 3}
    \label{fig:rankings_vase}
\end{figure}

\begin{figure}[!h]
    \centering

    \begin{subfigure}{0.49\linewidth} 
         \centering
         \includegraphics[width=\linewidth]{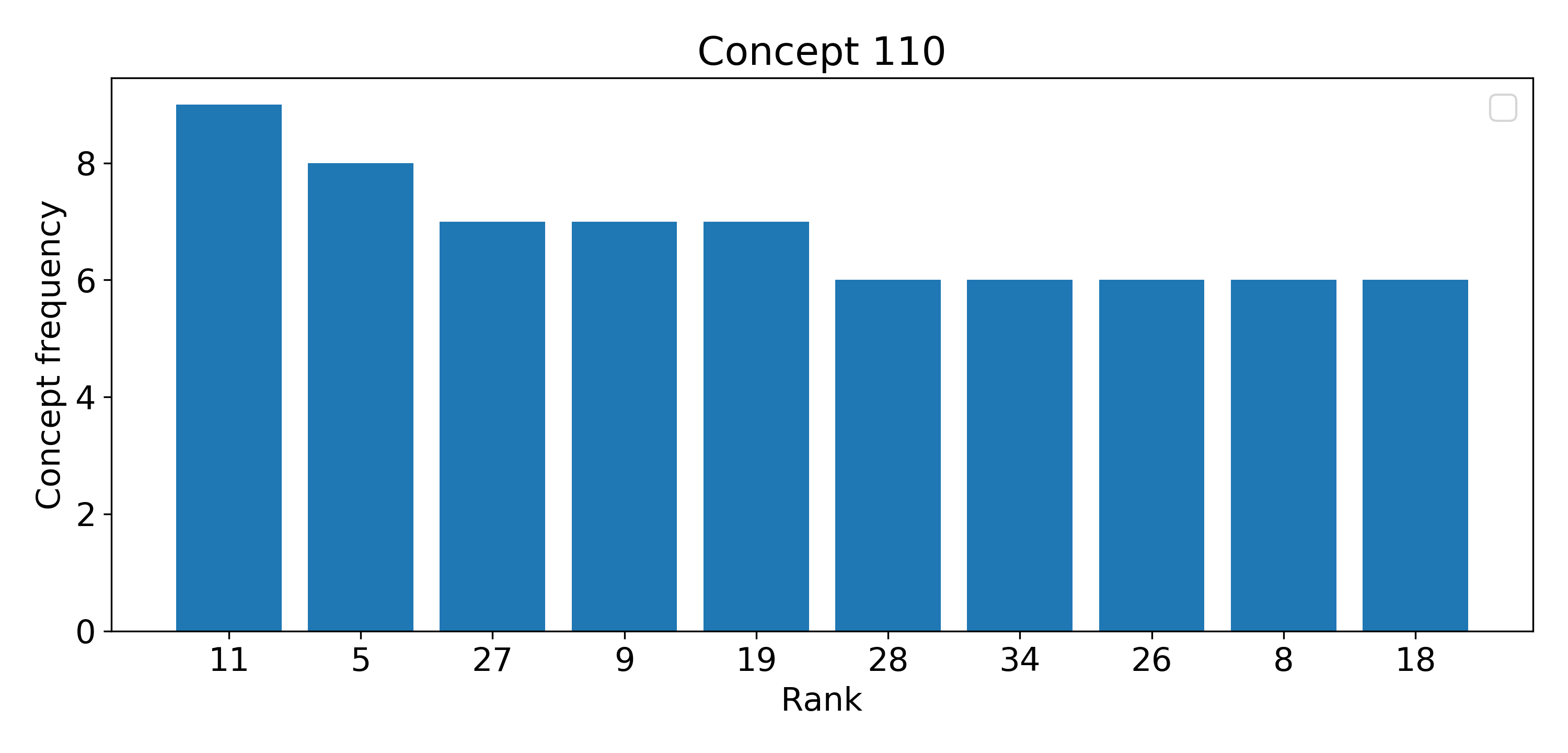}
     \end{subfigure}
    \hfill
    \begin{subfigure}{0.49\linewidth} 
         \centering
         \includegraphics[width=\linewidth]{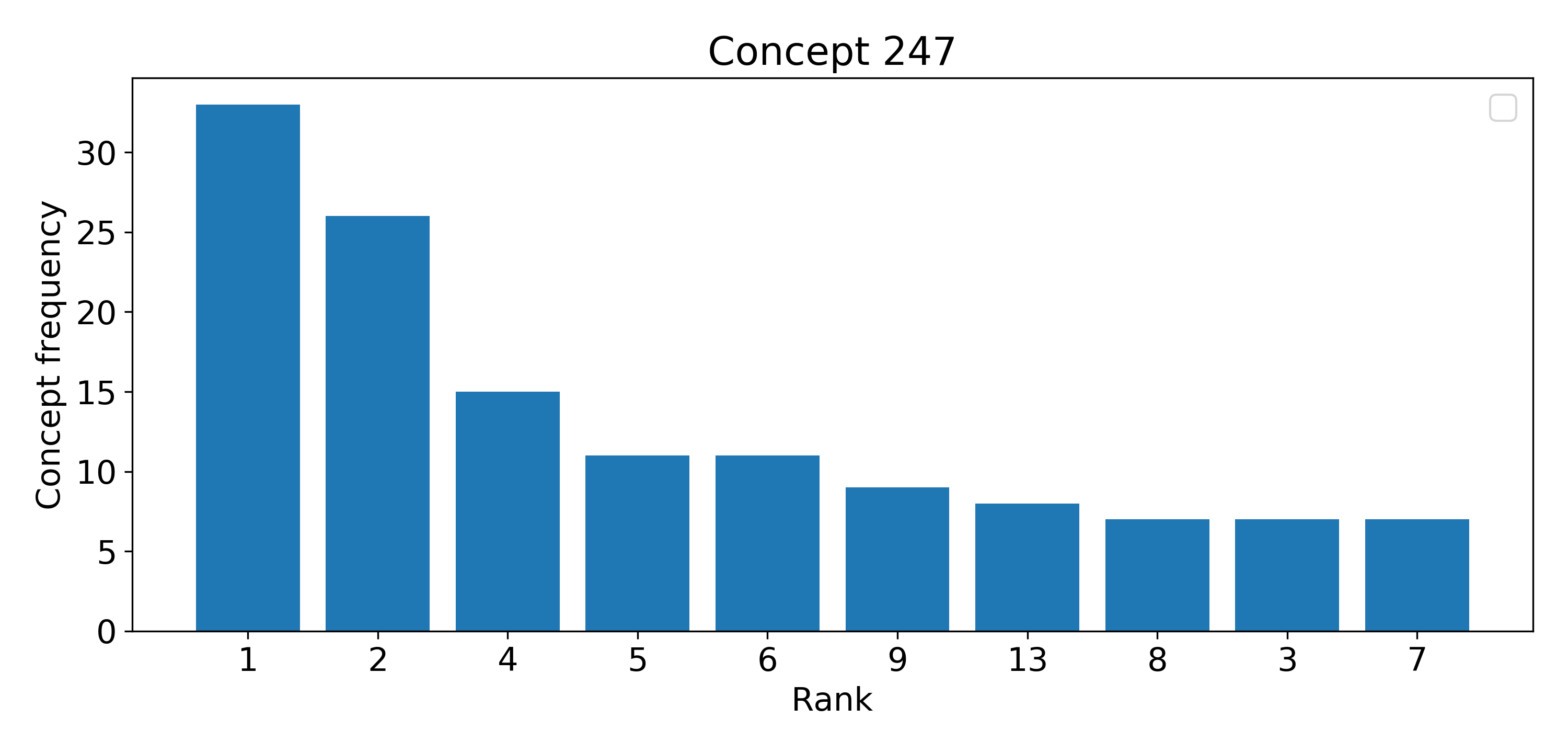}
     \end{subfigure}
     
    \begin{subfigure}{0.49\linewidth} 
         \centering
         \includegraphics[width=\linewidth]{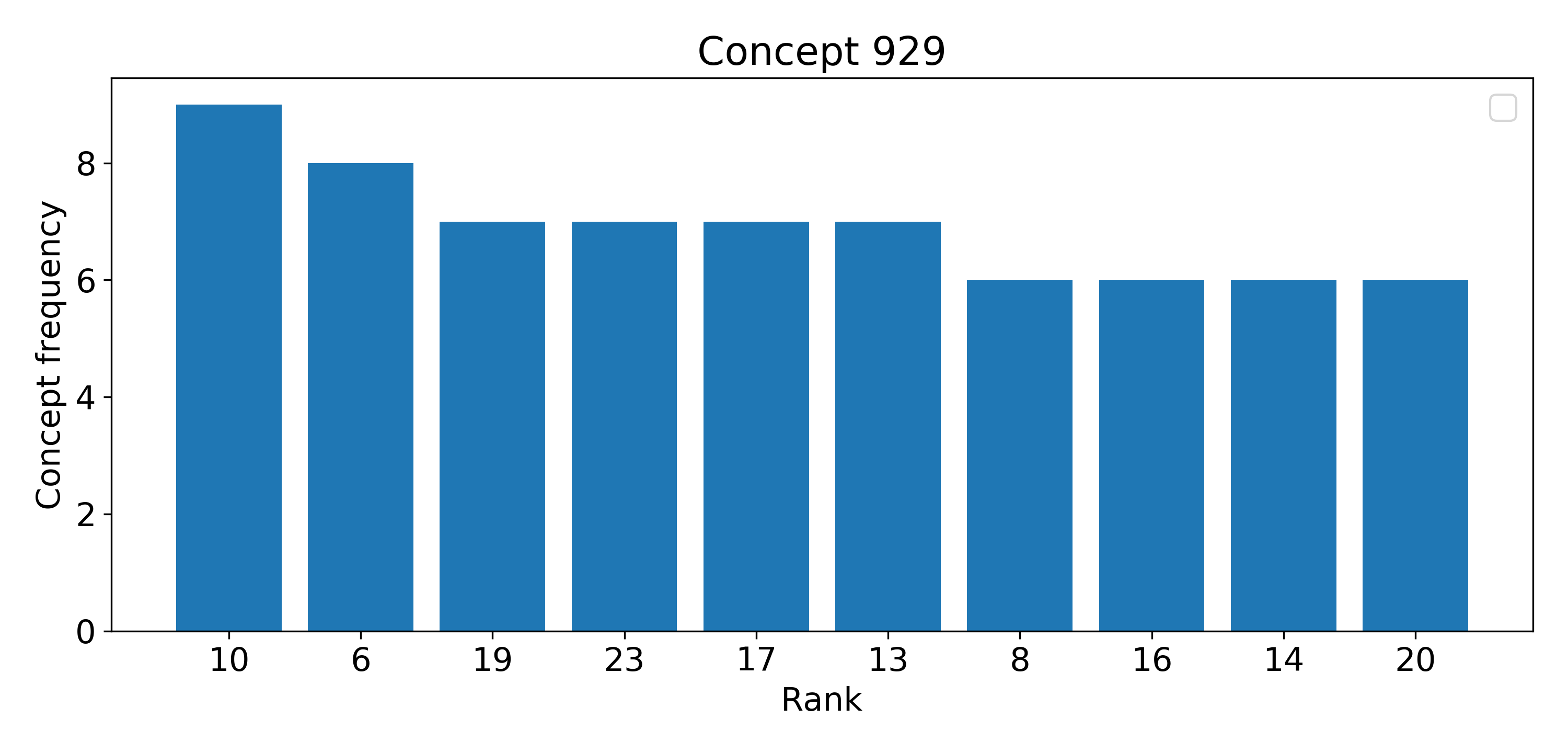}
     \end{subfigure}
    \hfill
    \begin{subfigure}{0.49\linewidth} 
         \centering
         \includegraphics[width=\linewidth]{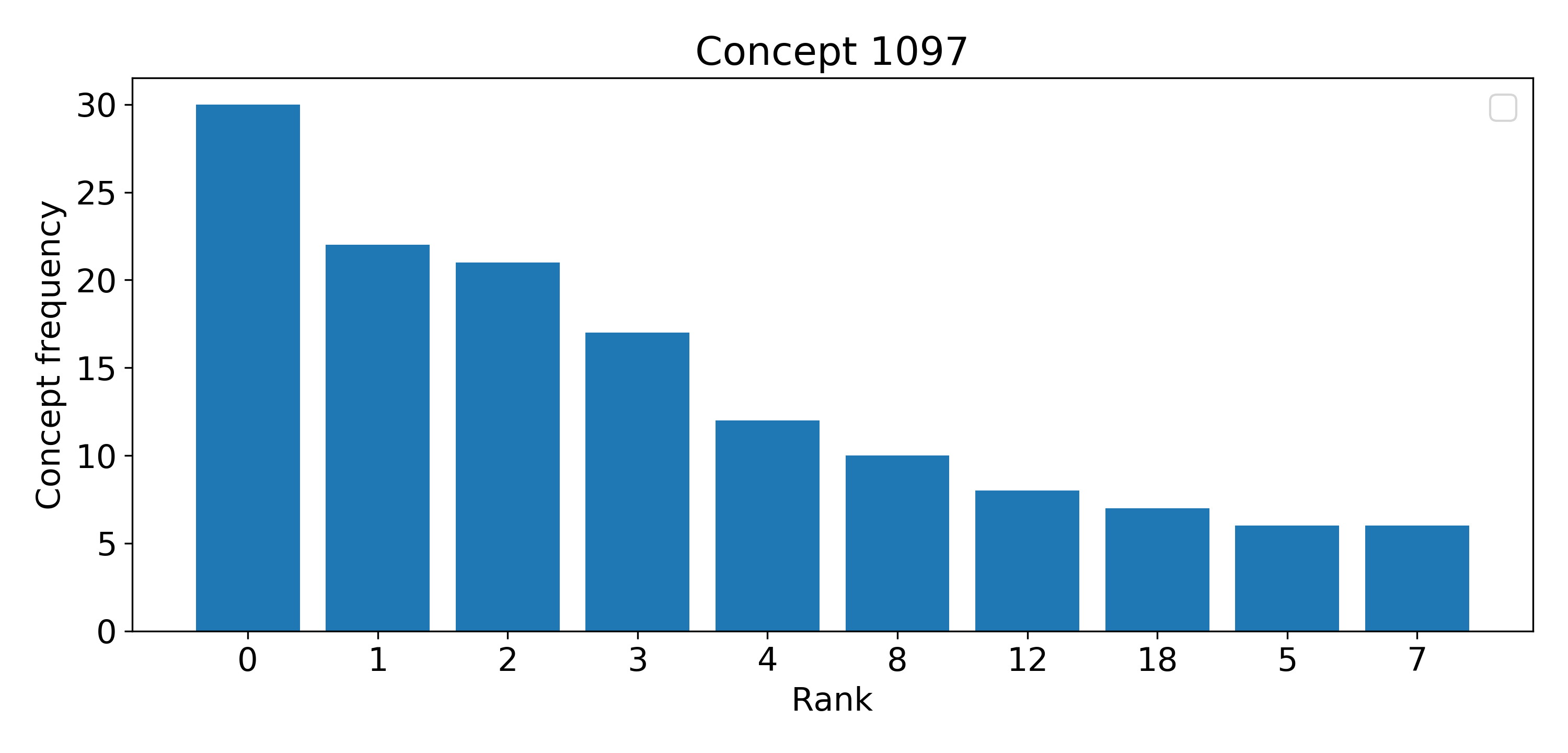}
     \end{subfigure}

    \begin{subfigure}{0.49\linewidth} 
         \centering
         \includegraphics[width=\linewidth]{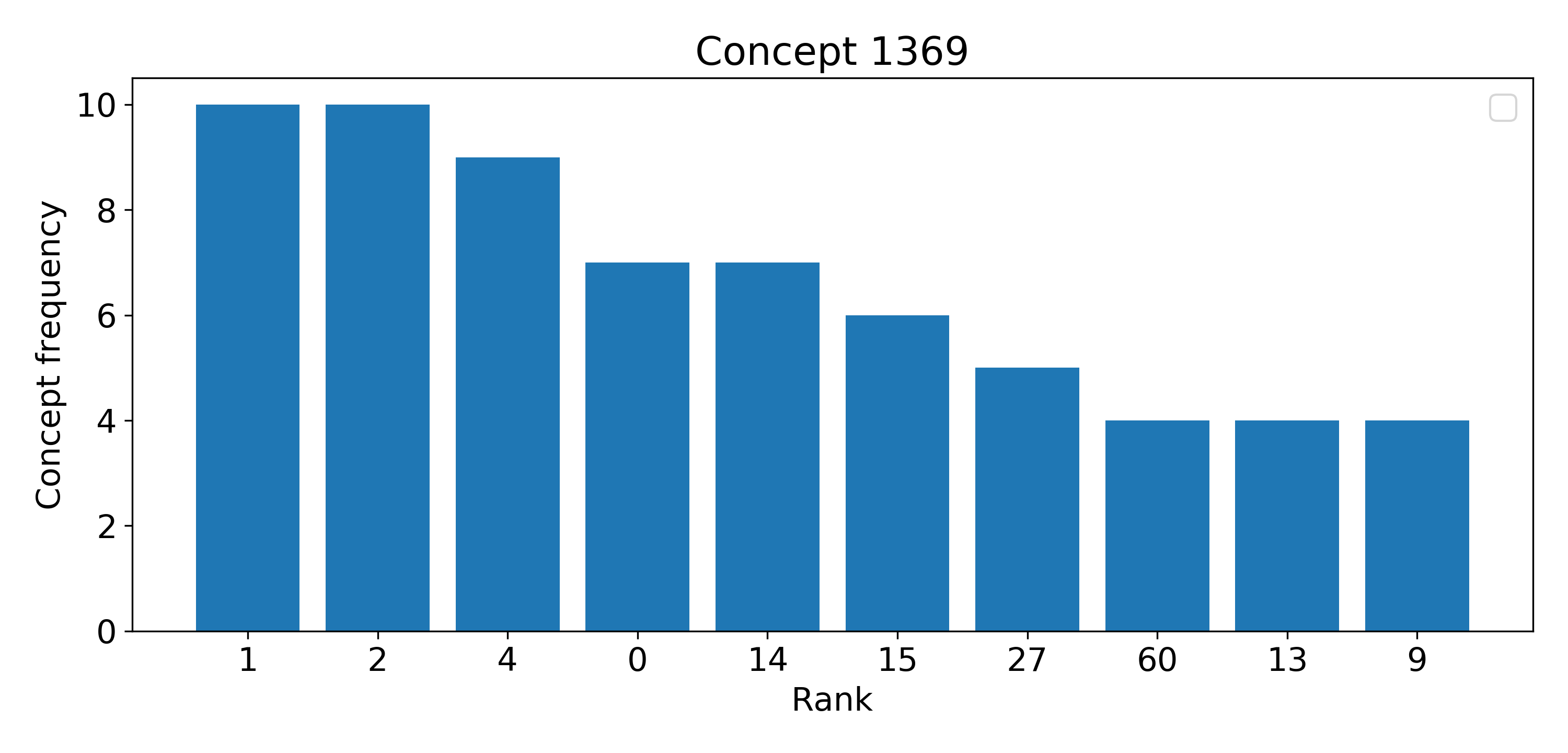}
     \end{subfigure}
    
    \caption{Ranks that are reached the most over all samples for the top-3 rule concepts for experiment 'PathMNIST-Train'. The 3 rules with the most covered samples (sample count in \textbf{bold}) are:\\
            (\textbf{213}) Cancerous, if concept 1097 is bottom right of concept 247\\
            (\textbf{197}) Cancerous, if sample contains concept 110\\
            (\textbf{115}) Cancerous, if concept 1369 is middle right of concept 929}
    \label{fig:rankings_pathmnist_train}
\end{figure}

\begin{figure}[!h]
    \centering

    \begin{subfigure}{0.49\linewidth} 
         \centering
         \includegraphics[width=\linewidth]{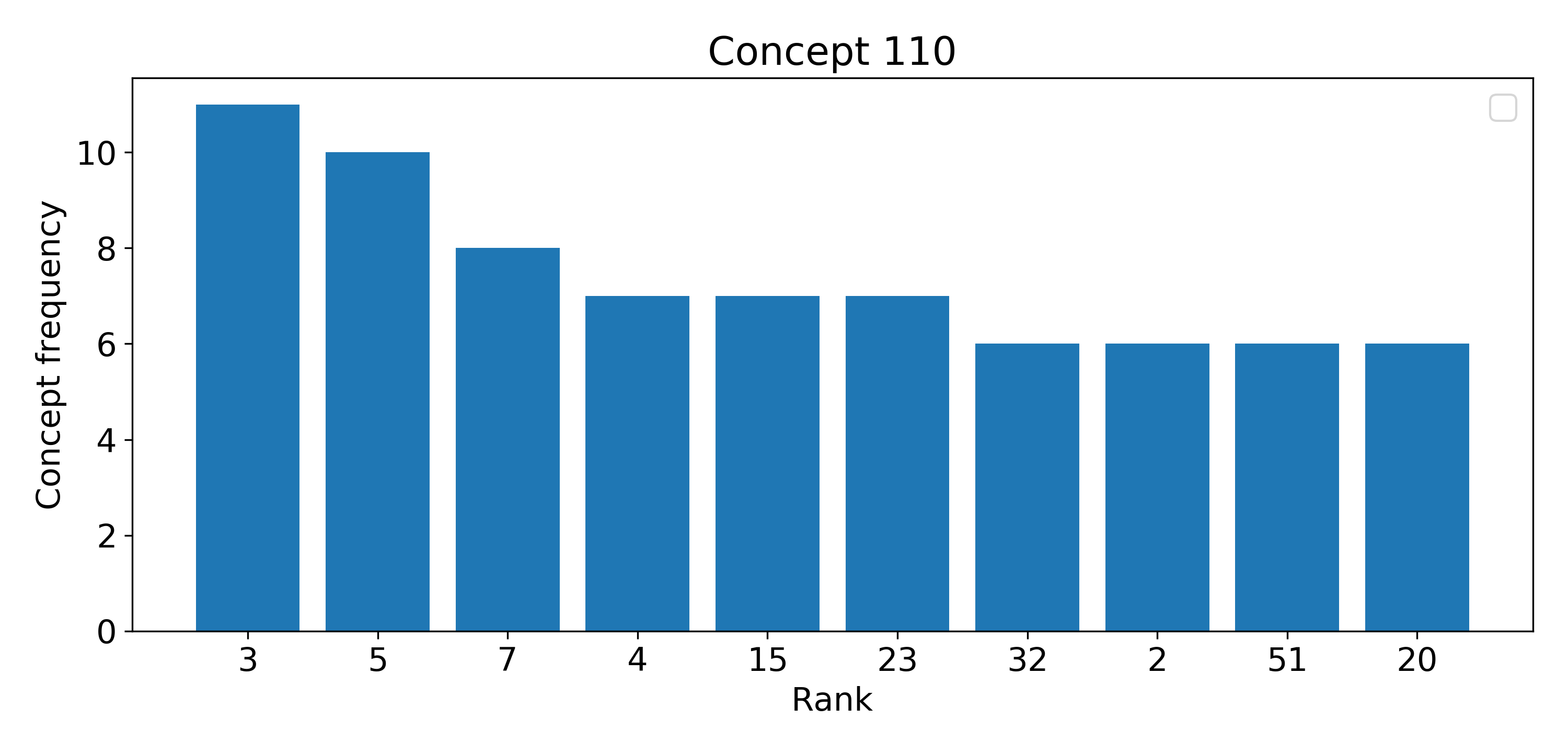}
     \end{subfigure}
    \hfill
    \begin{subfigure}{0.49\linewidth} 
         \centering
         \includegraphics[width=\linewidth]{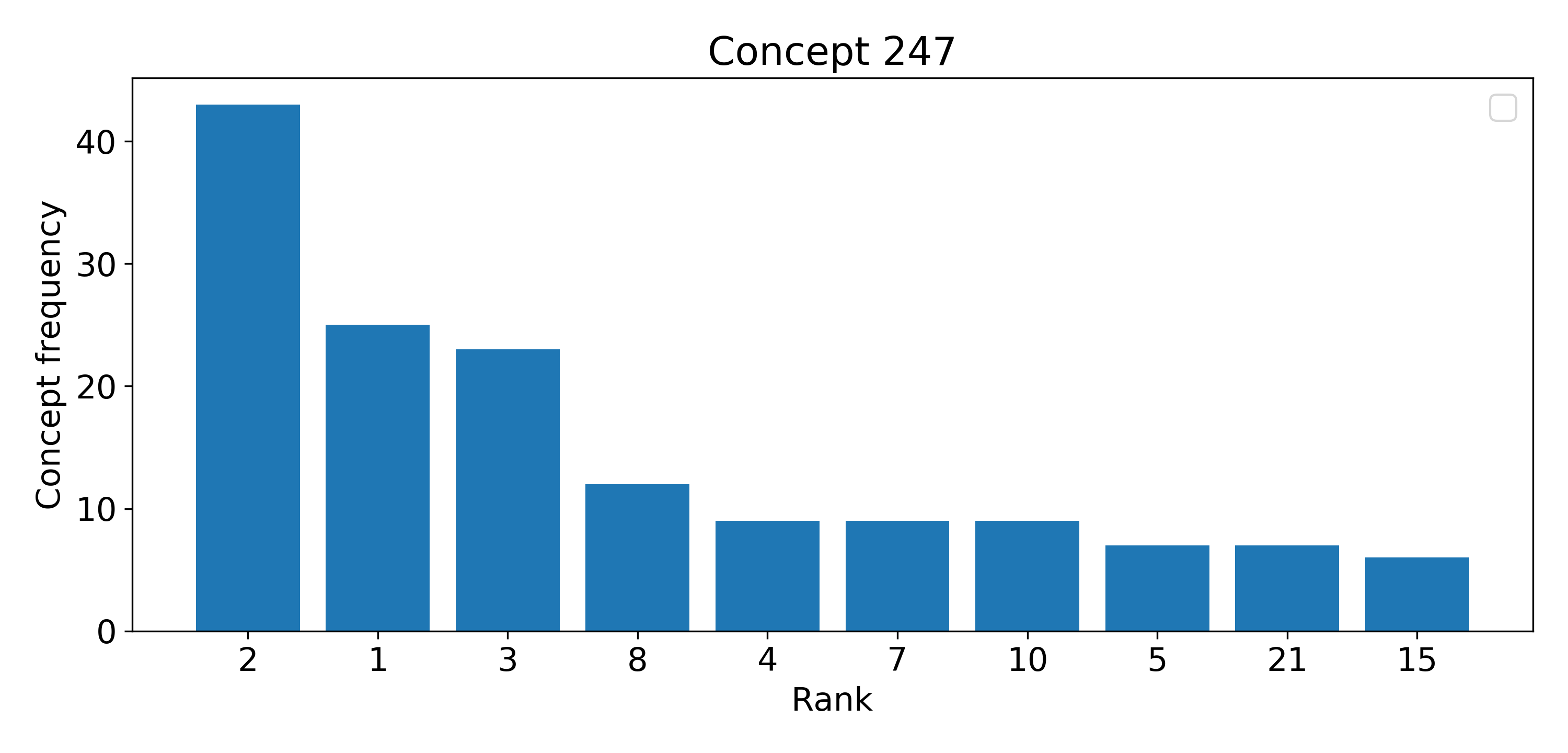}
     \end{subfigure}
     
    \begin{subfigure}{0.49\linewidth} 
         \centering
         \includegraphics[width=\linewidth]{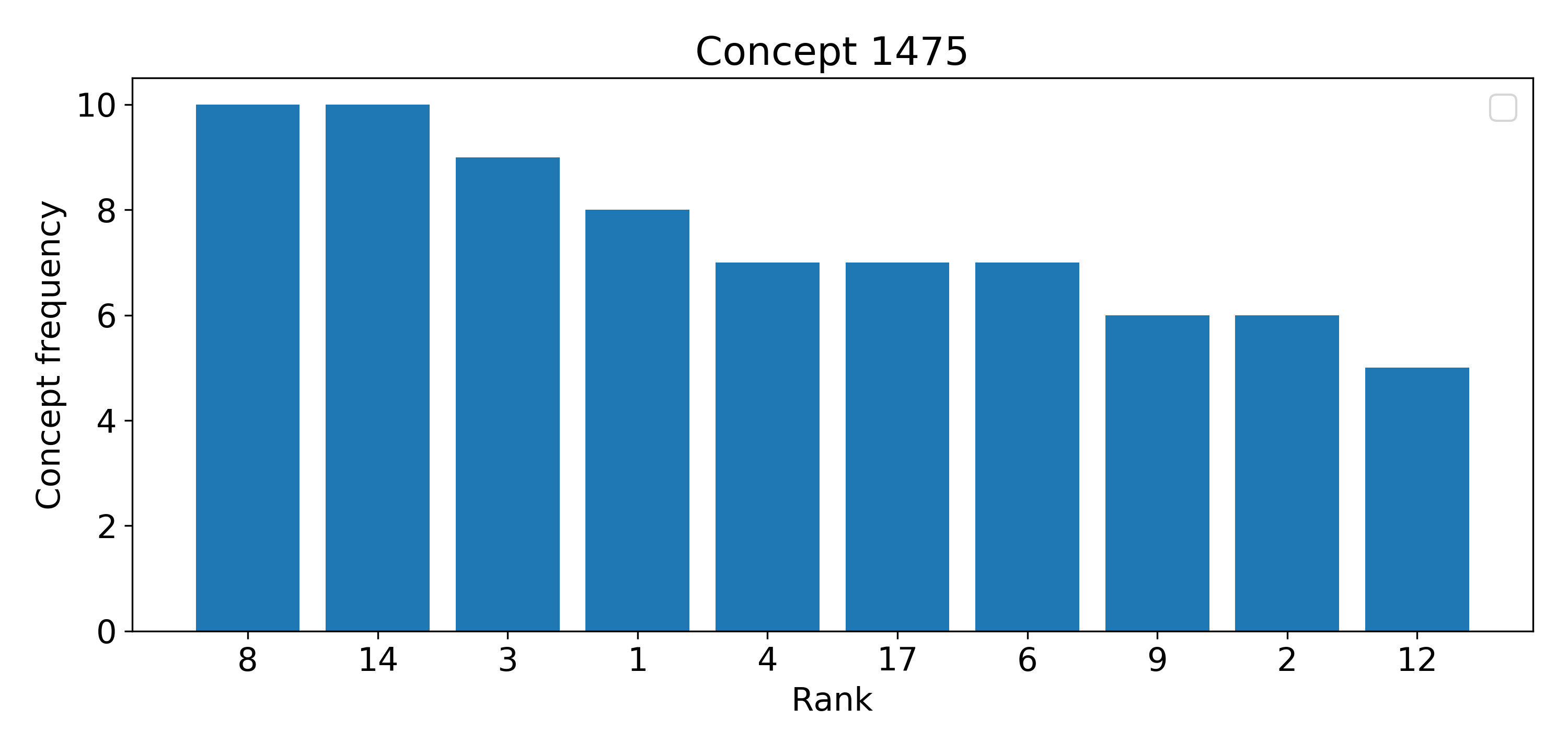}
     \end{subfigure}
    \hfill
    \begin{subfigure}{0.49\linewidth} 
         \centering
         \includegraphics[width=\linewidth]{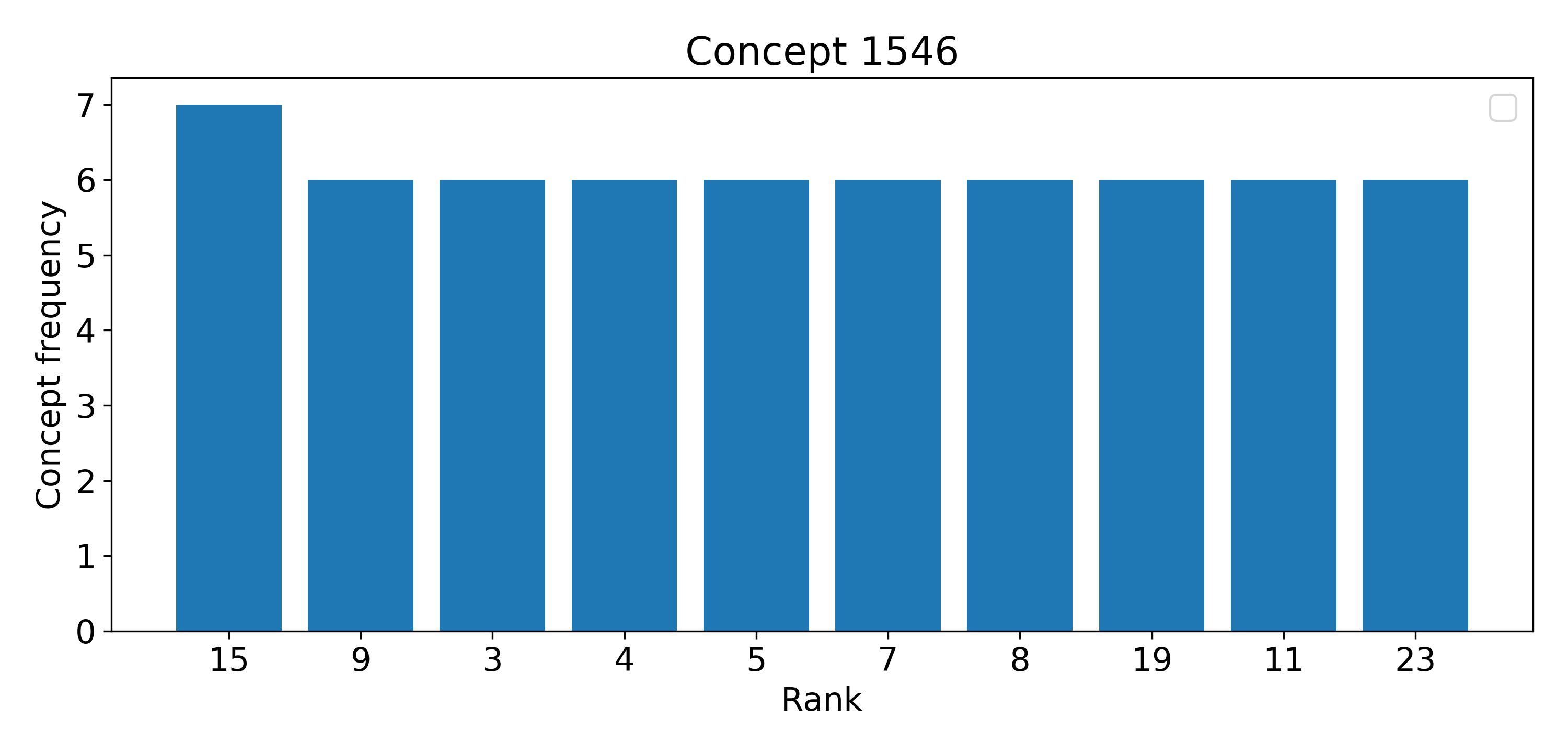}
     \end{subfigure}

    \begin{subfigure}{0.49\linewidth} 
         \centering
         \includegraphics[width=\linewidth]{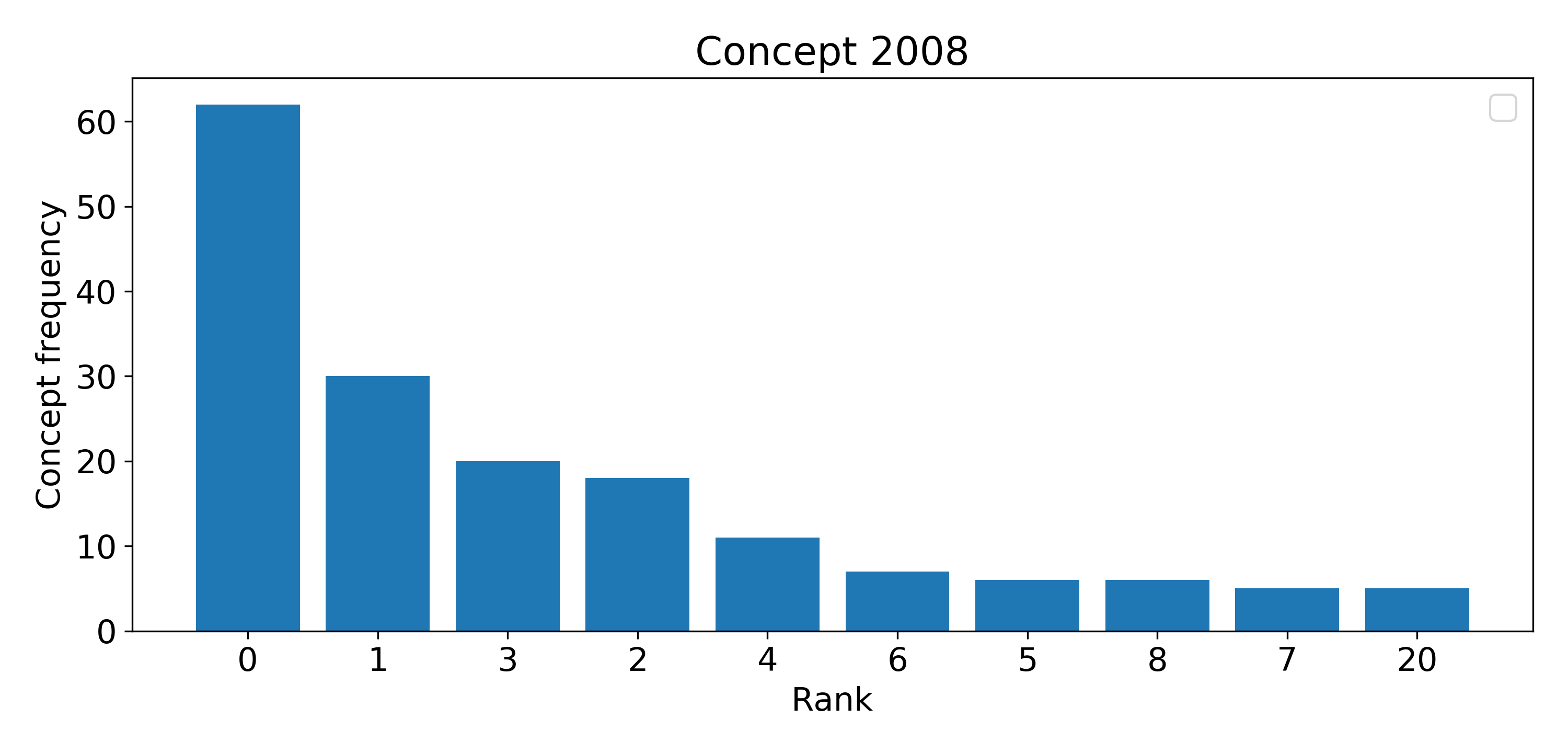}
     \end{subfigure}
    
    \caption{Ranks that are reached the most over all samples for the top-3 rule concepts for experiment 'PathMNIST-Test'. The 3 rules with the most covered samples (sample count in \textbf{bold}) are:\\
            (\textbf{208}) Cancerous, if concept 247 is bottom middle of concept 2008\\
            (\textbf{191}) Cancerous, if concept 110 is middle right of concept 2008\\
            (\textbf{166}) Cancerous, if concept 1475 is middle right of concept 1546}
    \label{fig:rankings_pathmnist_test}
\end{figure}

The Figures \ref{fig:rankings_picasso_train}, \ref{fig:rankings_picasso_test}, \ref{fig:rankings_adience_train_fm}, \ref{fig:rankings_adience_test_fm}, \ref{fig:rankings_adience_train_mf}, \ref{fig:rankings_adience_test_mf}, \ref{fig:rankings_teapot}, \ref{fig:rankings_vase}, \ref{fig:rankings_pathmnist_train}, \ref{fig:rankings_pathmnist_test} show occurrences of concepts from top-3 rules for the Aleph system (the 3 rules with the highest coverage of examples) given per concept rank. The ranking stems from the absolute value of the concept relevance. The x-axis shows the 10 ranks a given concept was ranked most frequently, calculated over all samples in a data set. Typically, high rankings (low numbers) in the 10 most occurring ranks are expected. 

\end{appendices}


\end{document}